\documentclass{article} 
\usepackage[preprint]{colm2026_conference}

\usepackage{microtype}
\usepackage{hyperref}
\usepackage{url}
\usepackage{booktabs}

\usepackage{bm}
\usepackage{bbm}
\usepackage{enumitem}
\usepackage{amsmath}
\usepackage{placeins}
\usepackage{graphicx}
\usepackage{multirow}
\usepackage{wrapfig}
\usepackage[T1]{fontenc}

\usepackage{lineno}

\definecolor{darkblue}{rgb}{0, 0, 0.5}
\hypersetup{colorlinks=true, citecolor=darkblue, linkcolor=darkblue, urlcolor=darkblue}

\title{Does Episodic Memory Help Close the Lexical Frequency Gap in Sensitivity to Syntactic Contrasts? A Test Using Retrieval-Augmented Language Models}

\author{Jing Liu\\
CoML Team, LSCP \\
ENS, Université PSL, EHESS, CNRS \\
\texttt{jing.liu@psl.eu} \\
\And
Najoung Kim \\
Department of Linguistics \\
Boston University \\
\texttt{najoung@bu.edu} \\
}

\begin{document}

\ifcolmsubmission
\linenumbers
\fi

\maketitle

\begin{abstract}
Grammatical knowledge and how it is empirically tested are typically considered robust to the frequency of the lexical items in the expressions. However, neural network-based models of grammaticality exhibit high sensitivity to lexical frequency. We draw upon Complementary Learning Systems theory to test the hypothesis that robustness to lexical frequency can arise via a hippocampal episodic memory mechanism, which enables rapid encoding and retrieval of specific experiences and allows learners to leverage them when processing rare patterns. We use retrieval-augmented language models as an instantiation of such an episodic memory mechanism (specifically, $k$-nearest-neighbor language models that augment parametric models with explicit instance storage), and test whether this augmentation helps close the lexical frequency gap that vanilla language models exhibit in syntactic contrast tests. 
Using syntactic contrasts with frequency-stratified test items, we find that retrieval augmentation narrows the performance gap between high- and low-frequency items, consistent with episodic memory compensating for weak parametric representations.
This benefit is consistent across different syntactic phenomena and across models pretrained on child-realistic and large-scale data. Additionally, we show that structural information is critical for effective retrieval, whereas semantic similarity alone provides little benefit. While these are promising proof-of-concept results supporting our hypothesis, the frequency gap is narrowed rather than fully closed. Based on our analyses, we propose preferential reweighting of retrieved instances, better representations and retrieval strategies for structural information, and flexible configurations of storage and retrieval as promising future directions for improving the implementation of episodic memory in language models.\footnote{Code is available at \url{https://github.com/Jing-L97/Generative\_replay}.}

\end{abstract}

\section{Introduction}

Language models are known to exhibit systematic lexical frequency biases: their performance on linguistic evaluations correlates with how often the lexical items in the test examples appear in training data, leading to low-frequency items posing a persistent challenge \citep{wei-etal-2021-frequency,kim2022uncontrolled,diehl-martinez-etal-2024-mitigating}. While frequency does influence human performance in empirical tests of grammatical knowledge such as acceptability judgments \citep[i.a.]{kempen2005relationship,bader2010toward,lau2017grammaticality}, recent work \citep{christensen2024complexity} has shown evidence for robustness to lexical frequency in delineating the critical contrasts. What accounts for this gap between human robustness and model fragility?

One possible answer comes from Complementary Learning Systems (CLS) theory 
\citep{mcclelland1995there,kumaran2016learning}, which proposes that the brain employs two distinct memory systems: a hippocampal system that rapidly encodes and retrieves specific experiences, and a neocortical system that gradually extracts generalizable statistical patterns. When parametric knowledge is weak, as it is for rare items, episodic memory can compensate by directly retrieving relevant past encounters \citep{davis2009complementary}. This division of labor has also been proposed to be relevant to linguistic processing \citep{borensztajn2011episodic,duff2012hippocampus}, leading us to hypothesize that episodic retrieval may reflect the mechanism that underlies human robustness to lexical frequency in syntactic processing.

In this work, we computationally test this hypothesis by asking whether augmenting language models with an explicit retrieval mechanism helps close the lexical frequency gap in syntactic contrast evaluations. We adopt $k$-nearest-neighbor language models ($k$NN-LM; \citealt{khandelwal2020knnlm}) as our proxy for episodic memory: the non-parametric datastore caches training instances for direct retrieval at inference time (analogous to hippocampal episodic traces), while the underlying parametric network consolidates statistical patterns through training (analogous to neocortical learning). The original $k$NN-LM has already been shown to be particularly helpful for rare patterns \citep{mallen2023trust,kandpal2023large}, and recent work shows that the information available in the context can be used more flexibly than the same information stored parametrically \citep{lampinen2025latent,chan2022transformers}, both consistent with the CLS prediction. At the same time, a recent finding that $k$NN-LMs do \textit{not} improve prediction of low-frequency tokens in standard language modeling evaluation \citep{nishida2025long} makes it a non-trivial and empirically open question whether retrieval benefits extend to the domain of syntactic knowledge.

We investigate this question through syntactic contrast tests across three different phenomena (subject-verb agreement, wh-questions, relative clauses) using frequency-stratified test items, under both child-realistic and large-scale pretraining. We further analyze what drives retrieval effectiveness by contrasting full versus semantics-only retrieval, by examining the role of structural similarity in retrieved instances, and by varying the configurations of retrieval (retrieval granularity, context window size, and number of retrieved neighbors). Our results show that retrieval augmentation consistently narrows the lexical frequency gap across all phenomena and both model scales, with greater gains for more syntactically complex phenomena. Nevertheless, the gap remains not fully closed. To this end, we analyze the limitations of the current modeling approach and posit preferential reweighting, improved representations and retrieval for structural information, and flexible configurations of storage and retrieval as promising future directions.

\section{Related Work}

\noindent \textbf{Frequency effects on linguistic evaluation in language models.}
Language models exhibit systematic frequency biases, where performance on linguistic evaluation often degrades when the test items contain low-frequency lexical items \citep{wei-etal-2021-frequency,kim2022uncontrolled}. In the context of syntactic contrasts, models perform worse on minimal pairs that contain low-frequency words, even when the underlying grammatical contrast is identical to high-frequency cases. \citet{diehl-martinez-etal-2024-mitigating} directly quantify this frequency bias in BLiMP (one of our syntactic test datasets) and propose a training-time intervention (Syntactic Smoothing) to reduce it, finding that substantial gaps persist even after mitigation.

\citet{wettig-etal-2023-mask} examine how the masking rate during training determines which co-occurring token pairs (measured via pointwise mutual information) are exposed to the model, shaping both convergence speed and final performance. Since pointwise mutual information is itself derived from corpus-level token frequencies, these findings are consistent with frequency distributions in pretraining data influencing model behavior, though this connection is inferential. More directly, \citet{kandpal2023large} show that language models struggle to acquire long-tail knowledge whose supporting documents are rare in the pretraining corpus, establishing a clear link between corpus frequency and what is learned.

Together, these findings establish that (lexical) frequency creates differential strength in parametric representations, with low-frequency items learned far less robustly. Our work asks a related question from a different methodological angle: rather than reducing frequency bias through better training, can it be compensated for at inference time through retrieval?

\noindent \textbf{Complementary Learning Systems theory.}
Complementary Learning Systems (CLS) theory \citep{mcclelland1995there, kumaran2016learning} proposes that the human brain employs two specialized memory systems: a hippocampal system that rapidly encodes specific episodes, and a neocortical system that gradually extracts statistical regularities across experiences. This allows new information to be acquired quickly without disrupting slower cortical learning, and computational implementations of CLS show that it enables both rapid memorization and gradual generalization \citep{mcclelland1995there}. Critically for our purposes, \citet{eichenbaum2001hippocampus} argues that the hippocampus does not merely store episodes but actively supports flexible use of past experiences through retrieval, enabling inferences that parametric learning alone may fail to achieve. Applied to language models, this predicts that when parametric representations are weak (as for low-frequency items), episodic retrieval should provide compensatory benefits by making specific relevant experiences directly accessible.  \citet{lampinen2025latent} in particular demonstrates that language models fail to generalize certain types of knowledge learned parametrically (e.g., relation reversals) but succeed when the same information is available in context, suggesting that episodic retrieval enables more flexible use of past experiences than parametric consolidation. Similarly, studies of in-context learning show that Transformers can apply procedures to information presented in context that they cannot apply to parametrically-stored knowledge \citep{chan2022transformers}. Our work extends these insights to syntactic processing by testing whether retrieval enables more robust generalization across frequency bands than parametric-only inferences.

\noindent \textbf{Retrieval-augmented language models.}  
Retrieval-augmented methods enhance language models by supplementing parametric knowledge with information drawn from an explicit memory at inference time. $k$NN-LMs \citep{khandelwal2020knnlm}, which we adopt in this work, interpolate the base model's next-token distribution with a distribution computed from nearest-neighbor contexts retrieved from a datastore of training instances. While prior work shows benefits for rare patterns and knowledge-intensive tasks \citep{borgeaud2022improving,kandpal2023large,mallen2023trust}, recent work finds that standard $k$NN-LMs do not improve prediction performance for low-frequency tokens in language modeling evaluation \citep{nishida2025long}, leaving open questions about retrieval benefits for linguistic processing.

\section{Method}
\label{sec:method}

We first show that frequency gaps in tests of grammatical knowledge exist in vanilla language models (``base models''), and then test whether $k$NN-LMs help close these gaps. To this end, we use targeted syntactic contrast datasets where we divide test instances based on the frequency of the content words they contain. We describe the details of the models, datasets, evaluation tasks and the retrieval procedure below. 

\subsection{Models and Training}

\noindent \textbf{Base models.} We evaluate two decoder-only Transformer models that vary in parameter size and training data scale: (1) GPT-2 XL (1.5B parameters), pretrained on large-scale web text \citep{radford2019language}, and (2) GPT-2 Small (124M parameters) pretrained on child-directed speech corpora with 50M words selected from the BabyLM Challenge \citep{hu2024findings}. Both use a standard autoregressive objective and serve as parametric-only baselines. Further details are provided in Appendix~\ref{app:model_dataset}.

\noindent \textbf{$k$NN language models.} Following \citet{khandelwal2020knnlm}, we augment base models with a datastore $\mathcal{D}$ containing context--target pairs. For each token $x_t$, we store the key-value pair: $(\bm{f}_\theta(\bm{x}_{<t}), x_t)$, where $\bm{f}_\theta(\bm{x}_{<t})$ is the layer-normalized contextual
representation fed as input to the feed-forward sublayer of the final Transformer block. For GPT-2, this corresponds to the output of the final block's second layer-normalization module and the input to its MLP. At inference time, we:
\vspace{-0.2cm}
\begin{enumerate}[itemsep=0.1pt, parsep=0pt]
    \item Compute the query: $\bm{q}_t = \bm{f}_\theta(\bm{x}_{<t})$
    \item Retrieve $k$ nearest neighbors via squared L2 distance
    \item Compute the $k$NN probability: $p_{\text{kNN}}(x_t|\bm{x}_{<t}) \propto \sum_{(\bm{k},v) \in \mathcal{N}} \mathbbm{1}_{x_t=v} \exp(-d(\bm{k},\bm{q}_t)/\tau)$
    \item Interpolate: $p(x_t|\bm{x}_{<t}) = \lambda p_{\text{kNN}} + (1-\lambda) p_{\text{LM}}$
\end{enumerate}

\noindent We fix $\lambda=0.25$ following \citet{khandelwal2020knnlm} and vary $k \in \{1, 16, 1024\}$, using FAISS \citep{johnson2019billion} for retrieval.

\subsection{Evaluation Task}

\textbf{Syntactic contrasts.} We evaluate sensitivity to three syntactic phenomena: Subject-Verb agreement (SV), Wh-questions (Wh), and Relative Clauses (RC). We collect examples in these categories from BLiMP \citep{warstadt2020blimp}, Zorro \citep{huebner2021babyberta}, and BIG-bench \citep{srivastava2023beyond}, filtering examples containing out-of-vocabulary items for each model. Each test item is a minimal pair with one grammatical and one ungrammatical sentence, where models must assign a higher probability to the grammatical sentence.

\noindent \textbf{Frequency stratification.} To test our hypothesis regarding lexical frequency, we further divided the evaluation data based on mean frequency across all the content words in the sentences (\textit{high-frequency} ($f > 10^4$) versus \textit{low-frequency} ($f < 10^3$); everything in-between \textit{mid-frequency}). Our main hypothesis concerns the contrast between the high- and low-frequency bands; we report the mid-frequency band throughout to give a complete picture of how retrieval benefits scale with frequency.

\noindent We emphasize that all frequency gap claims are scoped \emph{within} a single model/corpus: each gap and its narrowing under retrieval is measured against the same model's own baseline. Because the two pretraining corpora differ substantially in size and composition, the absolute frequency thresholds do not render items in the same band directly comparable across pretraining settings, and we therefore do not compare numerical trends across the two settings.

\begin{wraptable}{r}{0.48\textwidth}
\vspace{-5pt}
\caption{Syntactic contrast accuracy and perplexity (PPL) for baseline and kNN-LM models ($k=16$, $\tau=3$, sequence-level retrieval). kNN augmentation improves accuracy across all phenomena and both pretraining scales, with relative clauses showing the largest gains. Perplexity reductions are larger for the large-scale model. Bold indicates improvement over baseline.}
\vspace{6pt}
\label{tab:general_effects}
\centering
\small
\begin{tabular}{lcccc}
\toprule
\textbf{} & \multicolumn{2}{c}{\textbf{Child-realistic}} & 
\multicolumn{2}{c}{\textbf{Large-scale}} \\
 & baseline & kNN & baseline & kNN \\
\midrule
\textbf{PPL.} $\downarrow$ & 14.88 & \textbf{14.30} & 13.68 & \textbf{11.60} \\
\textbf{SV} $\uparrow$     & 0.76  & \textbf{0.82}  & 0.84  & \textbf{0.89} \\
\textbf{Wh} $\uparrow$     & 0.77  & \textbf{0.81}  & 0.81  & \textbf{0.86} \\
\textbf{RC} $\uparrow$     & 0.68  & \textbf{0.78}  & 0.65  & \textbf{0.84} \\
\bottomrule
\end{tabular}
\vspace{-5pt}
\end{wraptable}

\subsection{Retrieval Procedure}
\label{subsec:datastores}

We construct three datastore variants to disentangle the contributions of structural and semantic information: (1) \textit{Full Retrieval}: standard contextualized embeddings encoding both semantic and syntactic information; (2) \textit{Semantics-only (averaged)}: position-averaged contextualized embeddings from (1);\footnote{In practice, averaging contextualized embeddings may not fully scrub syntactic information; we return to this issue later.} and (3) \textit{Semantics-only (uncontextualized)}: summed GloVe \citep{pennington2014glove} and fastText \citep{bojanowski2017enriching} embeddings, capturing lexical semantics without structural information. We additionally vary three retrieval configuration parameters: granularity of retrieved instances (token, phrase, or sequence), context window size ($\tau$), and number of retrieved neighbors ($k$), to characterize which retrieval configurations are most effective across syntactic phenomena.

\section{Results}

\subsection{Retrieval Benefits Syntactic Contrasts}
We begin by investigating the general effectiveness of retrieval augmentation on syntactic contrasts. Table~\ref{tab:general_effects}\footnote{Note for all tables in this section: Baseline models use parametric knowledge only; kNN models are retrieval-augmented. Abbreviations: SV (subject-verb agreement), Wh (wh-questions), RC (relative clauses), PPL (perplexity). We use $k=16$, $\tau=3$, and sequence-level retrieval (see Section~\ref{subsec:datastores}) as the representative configuration, since this setting yields general gains in low-frequency conditions. However, as discussed in Section~\ref{sec:retrieval_configuration}, there may exist alternative configurations that outperform this representative configuration for individual phenomena.} shows that $k$NN augmentation improves performance across all three syntactic phenomena and both pretraining scales, as well as replicating previously reported perplexity improvements \citep{khandelwal2020knnlm}. Relative clauses benefit the most from retrieval augmentation across both model scales. This pattern holds consistently regardless of pretraining scale, though perplexity reductions are larger for the large-scale model. The disproportionate benefit for relative clauses suggests that retrieval is most helpful where parametric representations are weakest, which motivates us to analyze the performance gains stratified by frequency.

\subsection{Frequency Gap is Narrowed but not Fully Closed}
\label{sec:frequency_effects}

\begin{wraptable}{r}{0.55\textwidth}
\centering
\caption{Frequency-stratified syntactic contrast performance($k=16$, $\tau=3$; sequence-level retrieval). ER denotes error reduction,
$\Delta\mathrm{acc}/(1-\mathrm{acc}_{\mathrm{base}})$. Bold indicates improvement over baseline.}
\vspace{6pt}
\label{tab:frequency_effects}
\small
\setlength{\tabcolsep}{4pt}
\renewcommand{\arraystretch}{0.96}
\begin{tabular}{@{}llcccccc@{}}
\toprule
& & \multicolumn{3}{c}{\textbf{Child-realistic}} &
    \multicolumn{3}{c}{\textbf{Large-scale}} \\
\cmidrule(lr){3-5} \cmidrule(lr){6-8}
& & base & kNN & ER (\%) & base & kNN & ER (\%) \\
\midrule
\multirow{3}{*}{\textbf{SV}}
& high & 0.91 & 0.90 & $-11.1$ & 0.96 & 0.95 & $-25.0$ \\
& mid  & 0.82 & \textbf{0.83} & 5.6
       & 0.90 & 0.89 & $-10.0$ \\
& low  & 0.60 & \textbf{0.76} & 40.0
       & 0.71 & \textbf{0.83} & 41.4 \\
\midrule
\multirow{3}{*}{\textbf{Wh}}
& high & 0.92 & 0.88 & $-50.0$ & 0.93 & 0.93 & 0.0 \\
& mid  & 0.83 & \textbf{0.86} & 17.6
       & 0.90 & \textbf{0.91} & 10.0 \\
& low  & 0.61 & \textbf{0.70} & 23.1
       & 0.69 & \textbf{0.74} & 16.1 \\
\midrule
\multirow{3}{*}{\textbf{RC}}
& high & 0.88 & \textbf{0.89} & 8.3
       & 0.89 & \textbf{0.95} & 54.5 \\
& mid  & 0.73 & \textbf{0.81} & 29.6
       & 0.82 & \textbf{0.89} & 38.9 \\
& low  & 0.48 & \textbf{0.65} & 32.7
       & 0.41 & \textbf{0.70} & 49.2 \\
\bottomrule
\end{tabular}
\end{wraptable}

Next, we examine whether retrieval helps close the lexical frequency gap. Table~\ref{tab:frequency_effects} confirms the existence of frequency gaps in both baseline models across three syntactic phenomena, with performance on low-frequency items consistently lower than on high-frequency items, and mid-frequency items falling in between. With retrieval augmentation, the high--low performance gap narrows across all phenomena and both model scales. In terms of error reduction ($\Delta\text{acc}/(1-\text{acc}_{\text{base}})$), low-frequency items show substantial relative gains (e.g., $49.2\%$ for RCs in the large-scale model), though we note that this metric can be misleading near ceiling, where small absolute changes in already-high accuracy produce large negative percentages (e.g., child-realistic high-frequency Wh, $0.92 \rightarrow 0.88$, yields $-50\%$). Raw accuracies should be used to adjust the interpretations in such cases. 

Statistical analysis supports the interpretation that narrowing is driven primarily by gains on low-frequency items. Bootstrapped $95\%$ confidence intervals on the per-frequency band performance delta (kNN $-$ baseline) confirm a robust separation: for the child-realistic model, low-frequency items show $\Delta = 0.140$ ($95\%$ CI $[0.115, 0.166]$) versus $\Delta = -0.013$ ($95\%$ CI $[-0.030, 0.004]$) for high-frequency items, with non-overlapping intervals (large-scale model shows the same pattern; see Appendix~\ref{app:stats}). 

A mixed-effects logistic regression using treatment coding with high-frequency items and the baseline model as reference levels (\texttt{correct $\sim$ Frequency * ModelType + (1$\mid$item) + (1$\mid$phenomenon)}) further shows that the retrieval effect is significantly larger for low-frequency than for high-frequency items (Frequency [low] $\times$ ModelType [kNN] interaction: $\beta = 0.52$, $p < 0.001$. The effect of retrieval within the high-frequency band is not significant (ModelType [kNN]: $\beta = -0.04$, $p = 0.62$; Appendix~\ref{app:stats}). 

Nevertheless, this compensation is not sufficient to eliminate the frequency gap: low-frequency items still yield lower accuracy than high-frequency items across all tested phenomena and both model scales, even after retrieval augmentation.

\section{Analysis}

\noindent The analyses in this section fix the retrieval configuration to $k=16$, $\tau=3$ with sequence-level retrieval, which isolates the contribution of the factors studied from the effects of configuration. 

\subsection{Importance of Structural vs. Semantic Information}
\label{sec:multidimensional_binding}

We proceed to investigate which aspects of the retrieved information drive the observed benefits by examining (1) the effect of semantics-only retrieval, and (2) the effect of retrieving structurally similar instances.

\begin{figure*}[ht]
    \centering
    \begin{minipage}[t]{0.32\textwidth}
        \centering
        \includegraphics[width=\linewidth]{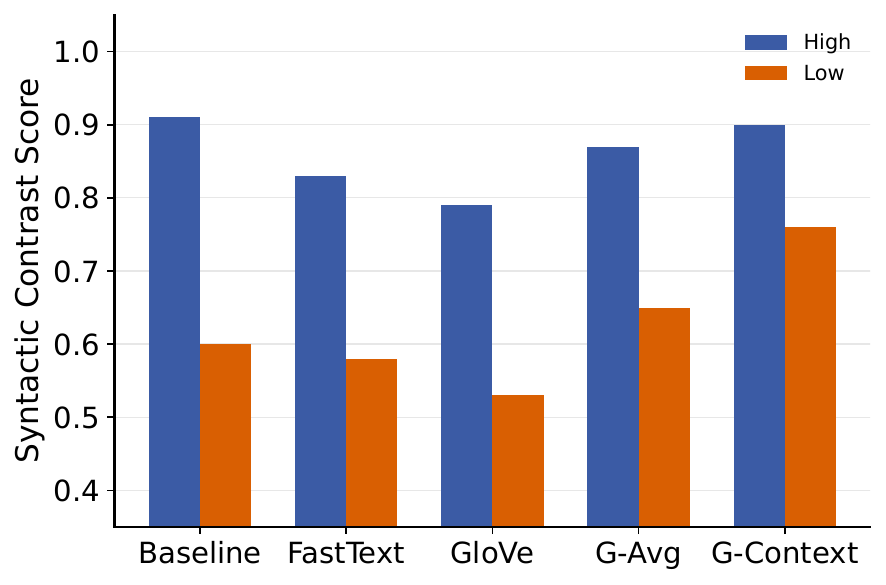}
    \end{minipage}
    \hfill
    \begin{minipage}[t]{0.32\textwidth}
        \centering
        \includegraphics[width=\linewidth]{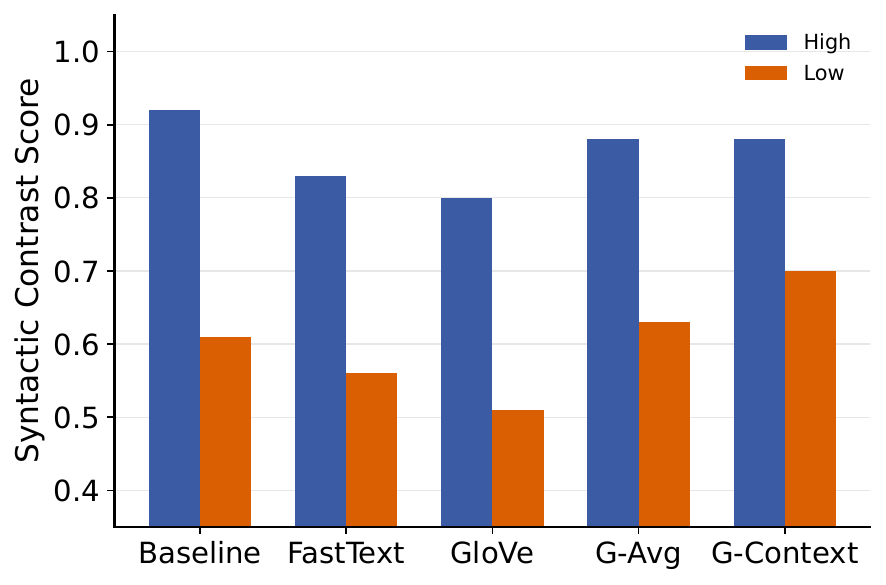}
    \end{minipage}
    \hfill
    \begin{minipage}[t]{0.32\textwidth}
        \centering
        \includegraphics[width=\linewidth]{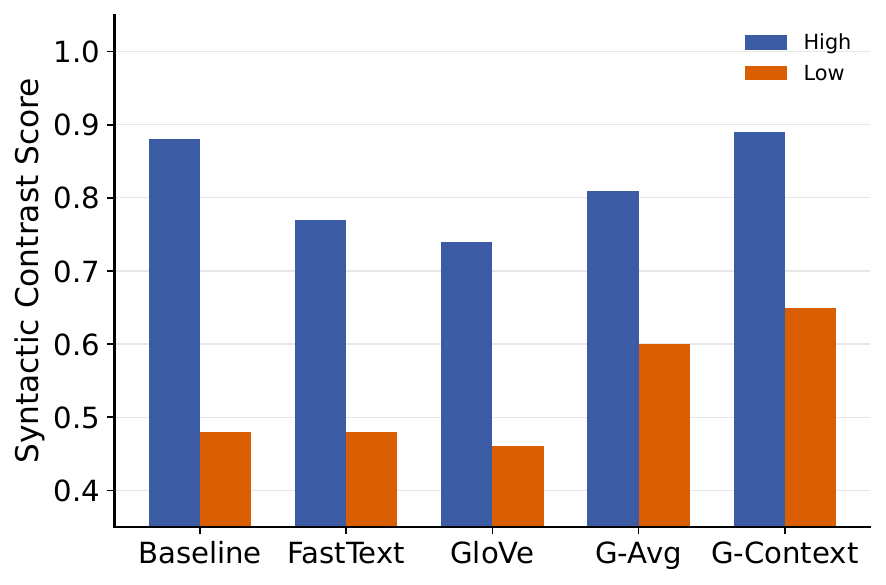}
    \end{minipage}

    \vspace{2mm} 

    \begin{minipage}[t]{0.32\textwidth}
        \centering
        \includegraphics[width=\linewidth]{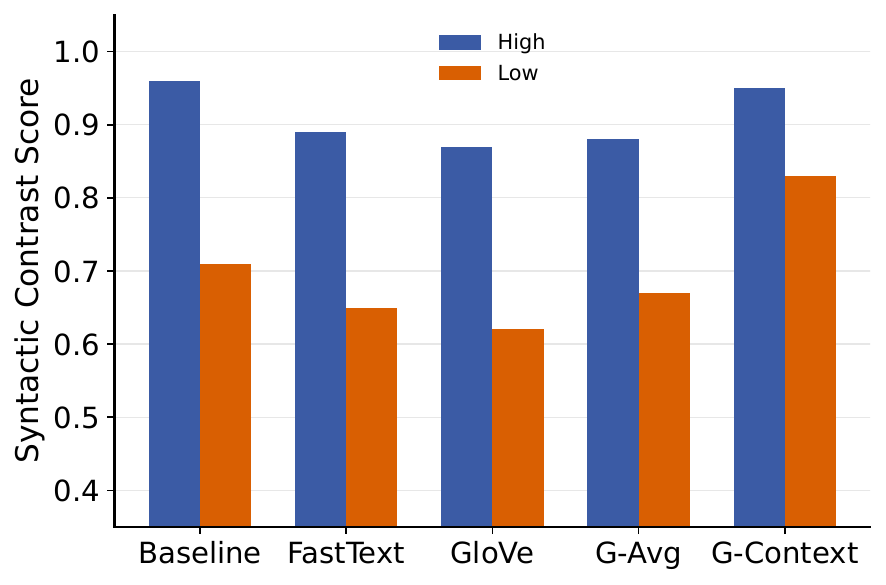}
        (a) Subject-verb agreement
    \end{minipage}
    \hfill
    \begin{minipage}[t]{0.32\textwidth}
        \centering
        \includegraphics[width=\linewidth]{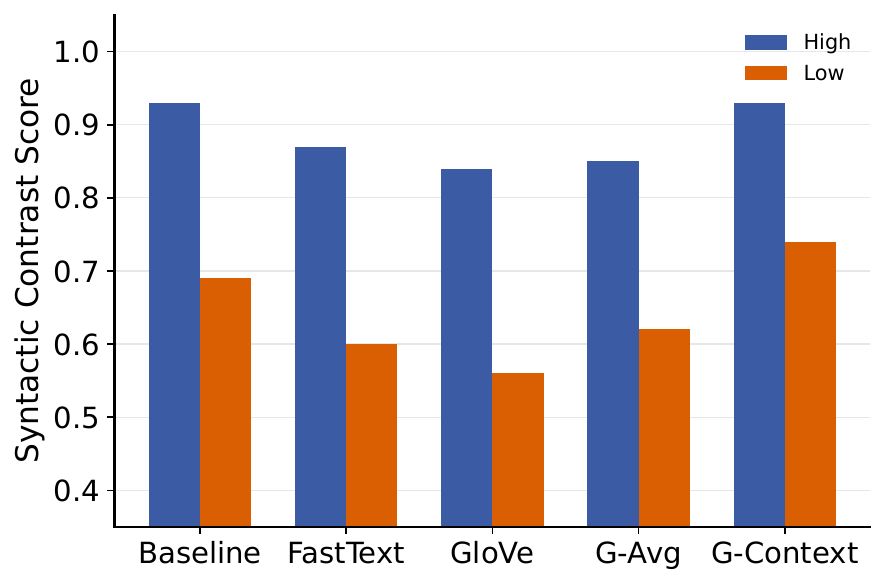}
        (b) Wh-questions
    \end{minipage}
    \hfill
    \begin{minipage}[t]{0.32\textwidth}
        \centering
        \includegraphics[width=\linewidth]{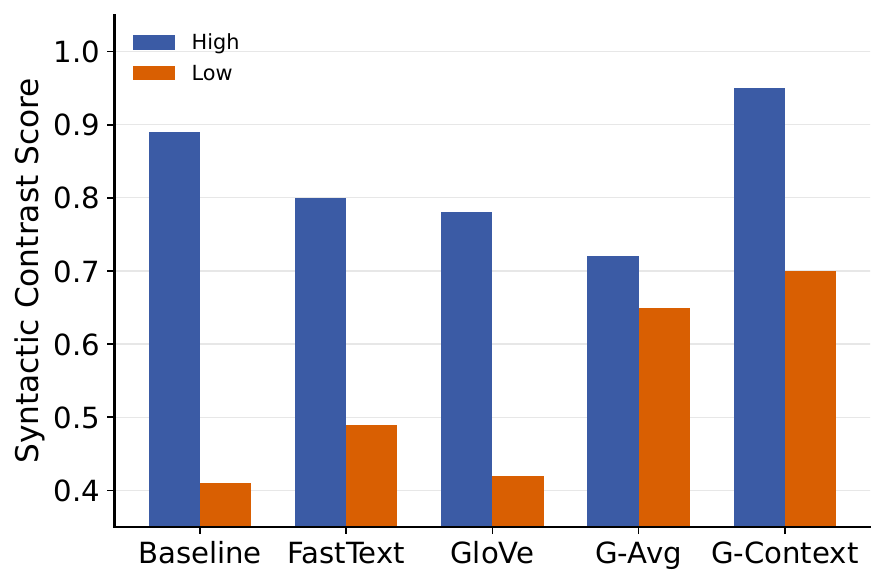}
        (c) Relative Clauses
    \end{minipage}
    
    \vspace{2mm} 

    \caption{Performance on syntactic contrasts using semantics-only retrieval strategies compared to full retrieval ($k=16$, $\tau=3$) of the child-realistic (top row) and large-scale (bottom row) models. Blue bars indicate high-frequency items and orange bars low-frequency items.
    }
    \label{fig:vector}
\end{figure*}

\noindent \textbf{Semantics-only retrieval is mostly ineffective.} 
Figure~\ref{fig:vector} shows that semantics-only retrieval (FastText, GloVe, G-Avg; see Section~\ref{subsec:datastores}) performs at or substantially \textit{below} the baseline without retrieval across most conditions. This shows that semantic similarity of the retrieved examples alone does not reliably improve syntactic judgments. One partial exception is GPT2-Averaged (G-Avg), where retrieval does lead to substantive gains over the baseline for low-frequency items for RCs, although performance on high-frequency items degrades by a large margin. FastText and GloVe also yield gains for low-frequency RCs in the large-scale setting, although relatively minor. There are two possible, non-mutually exclusive interpretations for this: (1) semantic information does sometimes yield benefits, as discussed in the next paragraph; or (2) that averaging of contextualized embeddings does not totally remove structural information.

\begin{figure*}[ht]
    \centering
    \begin{minipage}[t]{0.32\textwidth}
        \centering
        \includegraphics[width=\linewidth]{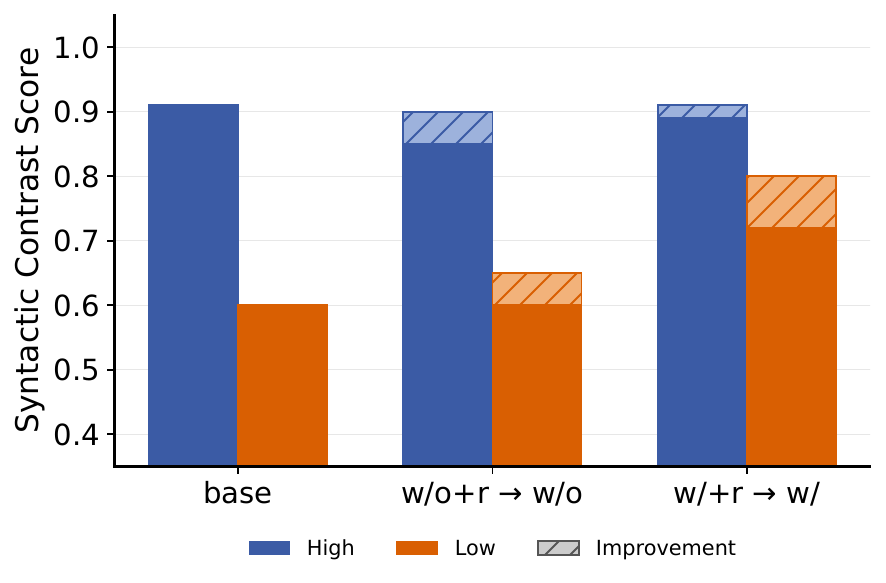}
    \end{minipage}
    \hfill
    \begin{minipage}[t]{0.32\textwidth}
        \centering
        \includegraphics[width=\linewidth]{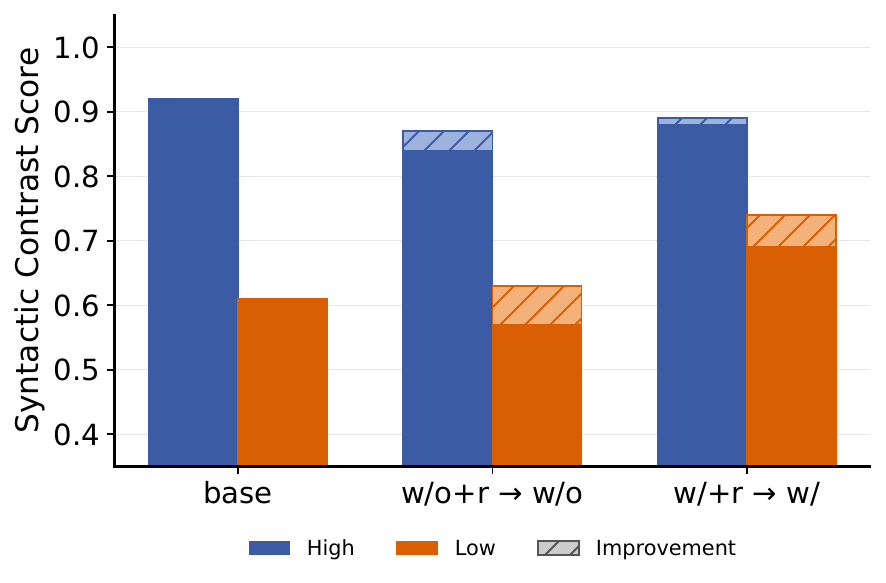}
    \end{minipage}
    \hfill
    \begin{minipage}[t]{0.32\textwidth}
        \centering
        \includegraphics[width=\linewidth]{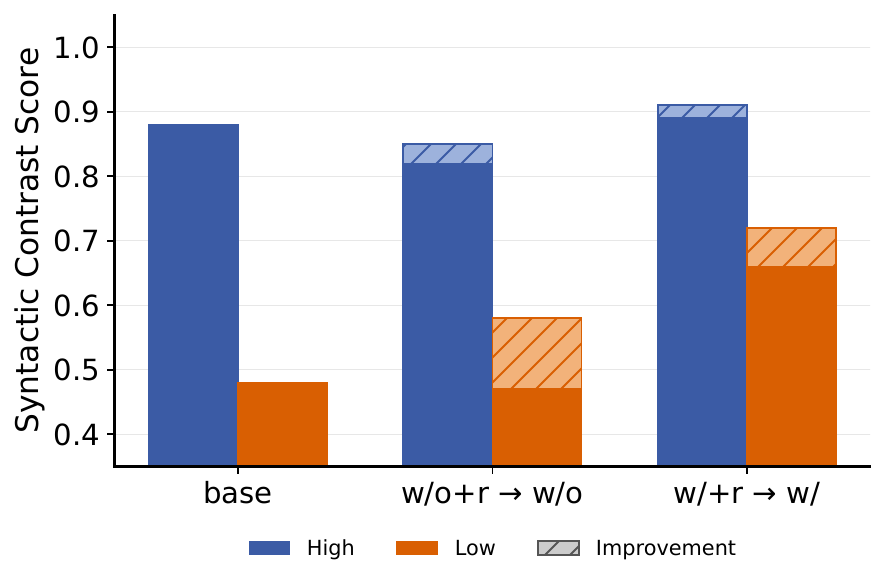}
    \end{minipage}
    
    \vspace{2mm} 

    \begin{minipage}[t]{0.32\textwidth}
        \centering
        \includegraphics[width=\linewidth]{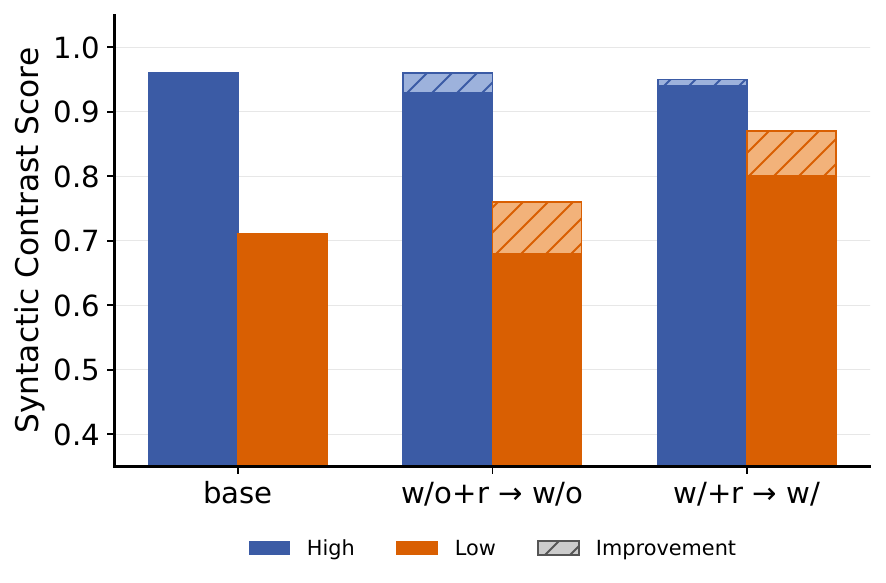}
        (a) Subject-verb agreement
    \end{minipage}
    \hfill
    \begin{minipage}[t]{0.32\textwidth}
        \centering
        \includegraphics[width=\linewidth]{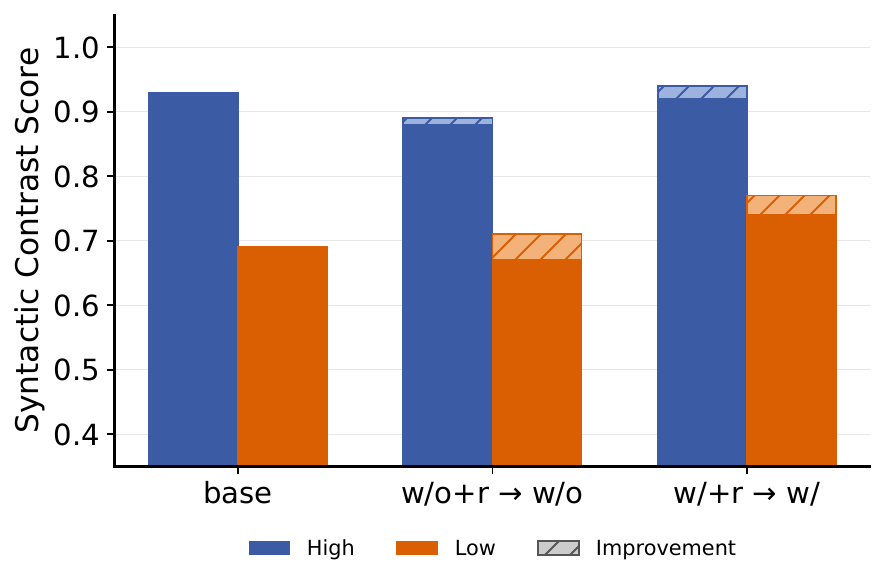}
        (b) Wh-questions
    \end{minipage}
    \hfill
    \begin{minipage}[t]{0.32\textwidth}
        \centering
        \includegraphics[width=\linewidth]{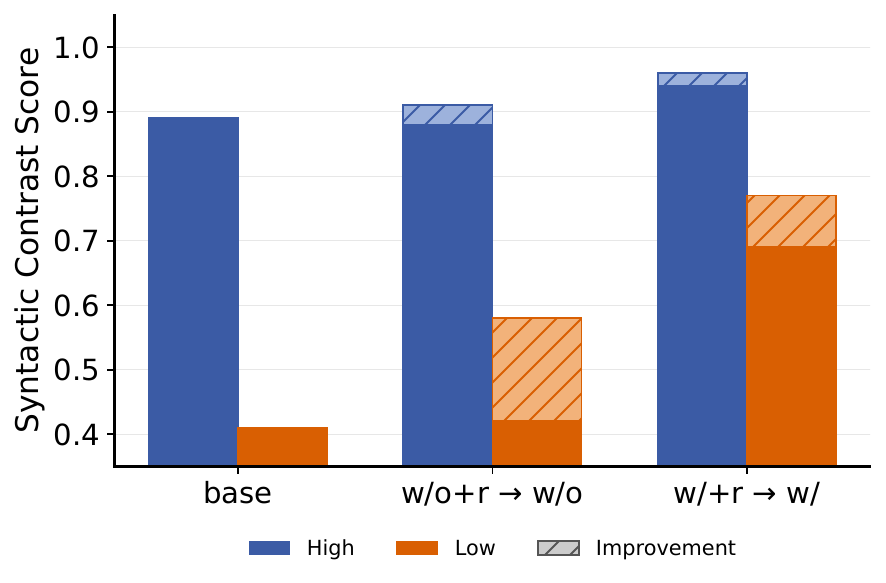}
        (c) Relative clauses
    \end{minipage}

    \caption{Performance when retrieved sequences contain target syntactic structures (with: w/) versus when they do not (without: w/o). Top row: child-realistic model; bottom row: large-scale model ($k=16$, $\tau=3$, sequence-level). Blue and orange bars correspond to high- and low-frequency items, respectively. Within each panel, the $x$-axis shows three conditions: \textbf{base} (no retrieval), \textbf{w/o+r $\rightarrow$ w/o} (retrieved set does not contain structurally matched examples, random selection $\rightarrow$ embedding-similarity), and \textbf{w/+r $\rightarrow$ w/} (retrieved set contains structurally matched examples, random selection $\rightarrow$ embedding-similarity). Solid bars denote random selection (\textbf{+r}); hatched segments denote the additional gain from embedding-similarity ranking over random selection.}
    \label{fig:datastore}
\end{figure*}

\noindent \textbf{Structural information is critical, and similarity-based example selection does not compensate for missing structure.} Figure~\ref{fig:datastore} examines retrieval benefits from two angles: whether the set of retrieved items contains structurally matching examples (\emph{with (w)} vs.\ \emph{without (w/o)}),\footnote{See Appendix~\ref{app:struc_simi} for our structural match metric.} and whether retrieval ranks them by embedding similarity or at random (\emph{+r}). Across all phenomena and both model scales, the \emph{w} bars consistently exceed \emph{w/o} for low-frequency items (orange bars) with the strongest effects for relative clauses. This illustrates the importance of structure-matched examples in the retrieved set. For low-frequency items, both structural matches and similarity-based selection contribute. However, similarity does not compensate for missing structural information. While it does add smaller, consistent gains overall, similarity-based selection without structural matches (\textit{w/o}: middle orange bars with hatched gains) consistently fails to outperform random selection with structural matches (\textit{w/+r}: right orange bars without hatched gains) for any model or phenomenon. Thus, structurally matched examples are the primary driver of frequency-selective retrieval benefits, with secondary contributions from similarity-based selection. Given the ineffectiveness of semantics-only retrieval discussed above, the embedding similarity gains likely reflect partial structural similarity between the retrieved examples and the target, in addition to semantic similarity.

\subsection{Effects of Retrieval Configuration}
\label{sec:retrieval_configuration}

We next analyze whether and how retrieval effectiveness is modulated by three configuration parameters: retrieval granularity (token, phrase, or sequence), number of neighbors ($k$), and context window size ($\tau$).

\begin{figure*}[ht]
    \centering
    \begin{minipage}[t]{0.32\textwidth}
        \centering
        \includegraphics[width=\linewidth]{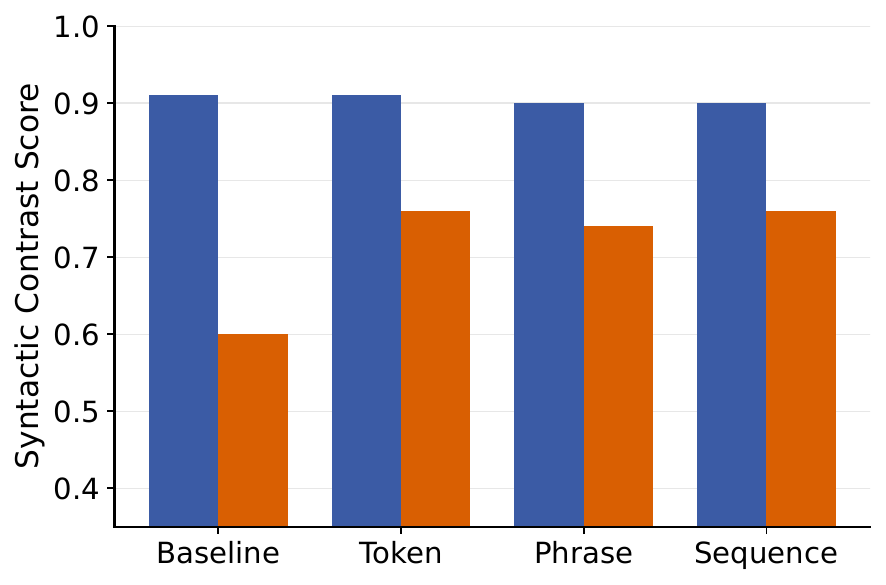}
    \end{minipage}
    \hfill
    \begin{minipage}[t]{0.32\textwidth}
        \centering
        \includegraphics[width=\linewidth]{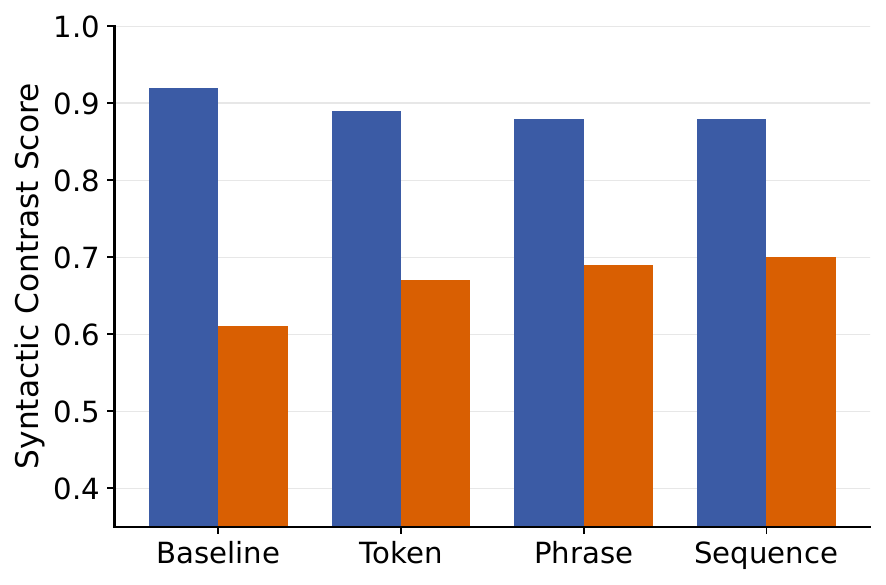}
    \end{minipage}
    \hfill
    \begin{minipage}[t]{0.32\textwidth}
        \centering
        \includegraphics[width=\linewidth]{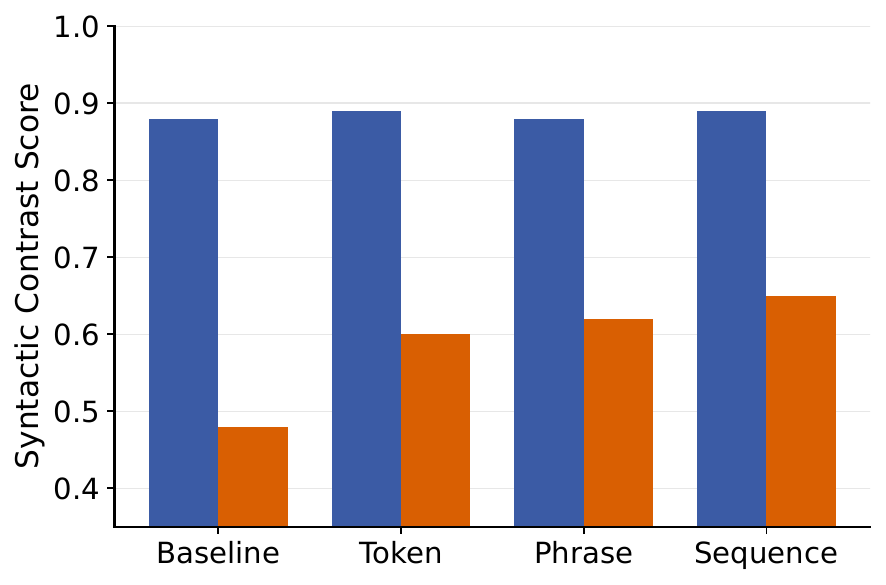}
    \end{minipage}
    
    \vspace{2mm} 

    \begin{minipage}[t]{0.32\textwidth}
        \centering
        \includegraphics[width=\linewidth]{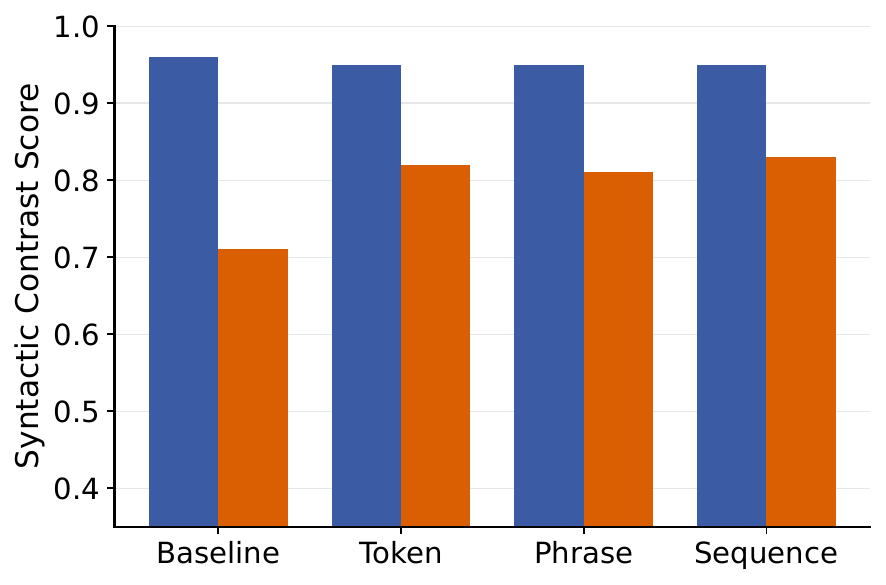}
        (a) Subject-verb agreement
    \end{minipage}
    \hfill
    \begin{minipage}[t]{0.32\textwidth}
        \centering
        \includegraphics[width=\linewidth]{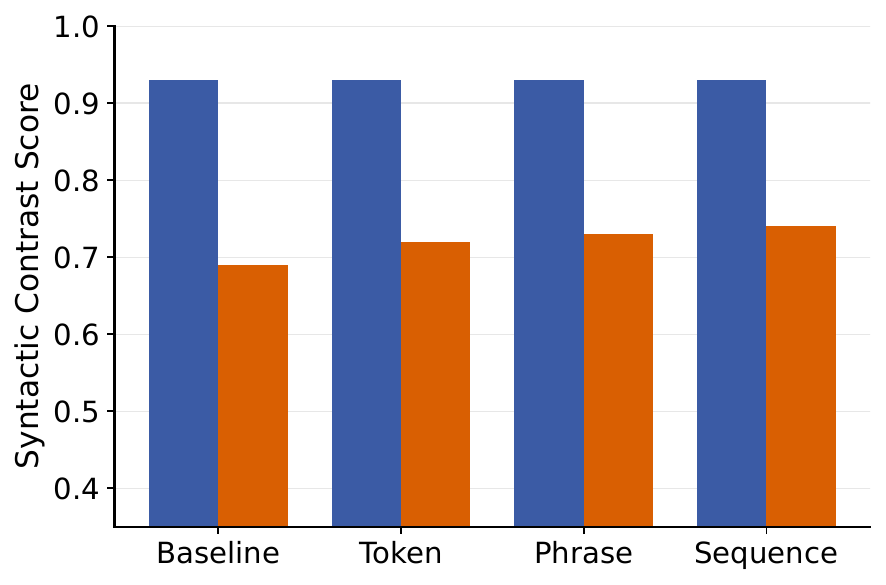}
        (b) Wh-questions
    \end{minipage}
    \hfill
    \begin{minipage}[t]{0.32\textwidth}
        \centering
        \includegraphics[width=\linewidth]{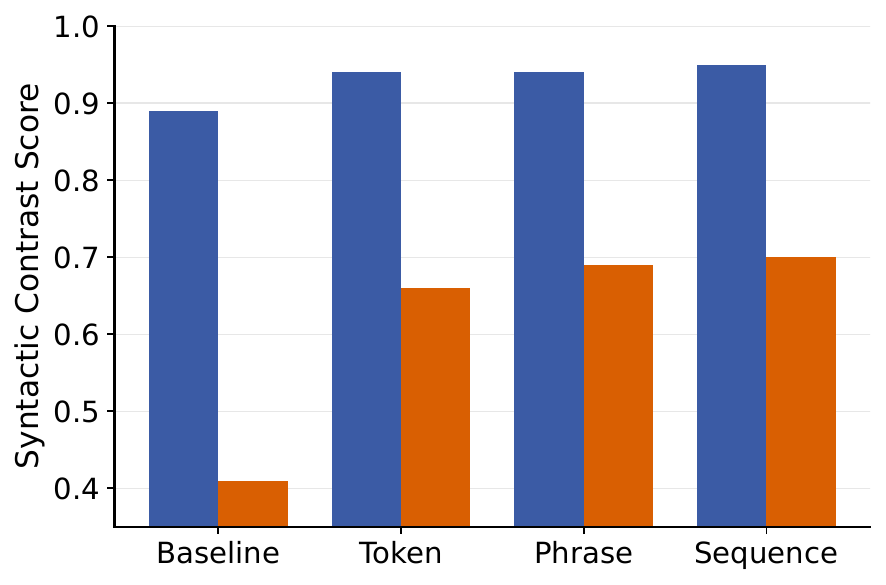}
        (c) Relative clauses
    \end{minipage}

    \caption{Performance across retrieval granularities (token, phrase, sequence) in child-realistic (top row) and large-scale (bottom row) models ($k=16$, $\tau=3$). Blue bars indicate high-frequency items and orange bars low-frequency items. Sequence-level retrieval is numerically the overall strongest, but the effect of retrieval granularity is overall small.}

    \label{fig:granularity}
\end{figure*}

\noindent \textbf{Effects of retrieval granularity.}
Figure~\ref{fig:granularity} shows that, at $k=16$ and $\tau=3$, sequence-level retrieval numerically provides the strongest or tied-strongest performance for low-frequency items across all phenomena and both model settings, but the effect of retrieval granularity overall is small. 

\begin{figure*}[ht]
    \centering
    \begin{minipage}[t]{0.32\textwidth}
        \centering
        \includegraphics[width=\linewidth]{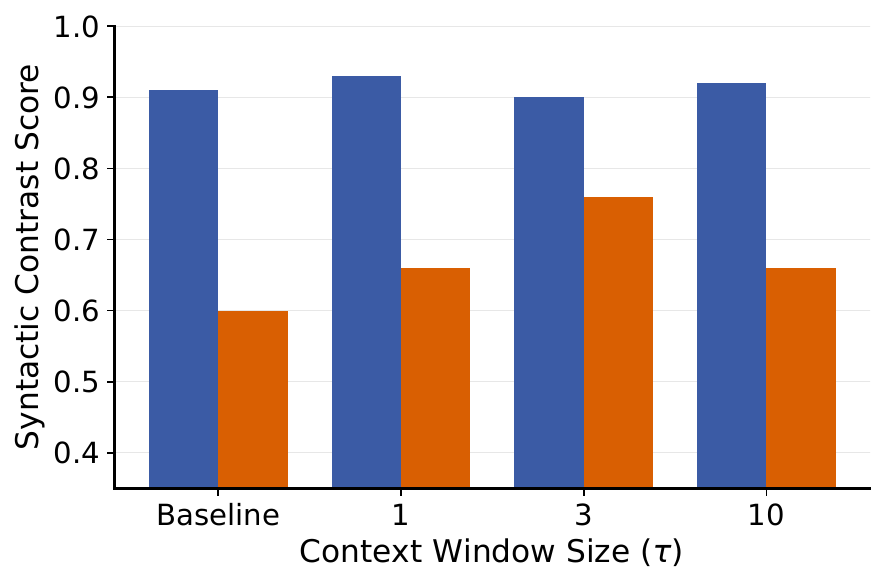}
    \end{minipage}
    \hfill
    \begin{minipage}[t]{0.32\textwidth}
        \centering
        \includegraphics[width=\linewidth]{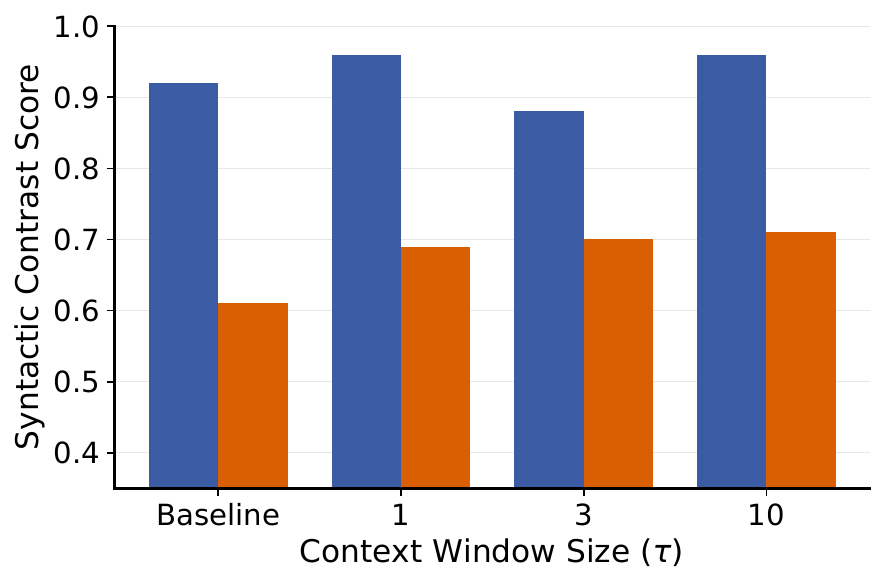}
    \end{minipage}
    \hfill
    \begin{minipage}[t]{0.32\textwidth}
        \centering
        \includegraphics[width=\linewidth]{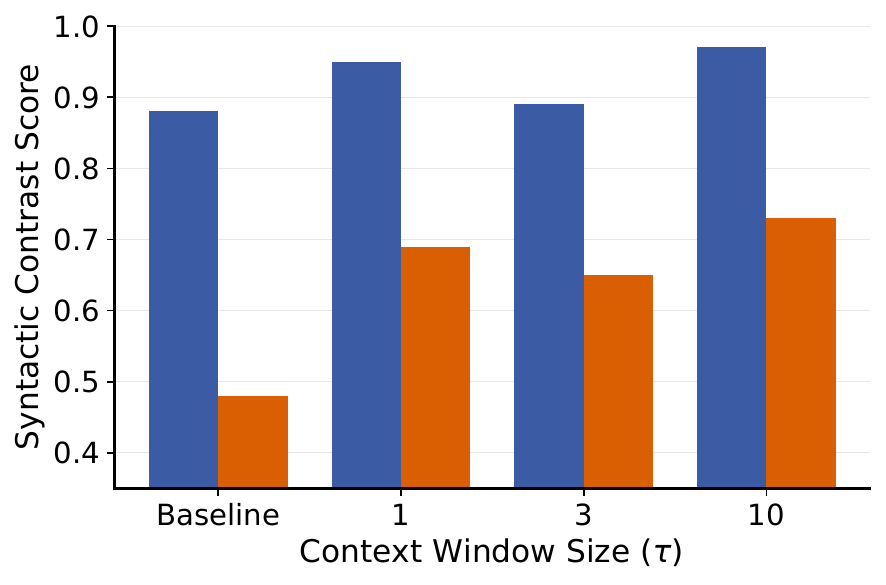}
    \end{minipage}

    \vspace{2mm}

    \begin{minipage}[t]{0.32\textwidth}
        \centering
        \includegraphics[width=\linewidth]{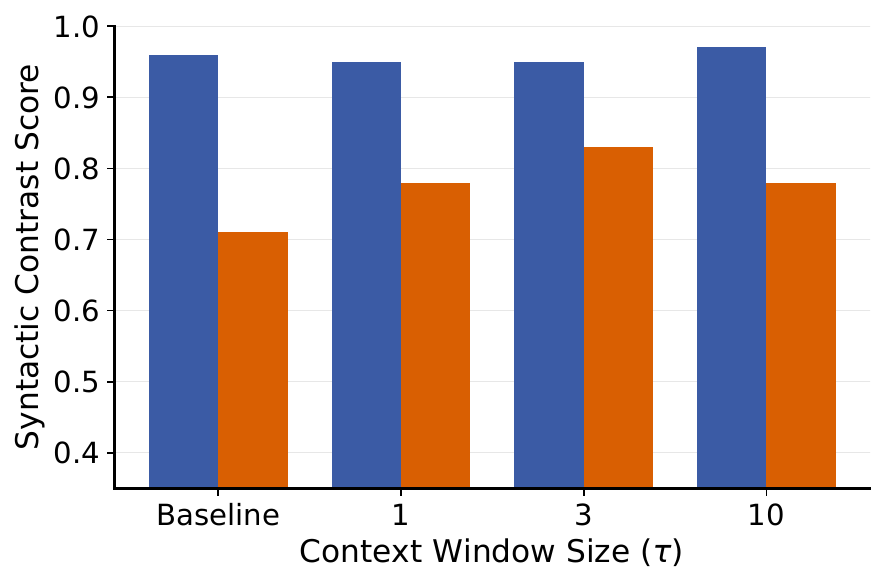}

        (a) Subject-verb agreement
    \end{minipage}
    \hfill
    \begin{minipage}[t]{0.32\textwidth}
        \centering
        \includegraphics[width=\linewidth]{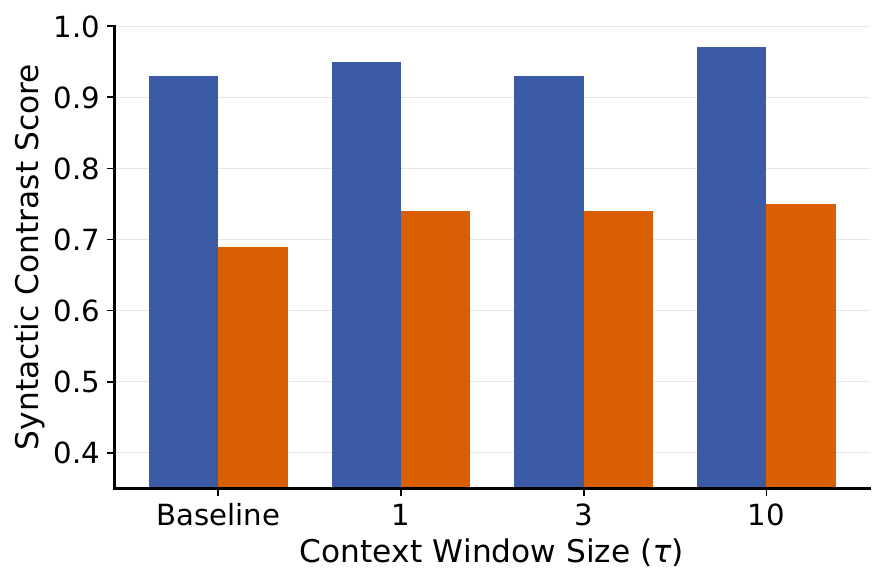}

        (b) Wh-questions
    \end{minipage}
    \hfill
    \begin{minipage}[t]{0.32\textwidth}
        \centering
        \includegraphics[width=\linewidth]{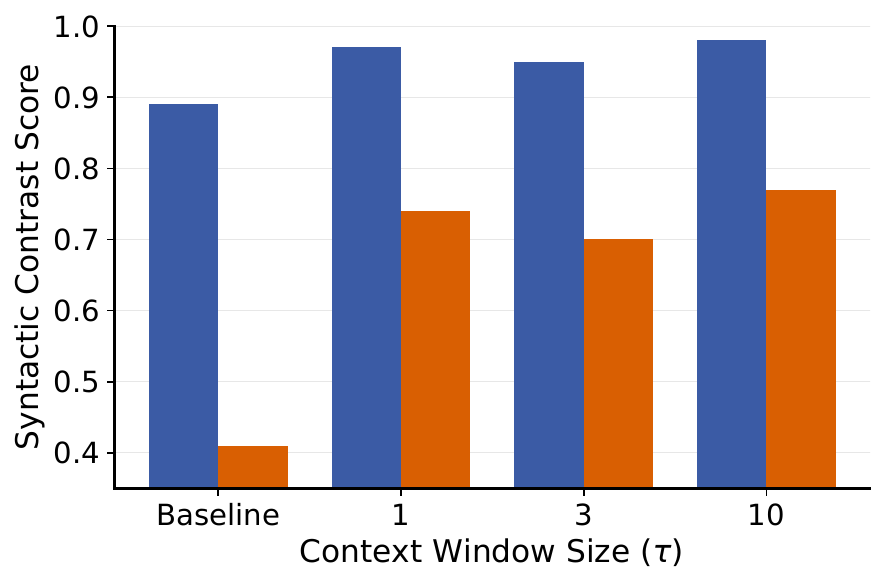}

        (c) Relative clauses
    \end{minipage}

    \caption{
        Performance across context-window sizes for the child-realistic
        (top row) and large-scale (bottom row) settings, using
        sequence-level retrieval with $k=16$. Blue bars indicate
        high-frequency items and orange bars indicate low-frequency
        items. The effects of context size are non-monotonic and
        phenomenon-, frequency band- and model-dependent.
    }
    \label{fig:context_effects}
\end{figure*}

\noindent \textbf{Effects of context window size.} 
Figure~\ref{fig:context_effects} shows that the effect of context window size is idiosyncratic. For RC, $\tau=10$ yields the highest accuracy across both model scales and both frequency bands, whereas for SV, $\tau=3$ yields the strongest performance for low-frequency. For Wh, the effect of $\tau$ is minimal, except $\tau=3$ leading to degradation in high-frequency items in the child-realistic model. Thus, increasing context window size does not uniformly improve performance, and the preferred window size depends on the phenomenon, frequency band, and model.

\begin{figure*}[ht]
    \centering

    \begin{minipage}[t]{0.32\textwidth}
        \centering
        \includegraphics[width=\linewidth]
        {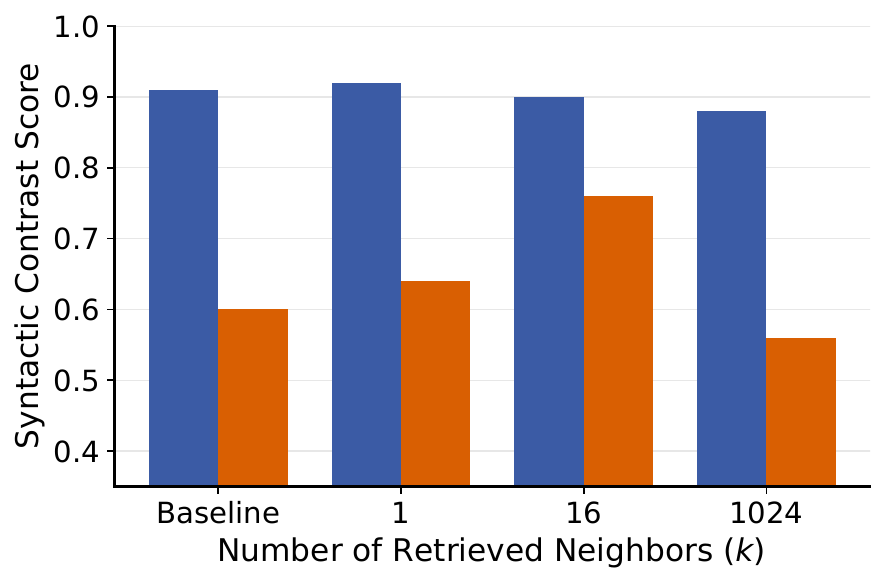}
    \end{minipage}
    \hfill
    \begin{minipage}[t]{0.32\textwidth}
        \centering
        \includegraphics[width=\linewidth]
        {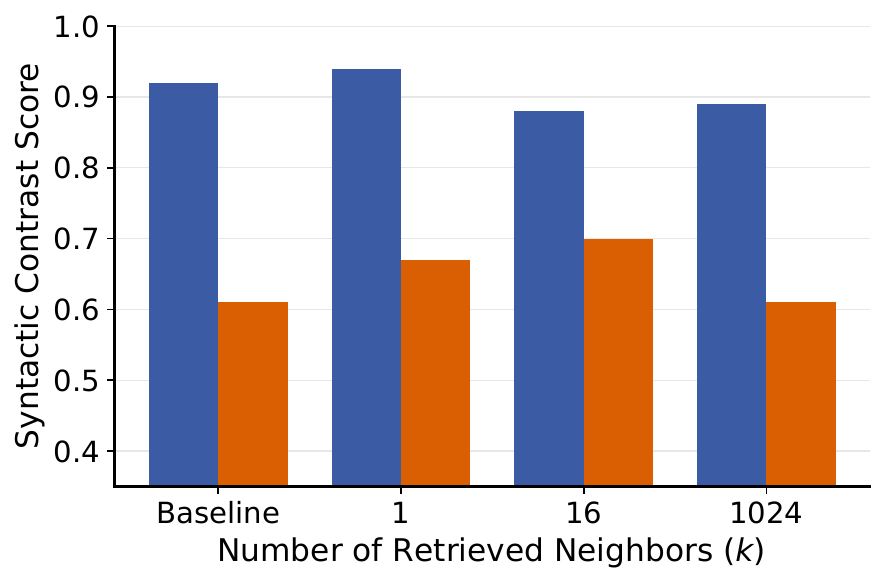}
    \end{minipage}
    \hfill
    \begin{minipage}[t]{0.32\textwidth}
        \centering
        \includegraphics[width=\linewidth]
        {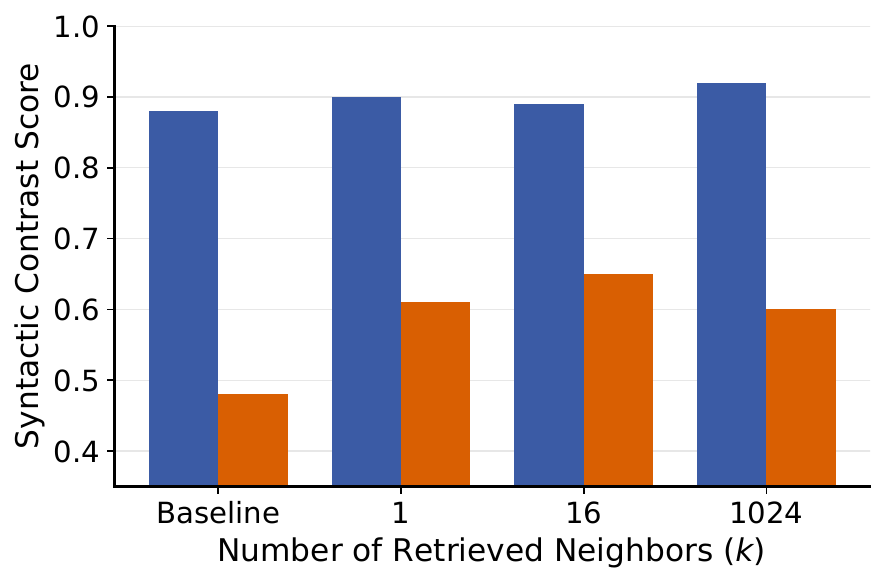}
    \end{minipage}

    \vspace{2mm}

    \begin{minipage}[t]{0.32\textwidth}
        \centering
        \includegraphics[width=\linewidth]
        {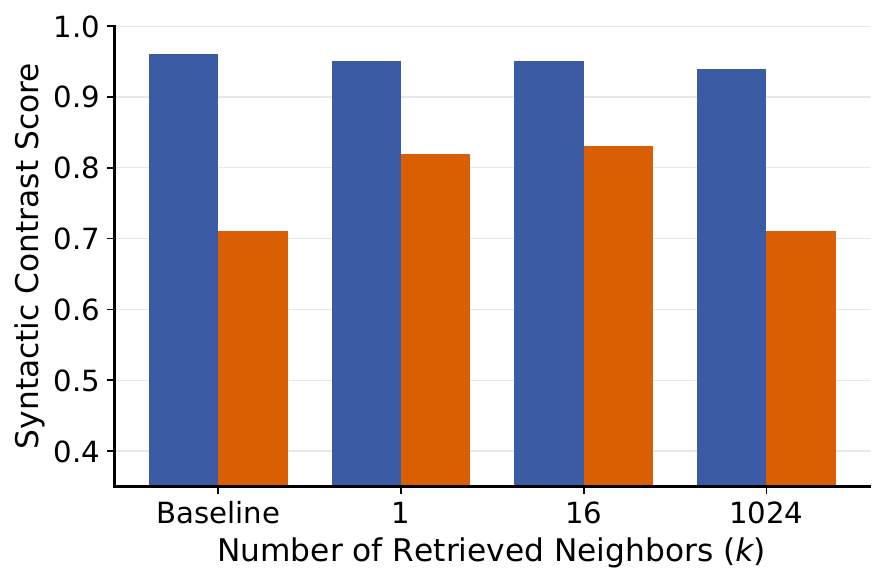}

        (a) Subject--verb agreement
    \end{minipage}
    \hfill
    \begin{minipage}[t]{0.32\textwidth}
        \centering
        \includegraphics[width=\linewidth]
        {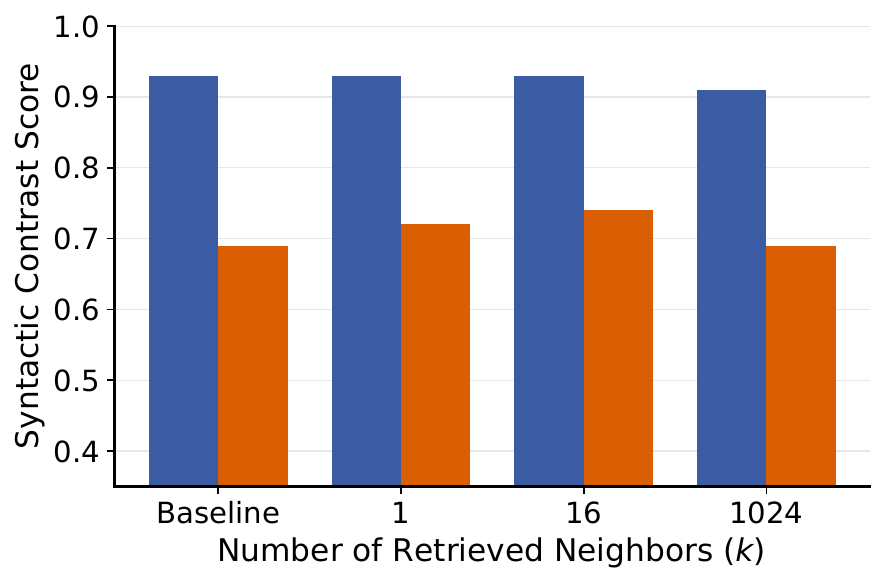}

        (b) Wh-questions
    \end{minipage}
    \hfill
    \begin{minipage}[t]{0.32\textwidth}
        \centering
        \includegraphics[width=\linewidth]
        {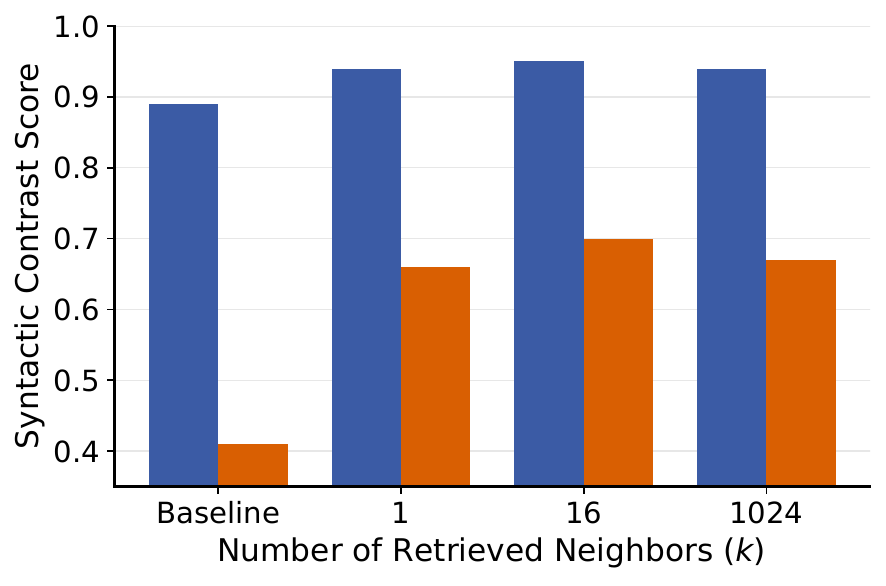}

        (c) Relative clauses
    \end{minipage}

    \caption{
        Performance across numbers of retrieved neighbors
        ($k\in\{1,16,1024\}$) for the child-realistic (top row) and large-scale (bottom row) settings, using $\tau=3$ and sequence-level retrieval. Blue bars indicate high-frequency items and orange bars indicate low-frequency items. For all six low-frequency conditions, $k=16$ outperforms both $k=1$ and $k=1024$. Effects on high-frequency items are smaller and mixed.
    }
    \label{fig:neighbor_effects}
\end{figure*}

\noindent \textbf{Effects of number of retrieved instances.} 
Figure~\ref{fig:neighbor_effects} shows that the number of retrieved instances has a non-monotonic effect on performance in the low-frequency regime. Moving from $k=1$ to a moderate number ($k=16$) consistently improves performance across most conditions, indicating the benefit of aggregating evidence from multiple exemplars. However, further increasing the number of neighbors ($k=1024$) leads to performance degradation, likely due to noise from distant and less relevant neighbors. This pattern holds across models, granularities, and syntactic phenomena, suggesting the optimal choice lies at intermediate values of $k$. The effect for high-frequency items is smaller and mixed. 

\noindent \textbf{No cross-phenomena optimal configuration.} 
Across the three parameters (retrieval granularity, context window size, and number of retrieved instances), no single configuration consistently yields optimal performance across all syntactic phenomena, model, and frequency strata. While sequence-level retrieval, larger context windows, and a moderate number of neighbors ($k=16$) provide the most robust overall performance (and therefore serves as our representative configuration), it does not uniformly dominate across all settings, particularly for simpler or high-frequency dependencies where more localized representations or smaller contexts can be competitive. These results suggest that effective retrieval is jointly modulated by dependency complexity, data frequency, and model capacity, and that optimal retrieval configurations are likely to be phenomenon-dependent rather than universally fixed.

\section{Discussion}

We tested whether an episodic retrieval mechanism as implemented via $k$NN-LMs can help close the lexical frequency gap that language models exhibit in syntactic contrast tests. Our findings reveal that retrieval augmentation provides substantial compensatory benefits for low-frequency items, and that both structural and semantic information contribute to the gains, although the role of structural information is more critical. However, the compensation is insufficient: the frequency gap is narrowed but not fully closed. In the spirit of \citet{sullivan2024machine}, we regard the $k$NN-LM as one instantiation of episodic retrieval that supports a \emph{how-possibly} explanation. Specifically, our model focuses on one functional analogy to CLS, without making more specific implementation commitments. In this section, we discuss which modeling considerations may help narrow the empirical gap further in the future based on the limitations we observe in our analyses. Specifically, our analyses suggest that preferential reweighting of retrieved instances, improved representations and retrieval strategies for structural information, and an adaptive storage and retrieval mechanism that allows for flexible configurations of instances may be promising next steps.

\noindent \textbf{Preferential reweighting.} 
The hippocampus has been proposed to engage in preferential reweighting, where novel or rewarding instances are prioritized during replay and consolidation \citep{kumaran2016learning}. This kind of modulation is consistent with error-driven learning accounts \citep{chang2006dual,rescorla1972theory}, which predict that unexpected or surprising items trigger stronger updates. Standard $k$NN-LM as implemented in this study retrieves instances by similarity alone, without modulating retrieval by the rarity or informativeness of stored instances. This leads to suboptimal retrieval of relevant examples especially for more difficult phenomena like RCs in the child-realistic regime (e.g., the low rate of structurally similar instances retrieved for RCs: Table~\ref{tab:struc_rate}). Thus, improving strategies for retrieval and storage to identify better retrieval targets could close the frequency gap further. Related work on example selection and pruning \citep{toneva2018empirical} and domain-level data reweighting and resampling \citep{xie2023doremi} shows that non-uniform treatment of training data can improve efficiency or performance. Although these studies do not examine retrieval or lexical frequency, they motivate testing whether rarity- or informativeness-aware weighting of retrieved instances benefits low-frequency items.

\noindent \textbf{Better representations and retrieval strategies for structural information.} 
Our analysis shows that structural information is critical for retrieval benefits we observe, which is in line with episodic accounts of syntactic processing that focus on the structural aspects of the retrieved content \citep{borensztajn2011episodic}. In light of this observation, we note two potential issues that may contribute to the persistent frequency gap. First, as discussed above, retrieval based on representation similarity is not always effective in identifying structurally relevant instances, particularly in child-realistic models. Second, even when relevant instances are retrieved, the structural signals encoded in the representations may be too weak to fully bootstrap processing of low-frequency instances. Addressing the first issue may benefit from selective indexing and constraint-based filtering strategies \citep{he-etal-2021-efficient,drozdov-etal-2022-cant}, and the second may require expanding the datastore to maintain the full residual stream representations or exploring alternative pooling strategies in the datastore.

\noindent \textbf{Flexible configurations of storage and retrieval.} 
The absence of a universal retrieval configuration across different syntactic phenomena suggests that maximally effective episodic retrieval requires adaptive mechanisms that modulate granularity, context scope, and neighbor count.
Indeed, human episodic memory has been proposed to support this kind of flexibility \citep{duff2012hippocampus} and adaptive retrieval mechanisms grounded in metacognitive monitoring have been discussed in memory theory \citep{nelson1990metamemory}. Our $k$NN-LM uses a fixed configuration across all inputs. Developing retrieval strategies that adapt their configuration to the structural demands of the input represents a promising direction for closing the remaining frequency gap.

\noindent \textbf{Limitations.}
Our conclusions are limited to controlled English syntactic minimal-pair tests covering three phenomena and two language models; whether the observed pattern generalizes to naturalistic comprehension, production, or human behavior remains open. Moreover, the $k$NN-LM implements only a functional analogy to CLS and does not model biological hippocampal mechanisms, online episodic encoding, or consolidation. Finally, our semantic--structural comparisons do not cleanly isolate the two sources of information, because averaged contextual representations retain structural information and structural-match rates may reflect datastore coverage as well as representation and retrieval quality. Our results therefore motivate episodic retrieval as a possible mechanism rather than pinpoint the mechanism underlying human robustness to lexical frequency.

\section{Conclusion}

We tested whether an episodic retrieval mechanism can help close the lexical frequency gap that language models exhibit in syntactic contrast evaluations. Using $k$NN-LMs as a functional proxy for hippocampal episodic memory, we find retrieval consistently narrows the performance gap between high- and low-frequency items across syntactic phenomena and model/data scales, with structural information playing a critical role in driving this compensation. However, the frequency gap remains not fully closed---test items with low frequency lexical items still systematically lead to lower performance even after retrieval augmentation. The limitations of the current implementation of episodic memory identified through our analyses lead to three concrete future directions for improvement: implementation of a preferential reweighting mechanism, stronger structural representations and retrieval strategies, and adaptive retrieval configurations rather than fixed. More broadly, these results offer proof-of-concept support for the CLS hypothesis that episodic memory contributes to the robustness of syntactic processing to lexical frequency effects, while identifying areas for improvement for future modeling work.

\section*{Acknowledgments}

This work has received funding from the European Union's Horizon 2020 research and innovation programme under the Marie Skłodowska-Curie grant agreement No 945304 – Cofund AI4theSciences hosted by PSL University. We thank the three anonymous reviewers for their feedback.

\bibliography{colm2026_conference}
\bibliographystyle{colm2026_conference}

\appendix

\newpage

\section{Appendix}

\subsection{Pretrained models and datasets}
\label{app:model_dataset}

\noindent \textbf{Child-realistic pretraining data.} We constructed a dataset to approximate the language input a child receives during early development, drawing on child-directed speech (CDS) from CHILDES \citep{macwhinney2000childes}, transcribed conversational speech from the Switchboard Dialog Act Corpus \citep{godfrey1992switchboard,stolcke2000dialogue}, children's stories from the Children's Stories Text Corpus \citep{bensaid2021cstc,bensaid2021fairytailor}, and movie and television subtitles from OpenSubtitles \citep{lison2016opensubtitles}. These four sources also appear in the 2023 BabyLM corpus release
\citep{warstadt2023findings}. The complete dataset comprises 50.03M words, consistent with estimates of cumulative linguistic input by early childhood \citep{gilkerson2017mapping,frank2023bridging}. Table~\ref{tab:training_data} shows the distribution of constituent sub-datasets.

\begin{table}[ht]
\centering
\small
\vspace{8pt}
\caption{Composition of child-realistic pretraining corpus combining child-directed
speech, conversational dialogue, and children's books to approximate naturalistic
language exposure during early childhood. CDS = child-directed speech; SWBD = Switchboard; CSTC = Children's Stories Text Corpus; OSub = OpenSubtitles.}
\label{tab:training_data}
\setlength{\tabcolsep}{4pt}
\renewcommand{\arraystretch}{1.1}
\begin{tabular}{lccc}
\toprule
\textbf{Dataset} & \textbf{Domain} & \textbf{\# words} & \textbf{Prop. (\%)} \\
\midrule
CHILDES & CDS      & 15M    & 30.0 \\
SWBD    & Dialogue & 1.18M  & 2.4  \\
CSTC    & Books    & 3.22M  & 6.4  \\
OSub    & Dialogue & 30.63M & 61.2 \\
\midrule
\textbf{Total} & & \textbf{50.03M} & \textbf{100.0} \\
\bottomrule
\end{tabular}
\end{table}

\noindent \textbf{Large-scale pretraining data.} For the large-scale condition, we use GPT-2 XL (1.5B parameters) pretrained on the WebText corpus \citep{radford2019language}, which contains approximately 40GB of text scraped from internet sources.

\noindent \textbf{Model architectures and training.} Table ~\ref{tab:model_architecture} summarizes the architectural specifications for both models. The child-realistic model uses the GPT-2 Small architecture with 124M parameters, trained from scratch for 10 epochs, following the BabyLM Challenge setup \citep{hu2024findings}. The GPT-2 XL model uses the publicly available pretrained checkpoint from \citet{radford2019language}. The training details for the child-realistic model are provided in Table~\ref{tab:training_hyperparameters}.

\begin{table}[ht]
\centering
\small
\vspace{8pt}
\caption{Model architecture specifications. GPT-2 Small was trained from scratch on child-realistic data (50M words); GPT-2 XL uses the publicly released pretrained checkpoint from \citet{radford2019language} trained on WebText ($\sim$40GB).}
\label{tab:model_architecture}
\setlength{\tabcolsep}{4pt}
\renewcommand{\arraystretch}{1.1}
\begin{tabular}{@{}lcc@{}}
\toprule
\textbf{Parameter} & \textbf{GPT-2 Small} & \textbf{GPT-2 XL} \\
\midrule
\# parameters       & 124M  & 1.5B  \\
Transformer layers  & 12    & 48    \\
Hidden size         & 768   & 1600  \\
FFN size            & 3072  & 6400  \\
Attention heads     & 12    & 25    \\
Attention dropout   & 0.1  & 0.1   \\
Embedding dimension & 768   & 1600  \\
Vocabulary size     & 50257 & 50257 \\
\bottomrule
\end{tabular}
\end{table}

\begin{table}[ht]
\centering
\small
\vspace{8pt}
\caption{Training hyperparameters for GPT-2 Small on the child-realistic corpus, trained for 10 epochs using a standard autoregressive language modeling objective with cross-entropy loss.}
\label{tab:training_hyperparameters}
\setlength{\tabcolsep}{4pt}
\renewcommand{\arraystretch}{1.1}
\begin{tabular}{@{}ll@{}}
\toprule
\textbf{Hyperparameter} & \textbf{Value} \\
\midrule
Max sequence length & 1024         \\
Batch size          & 64           \\
Training epochs     & 10           \\
Learning rate       & 0.0001       \\
LR scheduler        & inverse sqrt \\
Warmup steps        & 1000         \\
Optimizer           & Adam         \\
Adam-$\beta_1$      & 0.9          \\
Adam-$\beta_2$      & 0.98         \\
Dropout             & 0.1          \\
Weight decay        & 0.01         \\
Gradient clipping   & 1.0          \\
\bottomrule
\end{tabular}
\end{table}

\subsection{Evaluation set construction}
\label{app:subset_selection}

To ensure focused evaluation on syntactic phenomena relevant to our research questions, we selected subsets corresponding to our target phenomena from three evaluation benchmarks: BLiMP \citep{warstadt2020blimp}, Zorro \citep{huebner2021babyberta}, and BIG-bench \citep{srivastava2023beyond} as discussed below.

\subsubsection{Source benchmarks}

\textbf{BLiMP.} The Benchmark of Linguistic Minimal Pairs for English \citep{warstadt2020blimp} consists of 67 sub-datasets, each containing 1,000 minimal pairs (67,000 pairs total) isolating specific contrasts in syntax, morphology, or semantics. The data is automatically generated according to linguist-crafted grammar templates, with 96.4\% aggregate human agreement with the labels. For subject--verb agreement, we selected the relational-noun distractor paradigm and four plural-agreement paradigms: regular and irregular plurals, each with a verb-number manipulation holding the subject fixed (variant 1) and a subject-number manipulation holding the verb fixed (variant 2). For wh-questions, we selected the object-gap, subject-gap, and long-distance subject-gap wh-question paradigms, together with the four wh-versus-that paradigms crossing gap presence and dependency length. For relative clauses, we selected the relative-clause distractor agreement paradigm.

\textbf{Zorro.} Zorro \citep{huebner2021babyberta} is a targeted grammar test suite designed for small-data, restricted-vocabulary evaluation. It comprises 23 paradigms spanning 13 grammatical phenomena, with 2,000 minimal pairs (4,000 sentences) per paradigm. Its lexical items were selected from whole-word entries in BabyBERTa's custom vocabulary, making the suite suitable for evaluating models trained on comparatively limited, child-directed linguistic input. For subject--verb agreement, we selected agreement across a prepositional phrase, agreement in questions with an auxiliary, and agreement in simple questions. For wh-questions, we selected the object and subject wh-question paradigms. For relative clauses, we selected agreement across a relative clause.

\textbf{BIG-bench.} The Beyond the Imitation Game Benchmark \citep{srivastava2023beyond} includes several linguistic evaluation tasks. For subject-verb agreement, we selected the simple-English, single-adverb, double-adverb, adverb-conjunction, name-plus-prepositional-phrase, noun-plus-prepositional-phrase, noun-plus-prepositional-phrase-and-adverb, Gulordava-nonce, and frequency-disbalance-nonce subtasks. For relative clauses, we selected the long-nested-inner and long-nested-outer subtasks. No BIG-bench items were used for the wh-question evaluation.

\subsection{Statistical analyses}
\label{app:stats}

\noindent \textbf{Bootstrapped confidence intervals on per-frequency band deltas.}
For each frequency band, we computed the performance delta (kNN $-$ baseline) and estimated $95\%$ confidence intervals via bootstrap resampling over test items ($10{,}000$ resamples). As seen in Table~\ref{tab:boot_ci}, in both settings, the intervals for low- and high-frequency items do not overlap, indicating a robust separation in retrieval gains.

\begin{table}[ht]
\centering
\small
\caption{Bootstrapped $95\%$ confidence intervals on the per-frequency band accuracy delta (kNN $-$ baseline), $10{,}000$ resamples over test items ($k=16$, $\tau=3$, sequence-level).}
\label{tab:boot_ci}
\begin{tabular}{llcc}
\toprule
Pretraining & Freq & $\Delta$ & $95\%$ CI \\
\midrule
\multirow{3}{*}{Child-realistic}
& high & \textit{$-0.013$} & \textit{$[-0.030,\ 0.004]$} \\
& mid  & \textit{0.040}    & \textit{$[0.022,\ 0.059]$} \\
& low  & \textit{0.140}    & \textit{$[0.115,\ 0.166]$} \\
\multirow{3}{*}{Large-scale}
& high & \textit{0.017}    & \textit{$[0.002,\ 0.033]$} \\
& mid  & \textit{0.023}    & \textit{$[0.008,\ 0.039]$} \\
& low  & \textit{0.153}    & \textit{$[0.125,\ 0.182]$} \\
\bottomrule
\end{tabular}
\end{table}

\noindent \textbf{Mixed-effects logistic regression.}
To test whether the effect of retrieval is larger for low-frequency items, we fit a mixed-effects logistic regression predicting correctness from Frequency (high/low) and Model Type (baseline/kNN), with random intercepts for item and for phenomenon. We used treatment coding, with high-frequency items and the baseline model as the reference levels:
\begin{center}
\texttt{correct $\sim$ Frequency * ModelType + (1$\mid$item) + (1$\mid$phenomenon)}
\end{center}

Coefficients are reported in Table~\ref{tab:mixed_effects}. Under this coding, the Model Type coefficient estimates the effect of kNN augmentation within the high-frequency band. This coefficient was not statistically significant ($\beta=-0.04$, $p=0.62$), indicating no detectable retrieval effect for high-frequency items. Crucially, the Frequency $\times$ Model Type interaction was positive and significant ($\beta=0.52$, $p<0.001$), indicating that the retrieval effect (in log odds) was larger for low-frequency than for high-frequency items. Together with the positive accuracy differences and bootstrap confidence intervals for low-frequency items in Table~\ref{tab:boot_ci}, this result supports the hypothesis that retrieval disproportionately benefits items for which parametric representations are weak, rather than providing a uniform improvement across frequency bands.

\begin{table}[ht]
\centering
\small
\caption{Mixed-effects logistic regression coefficients
(\texttt{correct $\sim$ Frequency * ModelType + (1$\mid$item) + (1$\mid$phenomenon)}; $k=16$, $\tau=3$, sequence-level).}
\label{tab:mixed_effects}
\begin{tabular}{lcccc}
\toprule
Term & $\beta$ & SE & $z$ & $p$ \\
\midrule
Frequency (low)                    & \textit{$-1.52$} & \textit{0.11} & \textit{$-13.8$} & \textit{$<0.001$} \\
ModelType (kNN)                    & \textit{$-0.04$} & \textit{0.08} & \textit{$-0.50$}  & \textit{0.62} \\
Frequency $\times$ ModelType       & \textit{0.52}    & \textit{0.10} & \textit{5.20}    & \textit{$<0.001$} \\
\bottomrule
\end{tabular}
\end{table}

\subsection{Structural match analysis}
\label{app:struc_simi}

For each evaluation item and retrieved sequence, we defined a binary
structural match indicator equal to 1 if the sequence contained the same target phenomenon---subject-verb agreement (SV), a wh-question (Wh), or a relative clause (RC)---and 0 otherwise. We identified these phenomena using spaCy \citep{honnibal2020spacy} as specified below:

\textbf{Subject-Verb detection.} We applied spaCy's dependency parser to identify subject-verb relationships. Instances were flagged if they contained \texttt{nsubj} (nominal subject) or \texttt{nsubjpass} (passive nominal subject) dependencies that link a noun phrase to a finite verb.

\textbf{Wh-Sentence detection.} We used part-of-speech (POS) tagging to identify wh-words (POS tags: \texttt{WDT}, \texttt{WP}, \texttt{WP\$}, \texttt{WRB}) at clause boundaries. Instances containing interrogative or relative wh-elements were marked for inclusion.

\textbf{Relative Clause detection.} We combined POS tagging with dependency parsing to identify relative clause structures. Specifically, we detected: (1) Relative pronouns (\texttt{WDT}, \texttt{WP}) or relativizers (\emph{that}, \emph{which}, \emph{who}); (2) Clausal complements with \texttt{relcl} (relative clause modifier) dependency relations; and (3) Appropriate structural embedding within a noun phrase.

Table~\ref{tab:struc_rate} reports the mean proportion of the 16 retrieved neighbors that contain the target syntactic structure, and Table~\ref{tab:struc_ttests} reports two-sample $t$-tests comparing the effect of retrieval with vs. without structural matches, each against the baseline and against each other.

\begin{table}[ht]
\centering
\small
\caption{Mean proportion of the 16 retrieved neighbors containing the
target syntactic structure ($k=16$, $\tau=3$, sequence-level retrieval). Structural matches remain particularly sparse for relative clauses in the child-realistic setting.}
\label{tab:struc_rate}
\begin{tabular}{lcccccc}
\toprule
& \multicolumn{2}{c}{\textbf{SV}}
& \multicolumn{2}{c}{\textbf{Wh}}
& \multicolumn{2}{c}{\textbf{RC}} \\
\cmidrule(lr){2-3}
\cmidrule(lr){4-5}
\cmidrule(lr){6-7}
& high & low & high & low & high & low \\
\midrule
Child-realistic & 0.74 & 0.70 & 0.49 & 0.62 & 0.09 & 0.15 \\
Large-scale     & 0.95 & 0.89 & 0.67 & 0.76 & 0.74 & 0.70 \\
\bottomrule
\end{tabular}
\end{table}

\begin{table*}[ht]
\centering
\small
\caption{$p$-values for structural-retrieval comparisons in the
child-realistic and large-scale models ($k=16$, $\tau=3$,
sequence-level retrieval).}
\label{tab:struc_ttests}
\begin{tabular}{llcccccc}
\toprule
& & \multicolumn{3}{c}{\textbf{Child-realistic}}
& \multicolumn{3}{c}{\textbf{Large-scale}} \\
\cmidrule(lr){3-5}
\cmidrule(lr){6-8}
Comparison & Freq & SV & Wh & RC & SV & Wh & RC \\
\midrule

\multirow{2}{*}{Baseline vs.\ Without}
& low  & 0.16 & 0.38 & 0.022
       & 0.071 & 0.49  & $<0.001$ \\
& high & 0.67 & 0.14 & 0.43
       & 0.76  & 0.008 & 0.28 \\

\multirow{2}{*}{Baseline vs.\ With}
& low  & 0.006 & 0.021 & $<0.001$
       & $<0.001$ & 0.004 & $<0.001$ \\
& high & 0.36 & 0.23 & 0.016
       & 0.44 & 0.52 & $<0.001$ \\

\multirow{2}{*}{With vs.\ Without}
& low  & 0.003 & 0.046 & $<0.001$
       & $<0.001$ & 0.031 & $<0.001$ \\
& high & 0.31 & 0.078 & 0.012
       & 0.44 & 0.001 & 0.001 \\

\bottomrule
\end{tabular}
\end{table*}

\noindent The tests support the claim that retrieval benefits are greater when the retrieved examples contain structurally matched examples, and reveal additional nuance: the structural-match advantage is more pronounced in low-frequency bands and for structurally complex phenomena (RCs), consistent with broader trends elsewhere in the manuscript.

\subsection{Additional Retrieval Configuration Results}
\label{app:retrieval_config}

\begin{figure*}[tp]
\centering
\begin{minipage}[t]{0.32\textwidth}
    \centering
    \includegraphics[width=\linewidth]{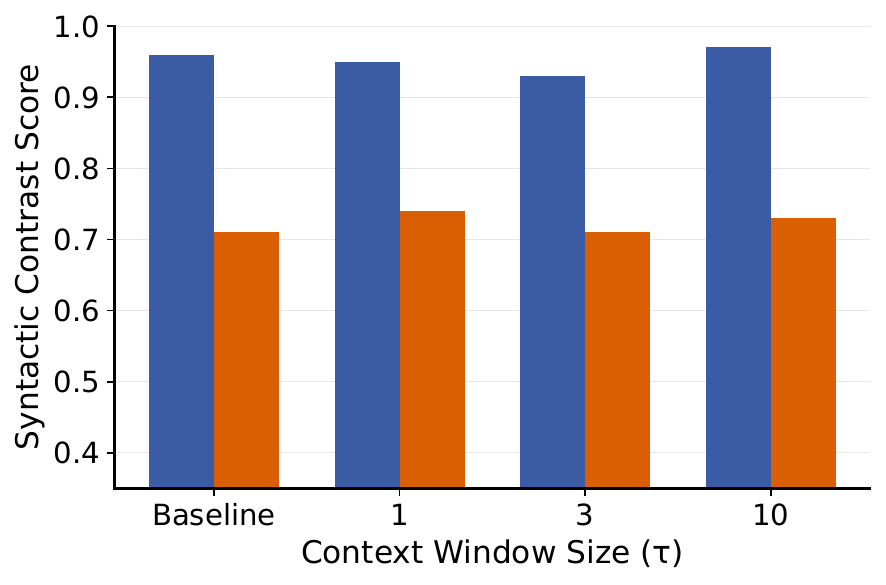}
    (a) SV -- token
\end{minipage}\hfill
\begin{minipage}[t]{0.32\textwidth}
    \centering
    \includegraphics[width=\linewidth]{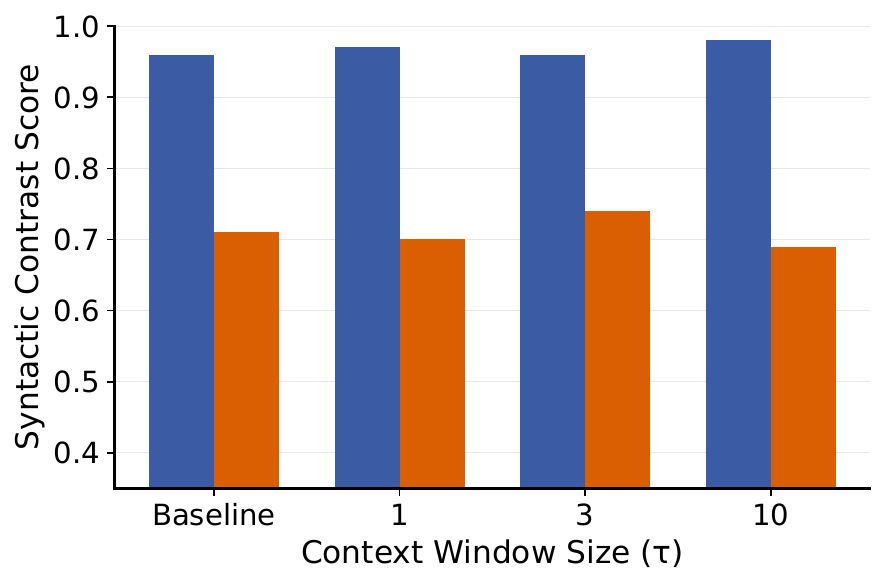}
    (b) SV -- phrase
\end{minipage}\hfill
\begin{minipage}[t]{0.32\textwidth}
    \centering
    \includegraphics[width=\linewidth]{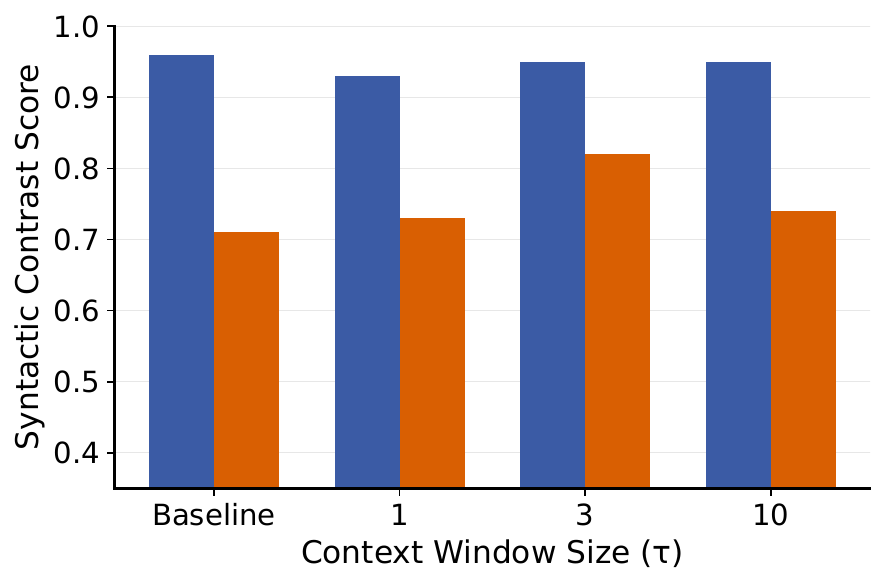}
    (c) SV -- sequence
\end{minipage}
\vspace{2mm}
\begin{minipage}[t]{0.32\textwidth}
    \centering
    \includegraphics[width=\linewidth]{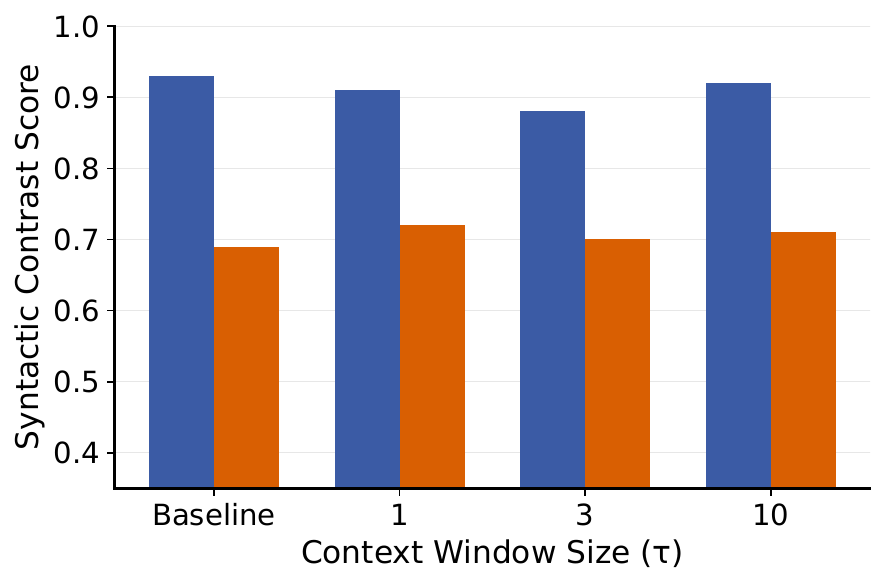}
    (d) Wh -- token
\end{minipage}\hfill
\begin{minipage}[t]{0.32\textwidth}
    \centering
    \includegraphics[width=\linewidth]{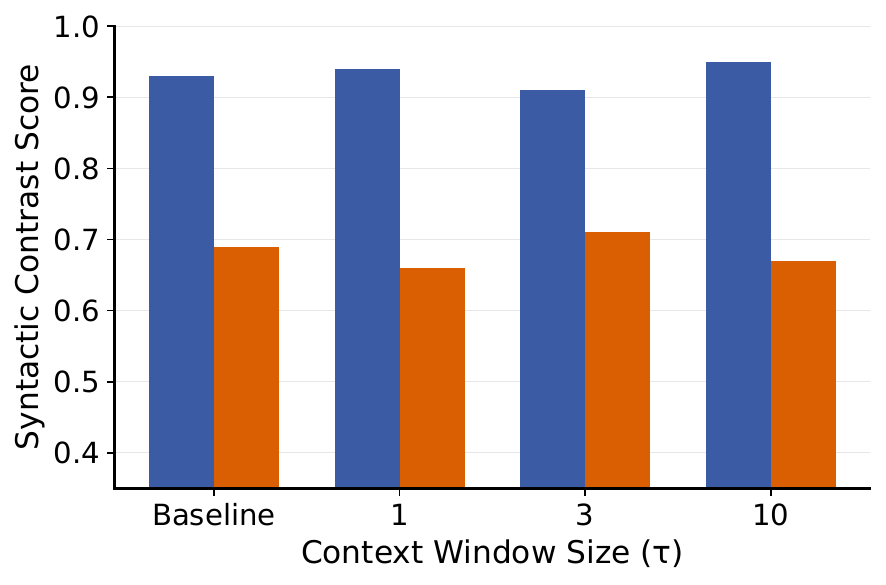}
    (e) Wh -- phrase
\end{minipage}\hfill
\begin{minipage}[t]{0.32\textwidth}
    \centering
    \includegraphics[width=\linewidth]{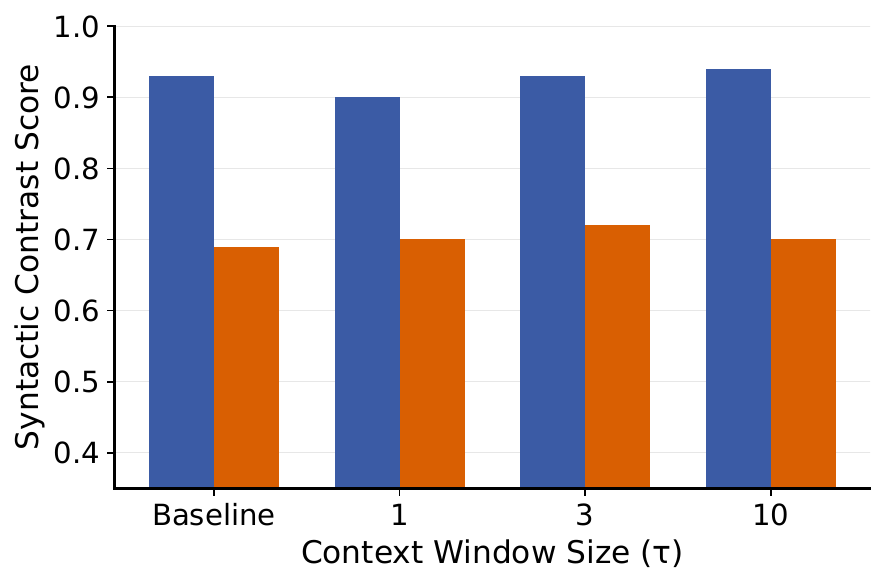}
    (f) Wh -- sequence
\end{minipage}
\vspace{2mm}
\begin{minipage}[t]{0.32\textwidth}
    \centering
    \includegraphics[width=\linewidth]{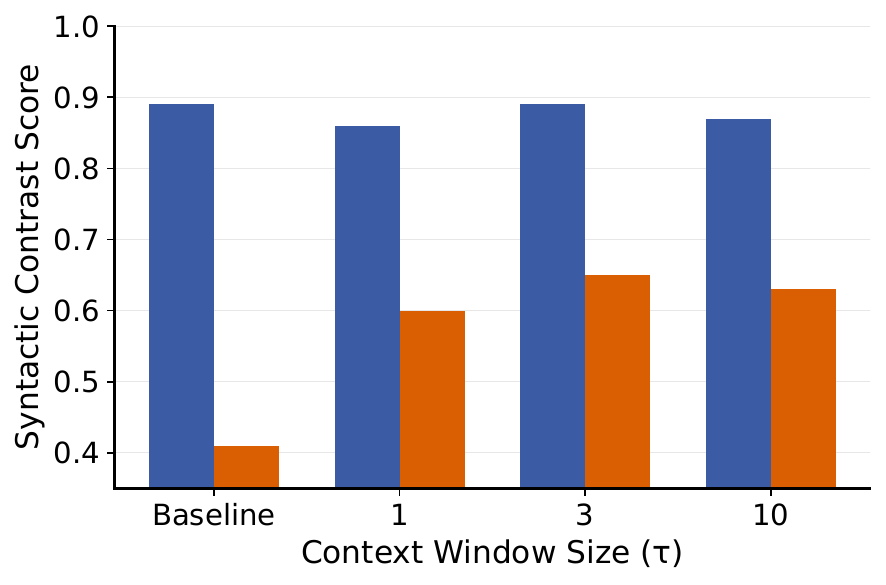}
    (g) RC -- token
\end{minipage}\hfill
\begin{minipage}[t]{0.32\textwidth}
    \centering
    \includegraphics[width=\linewidth]{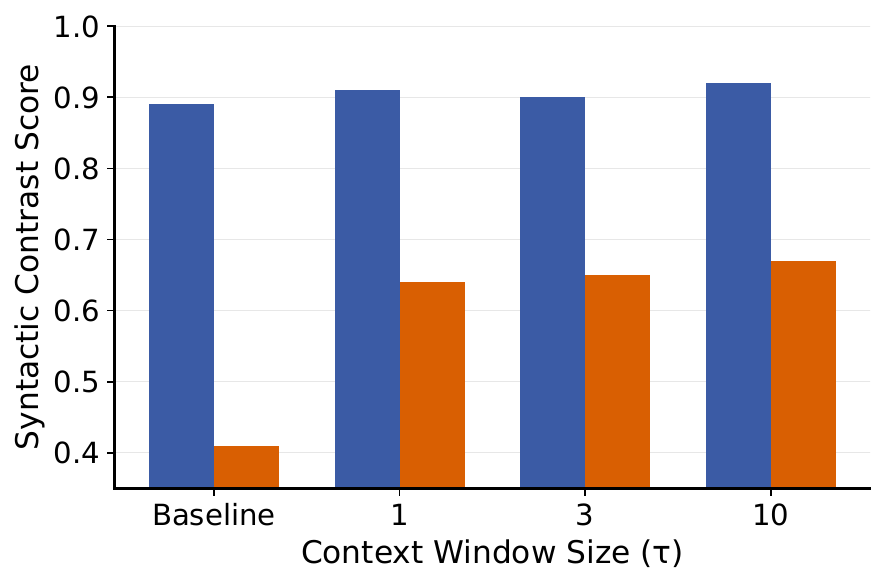}
    (h) RC -- phrase
\end{minipage}\hfill
\begin{minipage}[t]{0.32\textwidth}
    \centering
    \includegraphics[width=\linewidth]{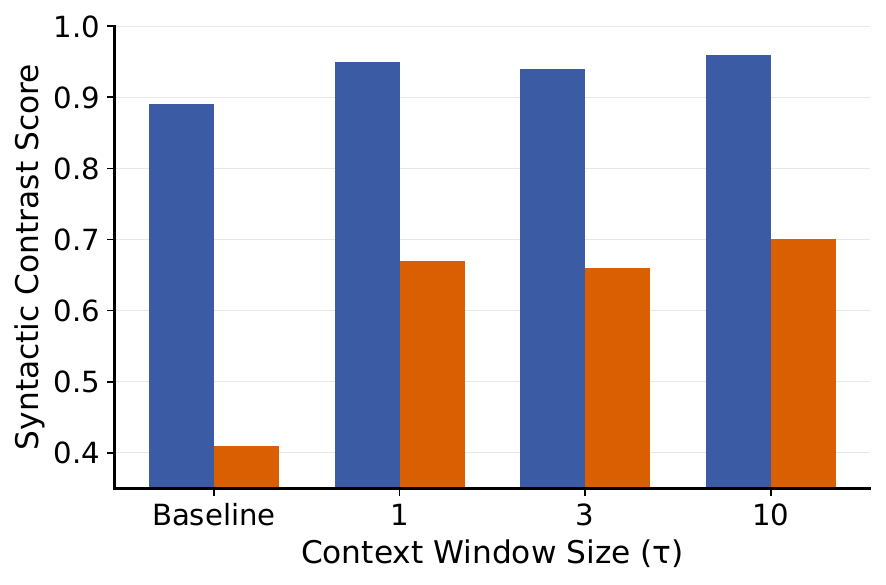}
    (i) RC -- sequence
\end{minipage}
\caption{Effect of context window size ($\tau \in \{1,3,10\}$) for GPT-2 XL at $k=1$. Blue bars indicate high-frequency items and orange bars indicate low-frequency items; baseline results are shown for reference.}
\label{fig:app_context_gpt2xl_k1}
\end{figure*}

\begin{figure*}[tp]
\centering
\begin{minipage}[t]{0.32\textwidth}
    \centering
    \includegraphics[width=\linewidth]{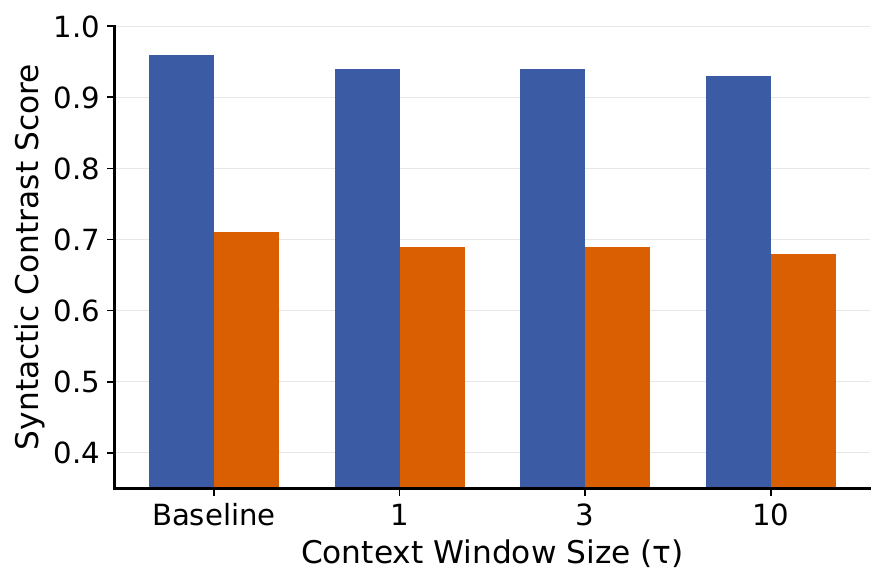}
    (a) SV -- token
\end{minipage}\hfill
\begin{minipage}[t]{0.32\textwidth}
    \centering
    \includegraphics[width=\linewidth]{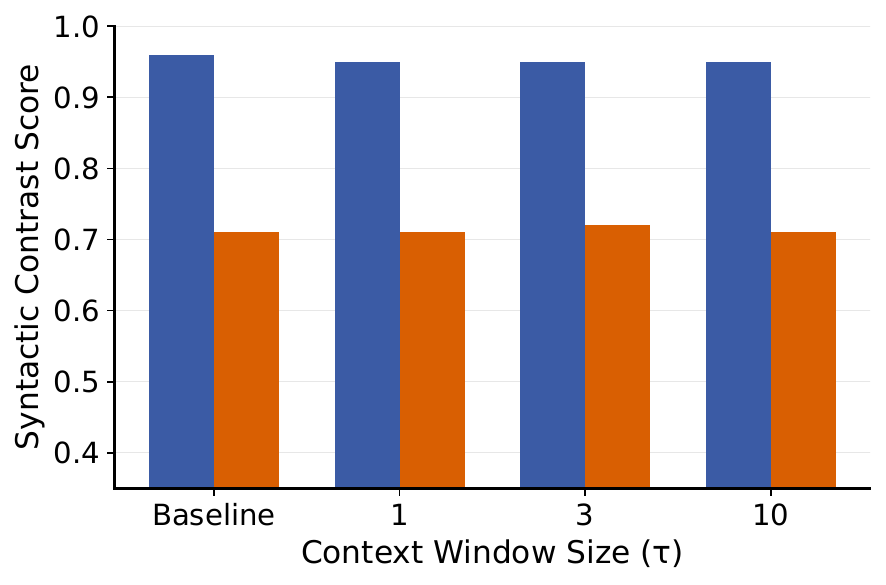}
    (b) SV -- phrase
\end{minipage}\hfill
\begin{minipage}[t]{0.32\textwidth}
    \centering
    \includegraphics[width=\linewidth]{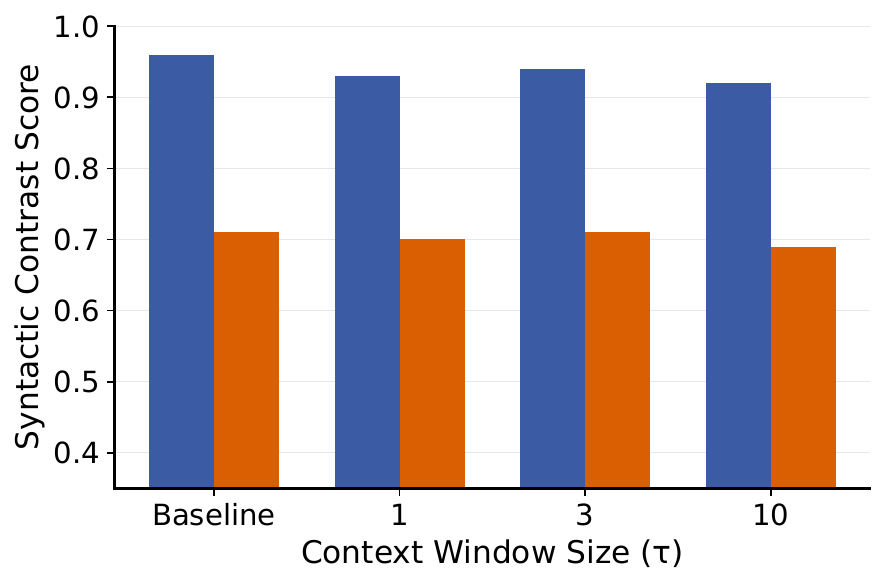}
    (c) SV -- sequence
\end{minipage}
\vspace{2mm}
\begin{minipage}[t]{0.32\textwidth}
    \centering
    \includegraphics[width=\linewidth]{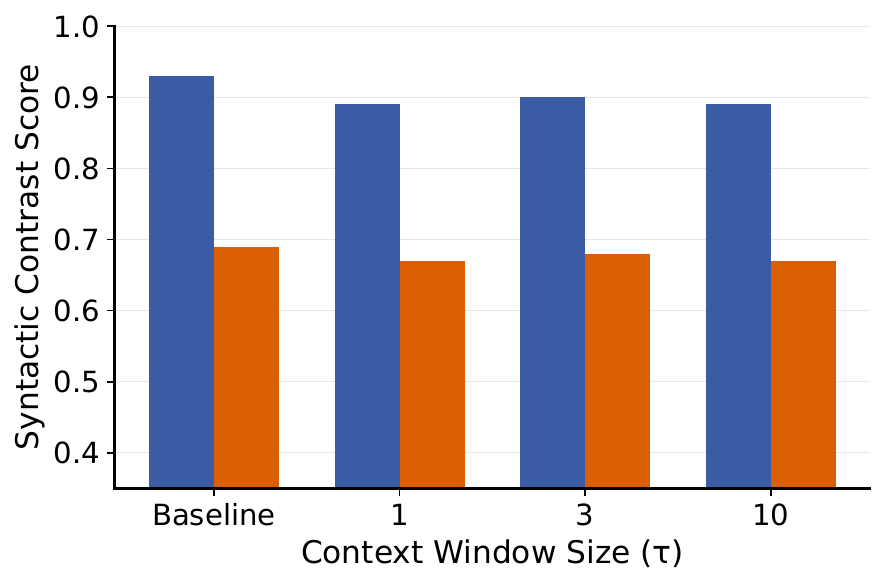}
    (d) Wh -- token
\end{minipage}\hfill
\begin{minipage}[t]{0.32\textwidth}
    \centering
    \includegraphics[width=\linewidth]{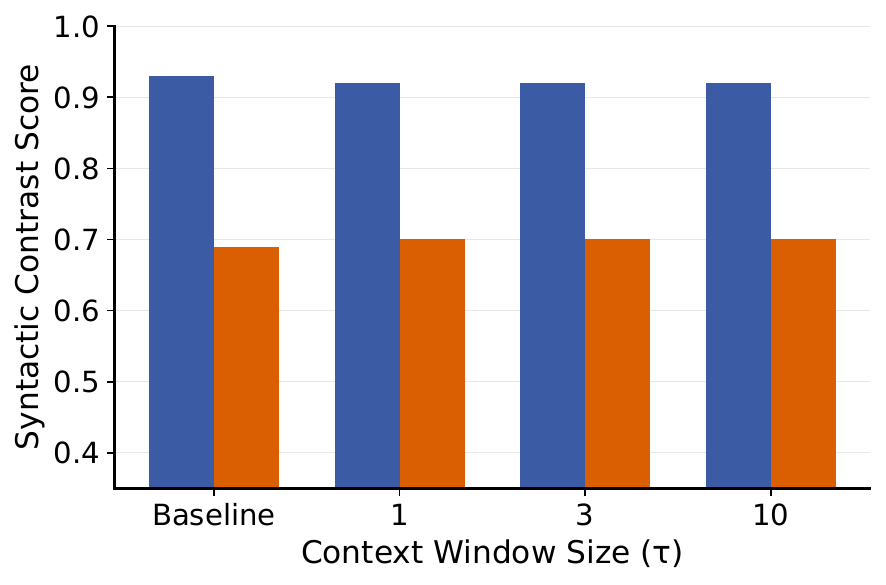}
    (e) Wh -- phrase
\end{minipage}\hfill
\begin{minipage}[t]{0.32\textwidth}
    \centering
    \includegraphics[width=\linewidth]{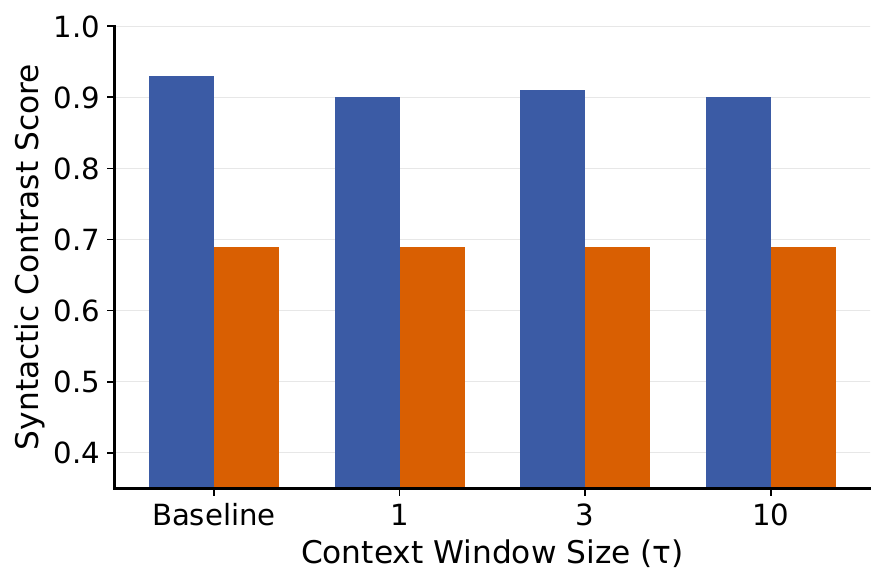}
    (f) Wh -- sequence
\end{minipage}
\vspace{2mm}
\begin{minipage}[t]{0.32\textwidth}
    \centering
    \includegraphics[width=\linewidth]{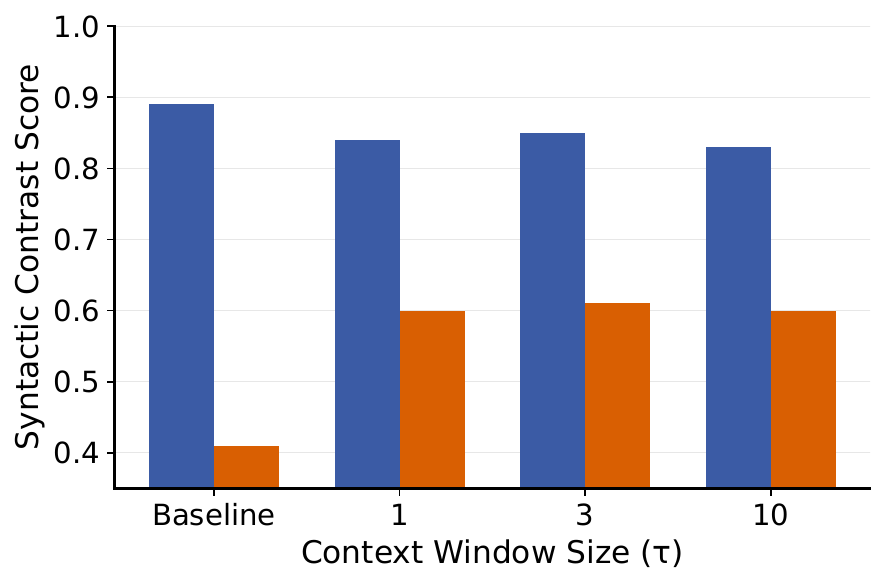}
    (g) RC -- token
\end{minipage}\hfill
\begin{minipage}[t]{0.32\textwidth}
    \centering
    \includegraphics[width=\linewidth]{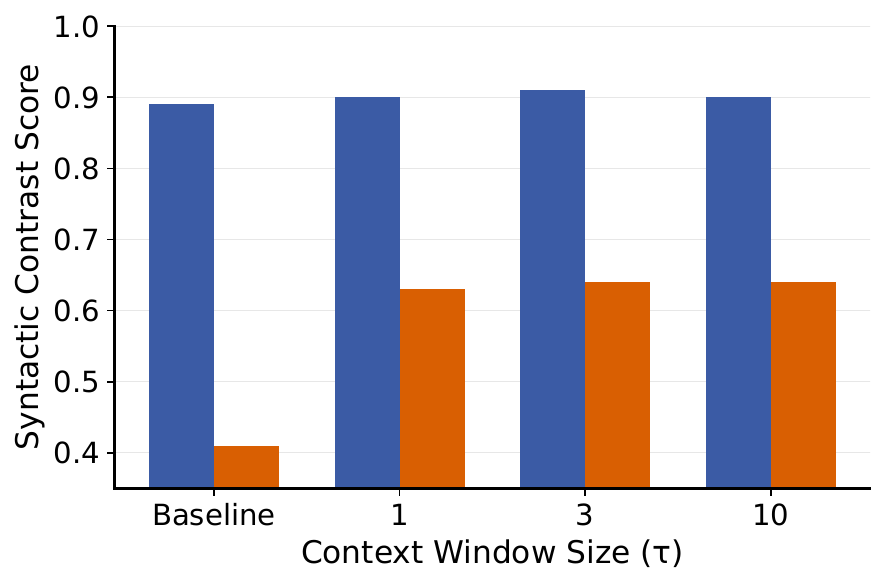}
    (h) RC -- phrase
\end{minipage}\hfill
\begin{minipage}[t]{0.32\textwidth}
    \centering
    \includegraphics[width=\linewidth]{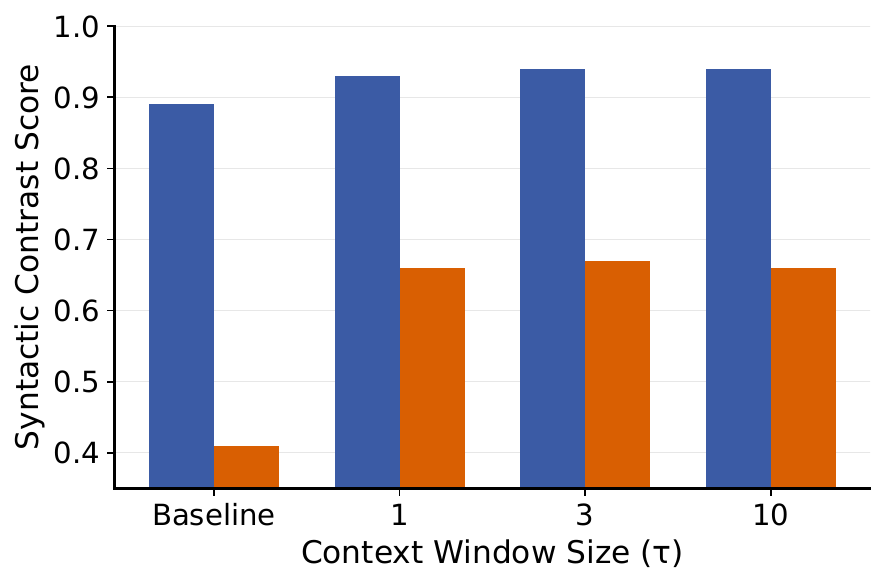}
    (i) RC -- sequence
\end{minipage}
\caption{Effect of context window size ($\tau \in \{1,3,10\}$) for GPT-2 XL at $k=1024$. Blue bars indicate high-frequency items and orange bars indicate low-frequency items; baseline results are shown for reference.}
\label{fig:app_context_gpt2xl_k1024}
\end{figure*}

\begin{figure*}[tp]
\centering
\begin{minipage}[t]{0.32\textwidth}
    \centering
    \includegraphics[width=\linewidth]{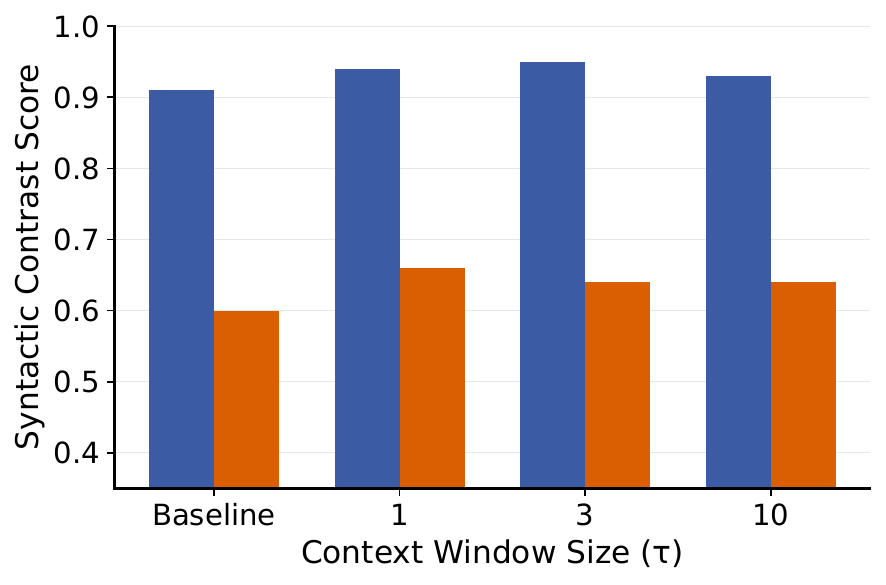}
    (a) SV -- token
\end{minipage}\hfill
\begin{minipage}[t]{0.32\textwidth}
    \centering
    \includegraphics[width=\linewidth]{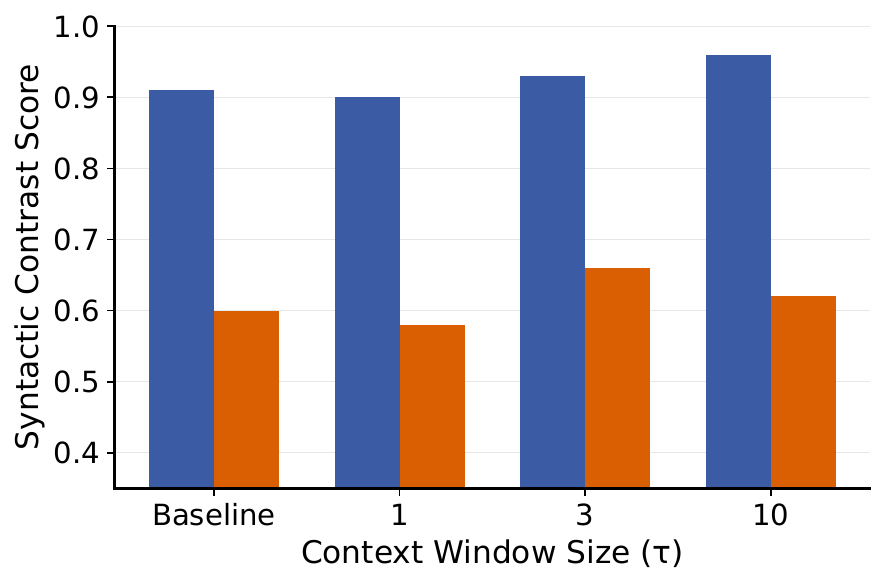}
    (b) SV -- phrase
\end{minipage}\hfill
\begin{minipage}[t]{0.32\textwidth}
    \centering
    \includegraphics[width=\linewidth]{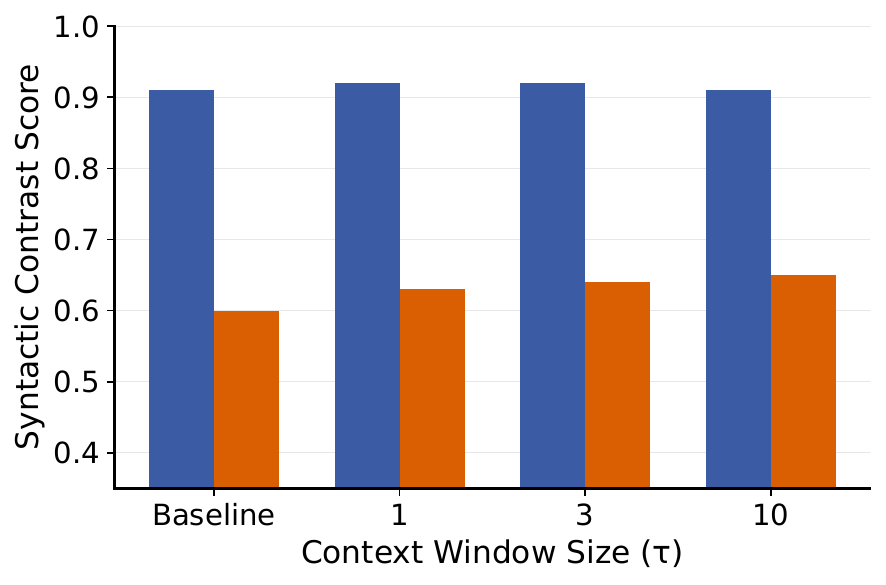}
    (c) SV -- sequence
\end{minipage}
\vspace{2mm}
\begin{minipage}[t]{0.32\textwidth}
    \centering
    \includegraphics[width=\linewidth]{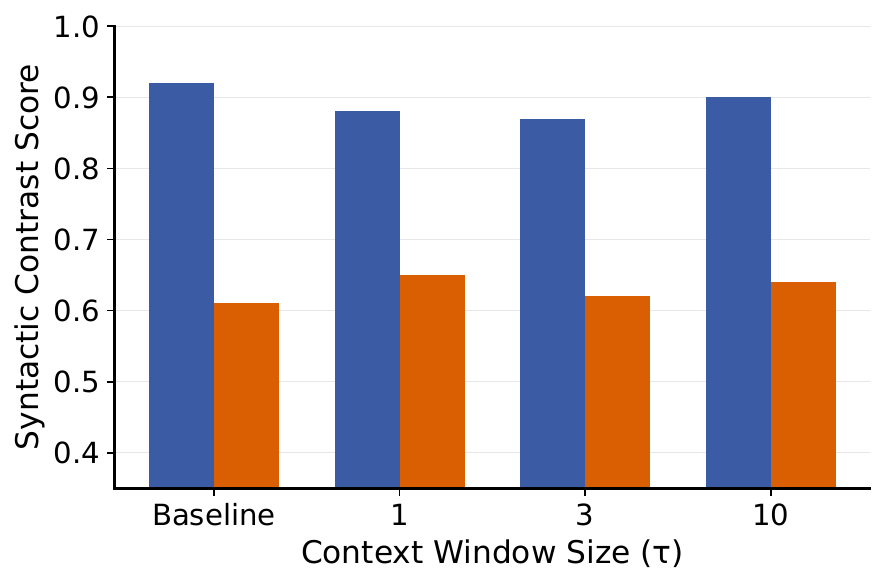}
    (d) Wh -- token
\end{minipage}\hfill
\begin{minipage}[t]{0.32\textwidth}
    \centering
    \includegraphics[width=\linewidth]{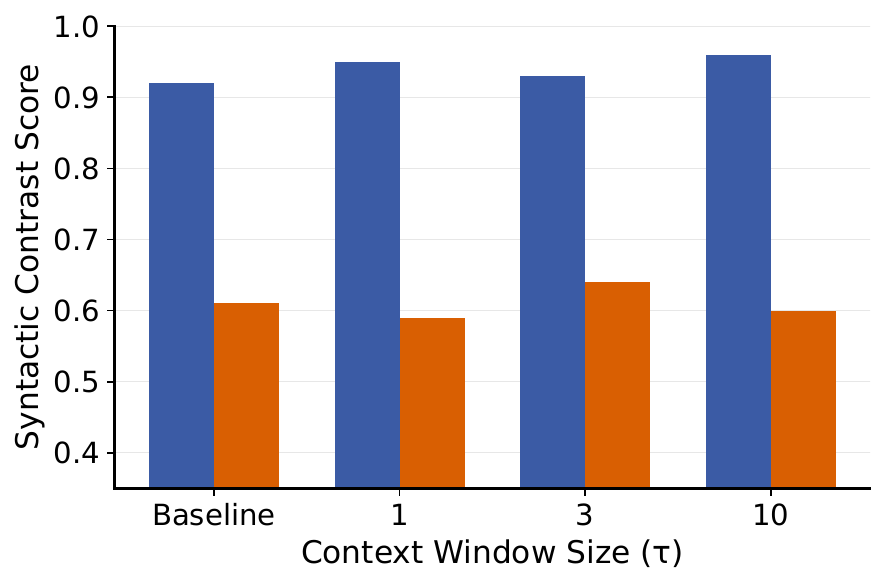}
    (e) Wh -- phrase
\end{minipage}\hfill
\begin{minipage}[t]{0.32\textwidth}
    \centering
    \includegraphics[width=\linewidth]{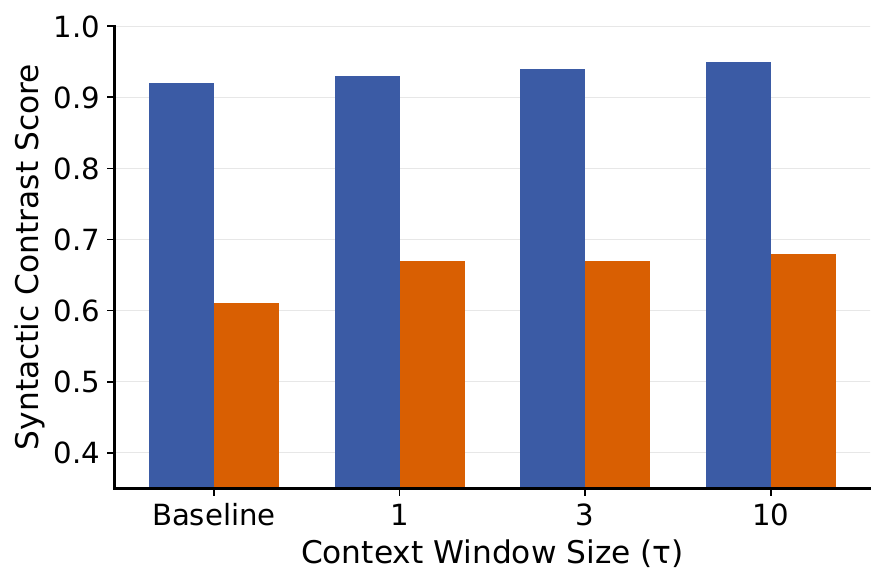}
    (f) Wh -- sequence
\end{minipage}
\vspace{2mm}
\begin{minipage}[t]{0.32\textwidth}
    \centering
    \includegraphics[width=\linewidth]{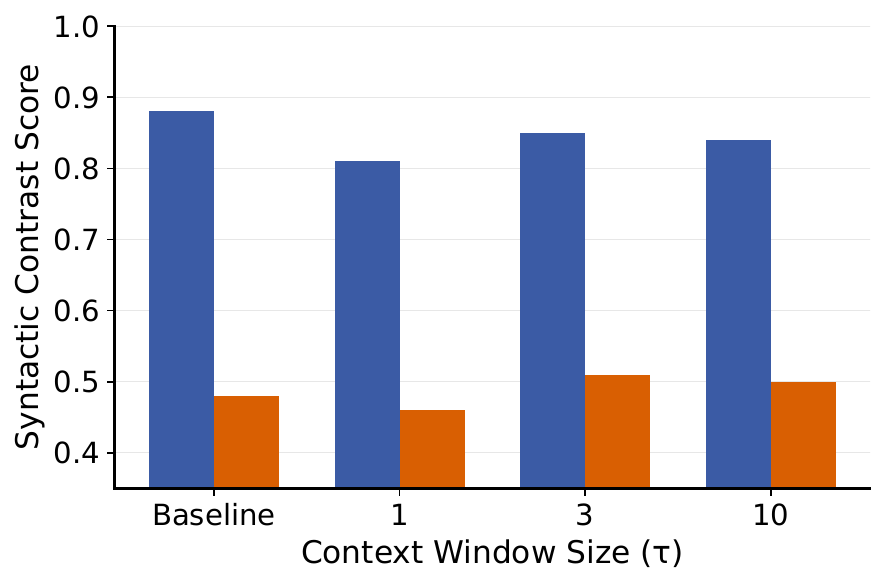}
    (g) RC -- token
\end{minipage}\hfill
\begin{minipage}[t]{0.32\textwidth}
    \centering
    \includegraphics[width=\linewidth]{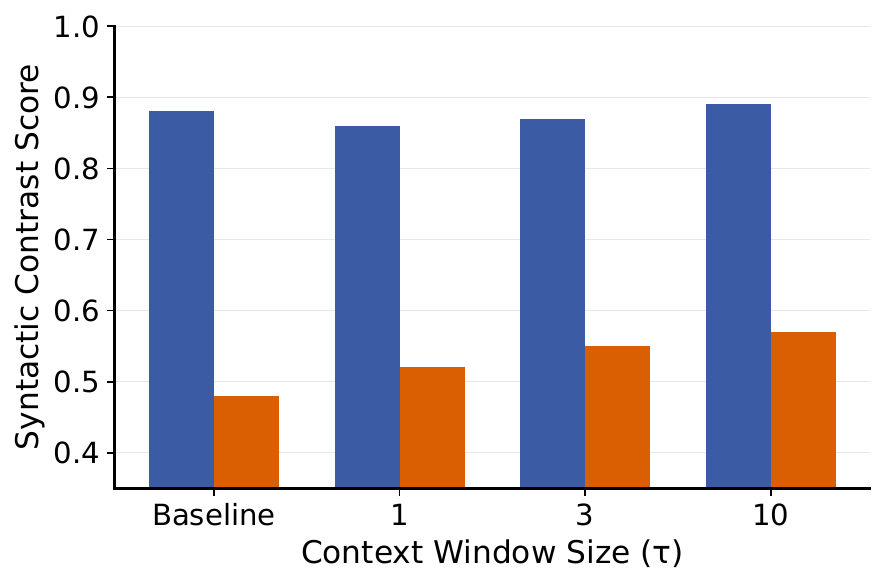}
    (h) RC -- phrase
\end{minipage}\hfill
\begin{minipage}[t]{0.32\textwidth}
    \centering
    \includegraphics[width=\linewidth]{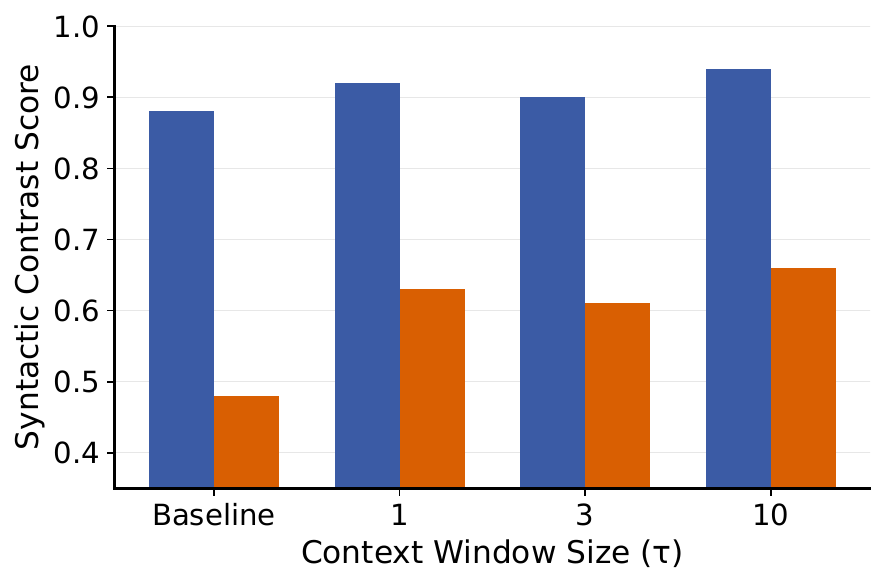}
    (i) RC -- sequence
\end{minipage}
\caption{Effect of context window size ($\tau \in \{1,3,10\}$) for the child-realistic model at $k=1$. Blue bars indicate high-frequency items and orange bars indicate low-frequency items; baseline results are shown for reference.}
\label{fig:app_context_child_k1}
\end{figure*}

\begin{figure*}[tp]
\centering
\begin{minipage}[t]{0.32\textwidth}
    \centering
    \includegraphics[width=\linewidth]{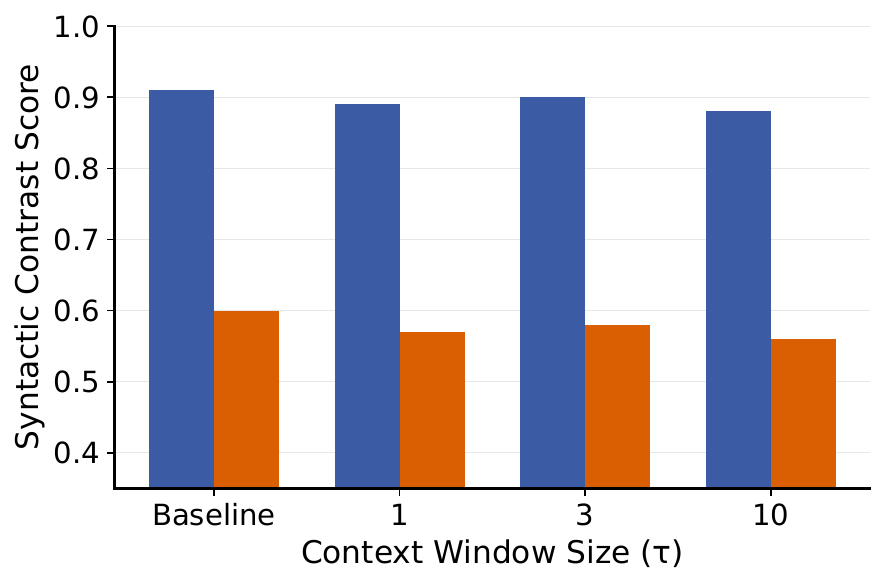}
    (a) SV -- token
\end{minipage}\hfill
\begin{minipage}[t]{0.32\textwidth}
    \centering
    \includegraphics[width=\linewidth]{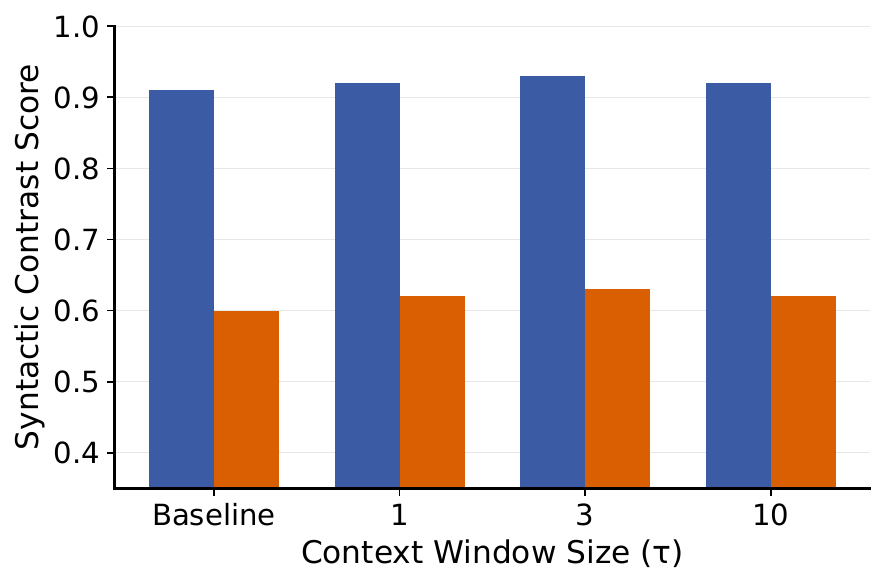}
    (b) SV -- phrase
\end{minipage}\hfill
\begin{minipage}[t]{0.32\textwidth}
    \centering
    \includegraphics[width=\linewidth]{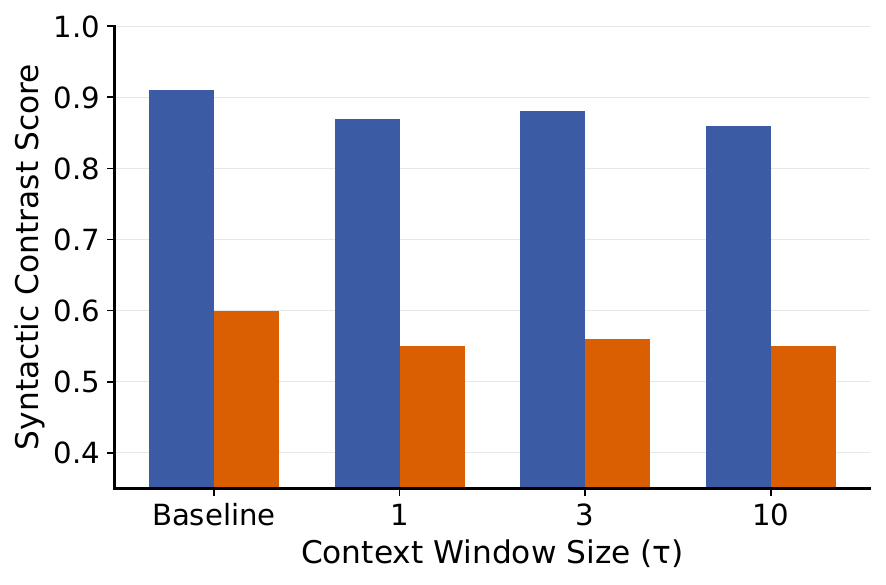}
    (c) SV -- sequence
\end{minipage}
\vspace{2mm}
\begin{minipage}[t]{0.32\textwidth}
    \centering
    \includegraphics[width=\linewidth]{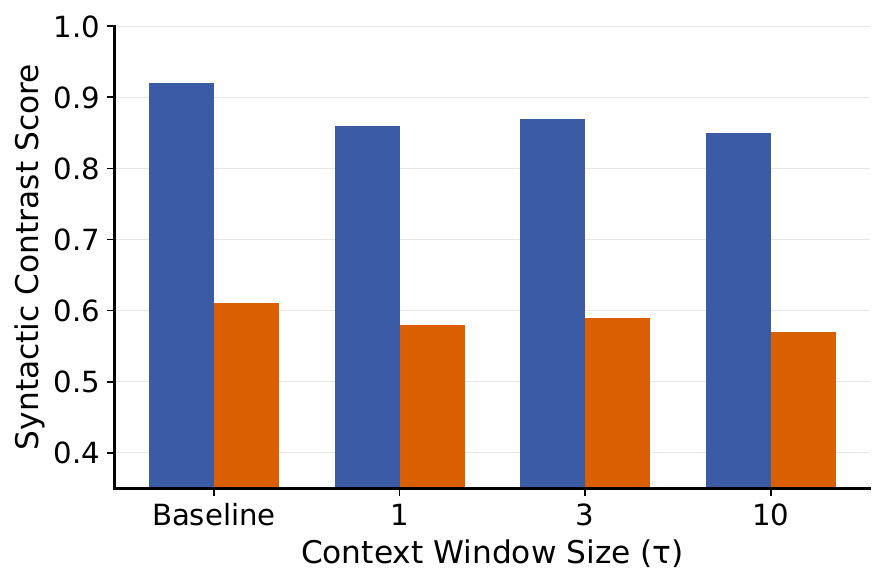}
    (d) Wh -- token
\end{minipage}\hfill
\begin{minipage}[t]{0.32\textwidth}
    \centering
    \includegraphics[width=\linewidth]{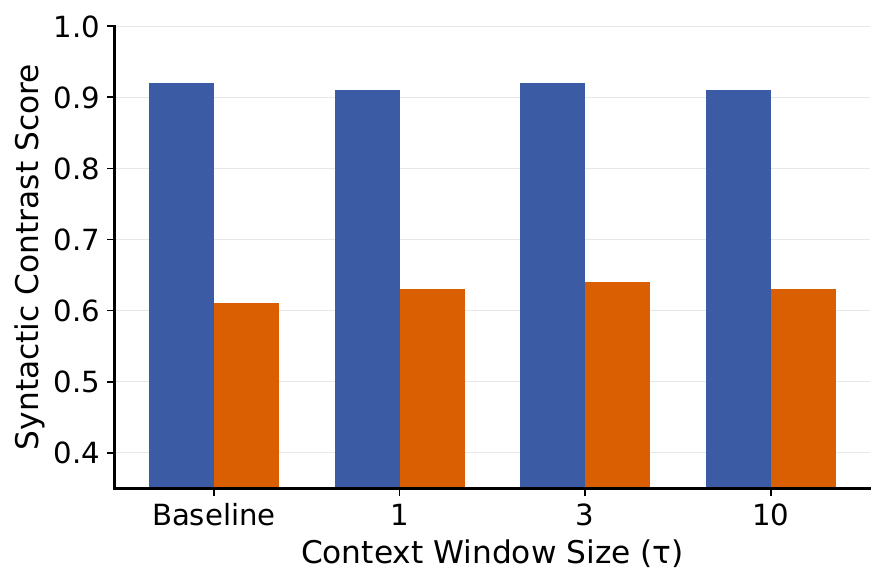}
    (e) Wh -- phrase
\end{minipage}\hfill
\begin{minipage}[t]{0.32\textwidth}
    \centering
    \includegraphics[width=\linewidth]{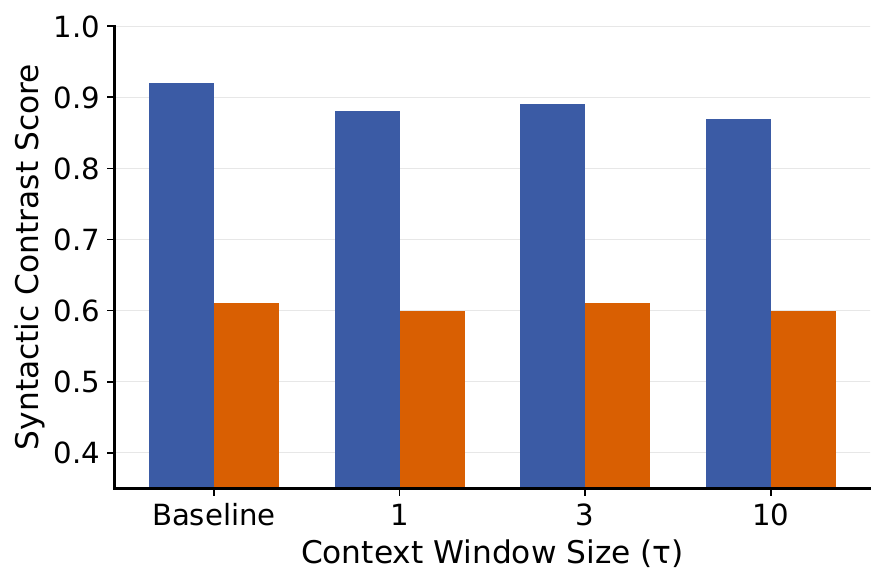}
    (f) Wh -- sequence
\end{minipage}
\vspace{2mm}
\begin{minipage}[t]{0.32\textwidth}
    \centering
    \includegraphics[width=\linewidth]{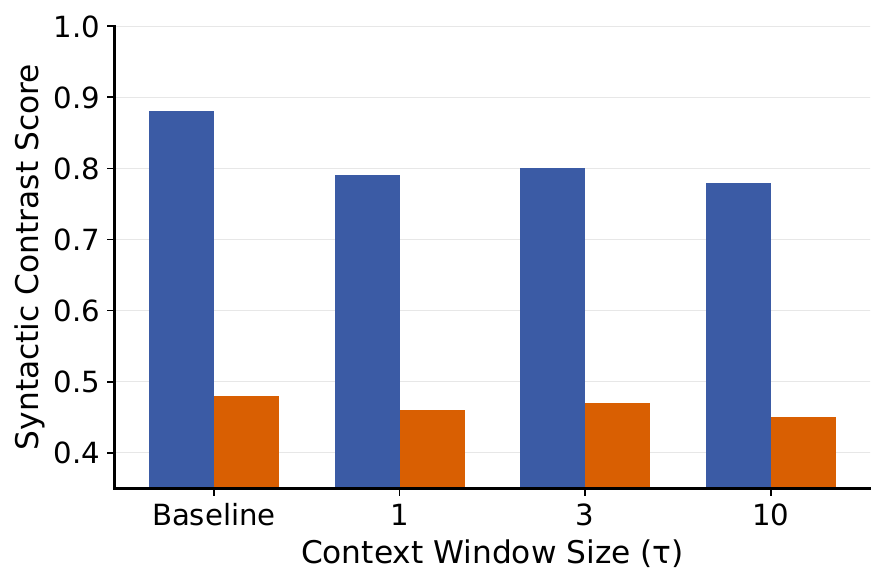}
    (g) RC -- token
\end{minipage}\hfill
\begin{minipage}[t]{0.32\textwidth}
    \centering
    \includegraphics[width=\linewidth]{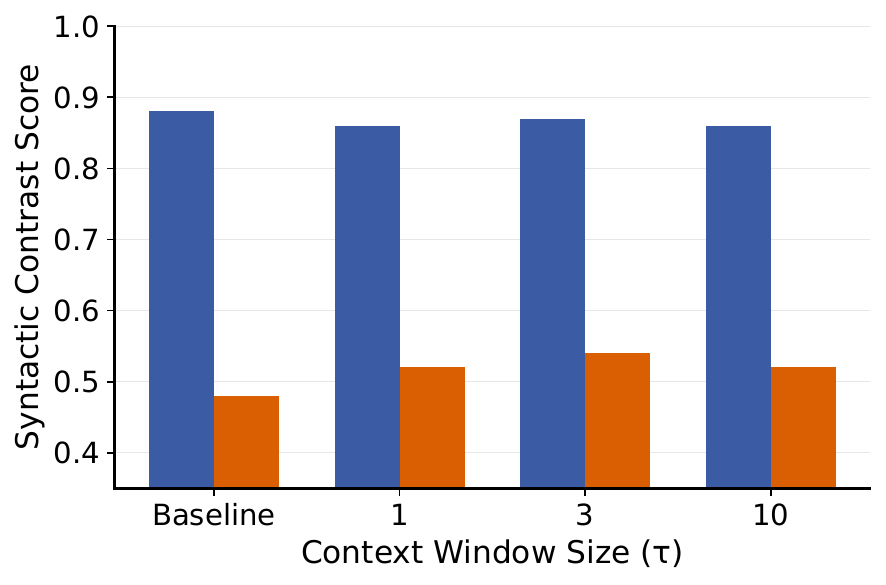}
    (h) RC -- phrase
\end{minipage}\hfill
\begin{minipage}[t]{0.32\textwidth}
    \centering
    \includegraphics[width=\linewidth]{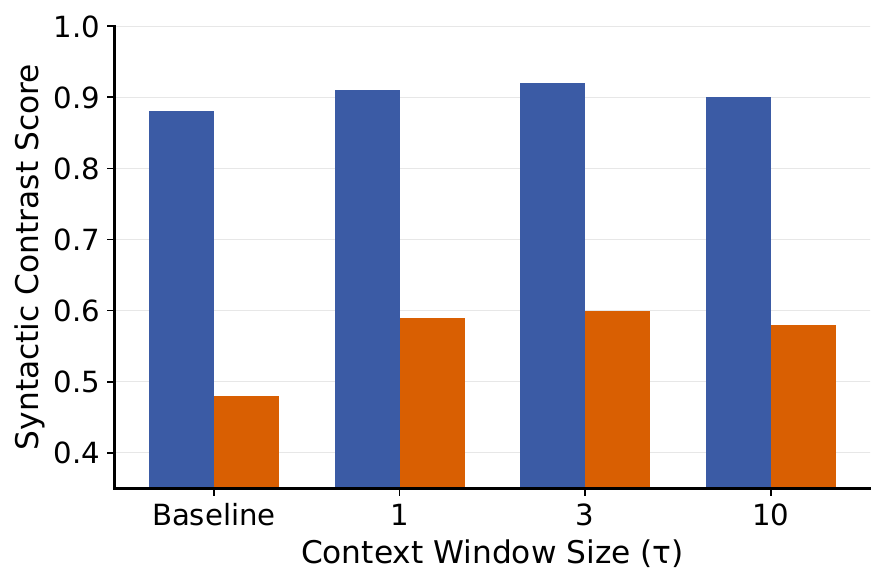}
    (i) RC -- sequence
\end{minipage}
\caption{Effect of context window size ($\tau \in \{1,3,10\}$) for the child-realistic model at $k=1024$. Blue bars indicate high-frequency items and orange bars indicate low-frequency items; baseline results are shown for reference.}
\label{fig:app_context_child_k1024}
\end{figure*}

\begin{figure*}[tp]
\centering
\begin{minipage}[t]{0.32\textwidth}
    \centering
    \includegraphics[width=\linewidth]{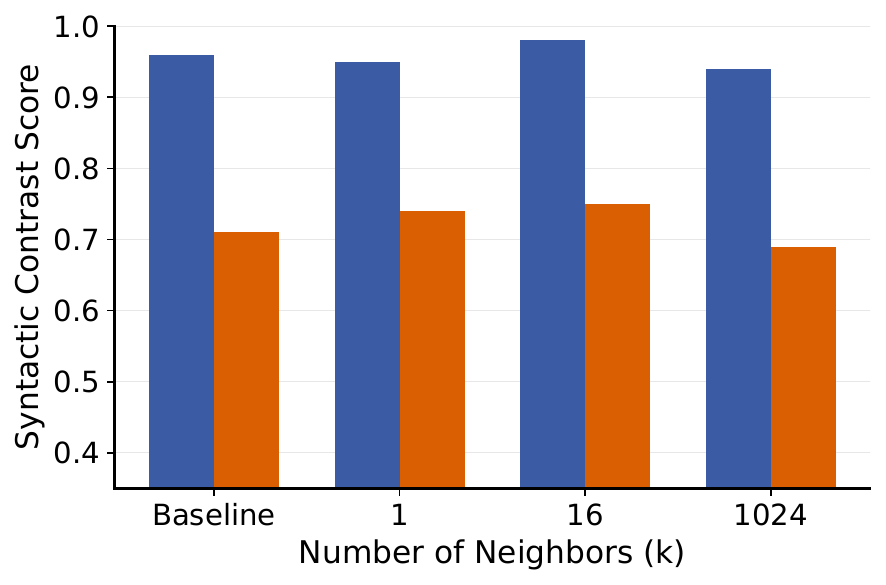}
    (a) SV -- token
\end{minipage}\hfill
\begin{minipage}[t]{0.32\textwidth}
    \centering
    \includegraphics[width=\linewidth]{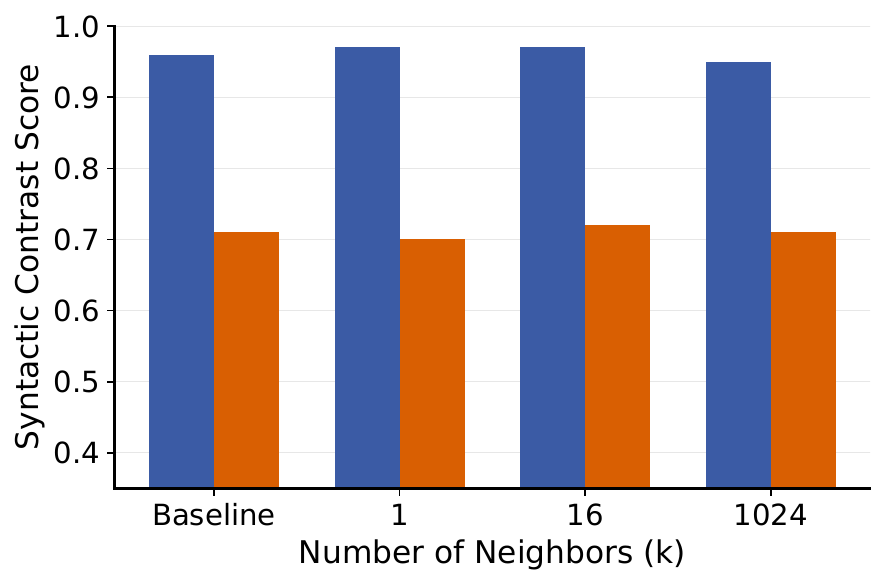}
    (b) SV -- phrase
\end{minipage}\hfill
\begin{minipage}[t]{0.32\textwidth}
    \centering
    \includegraphics[width=\linewidth]{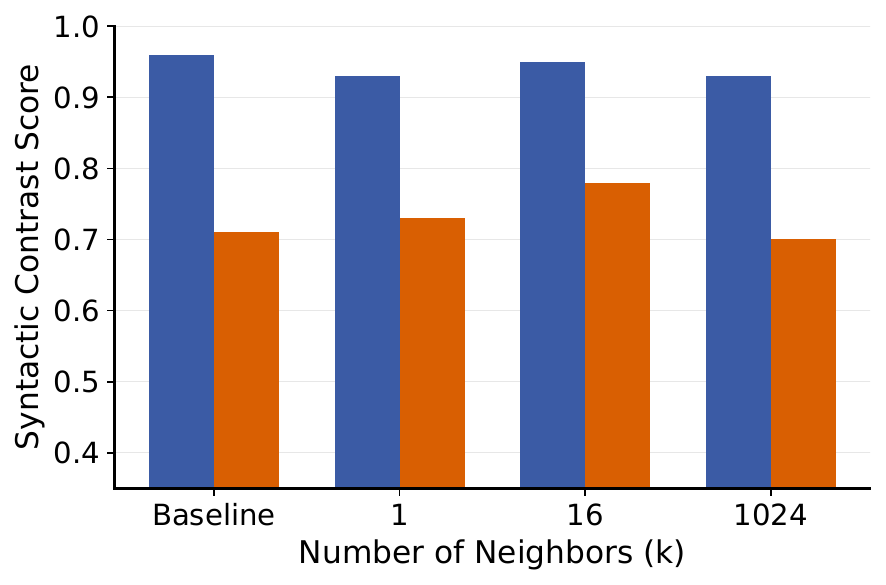}
    (c) SV -- sequence
\end{minipage}
\vspace{2mm}
\begin{minipage}[t]{0.32\textwidth}
    \centering
    \includegraphics[width=\linewidth]{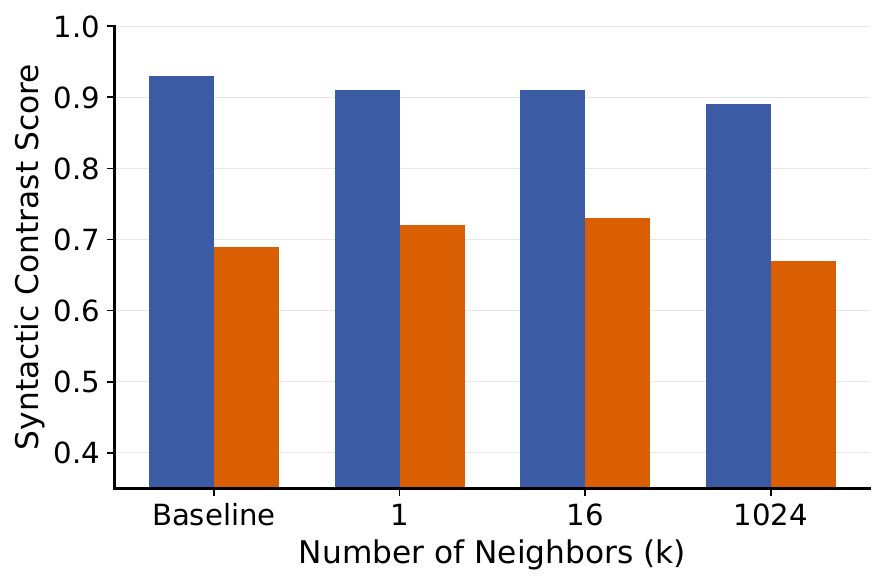}
    (d) Wh -- token
\end{minipage}\hfill
\begin{minipage}[t]{0.32\textwidth}
    \centering
    \includegraphics[width=\linewidth]{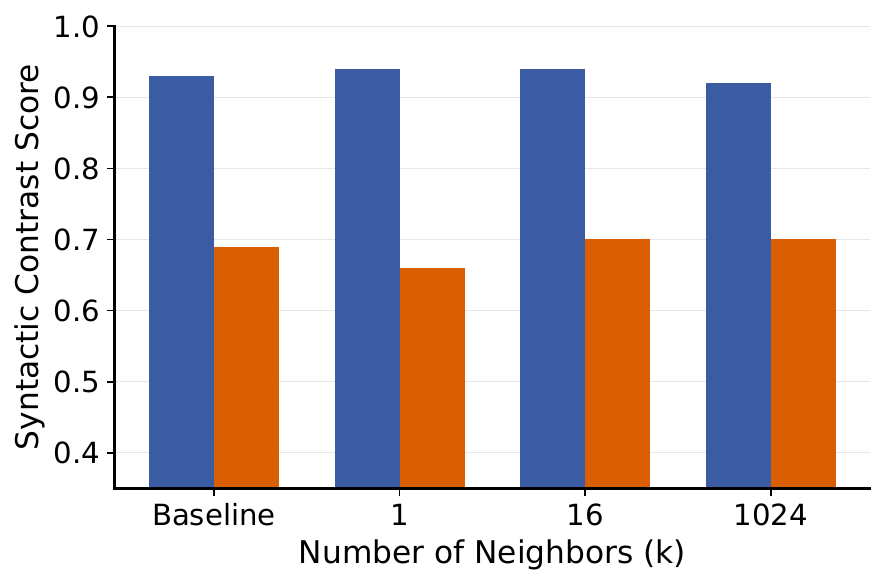}
    (e) Wh -- phrase
\end{minipage}\hfill
\begin{minipage}[t]{0.32\textwidth}
    \centering
    \includegraphics[width=\linewidth]{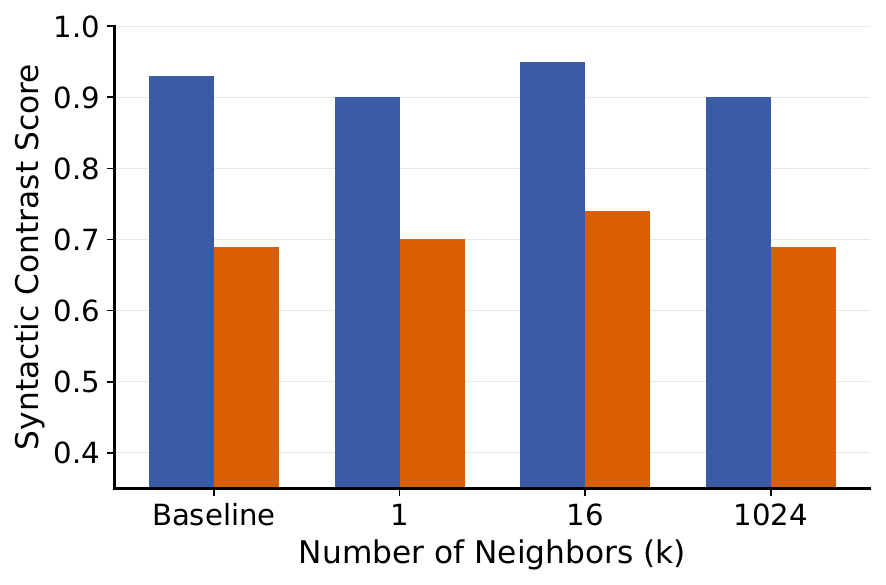}
    (f) Wh -- sequence
\end{minipage}
\vspace{2mm}
\begin{minipage}[t]{0.32\textwidth}
    \centering
    \includegraphics[width=\linewidth]{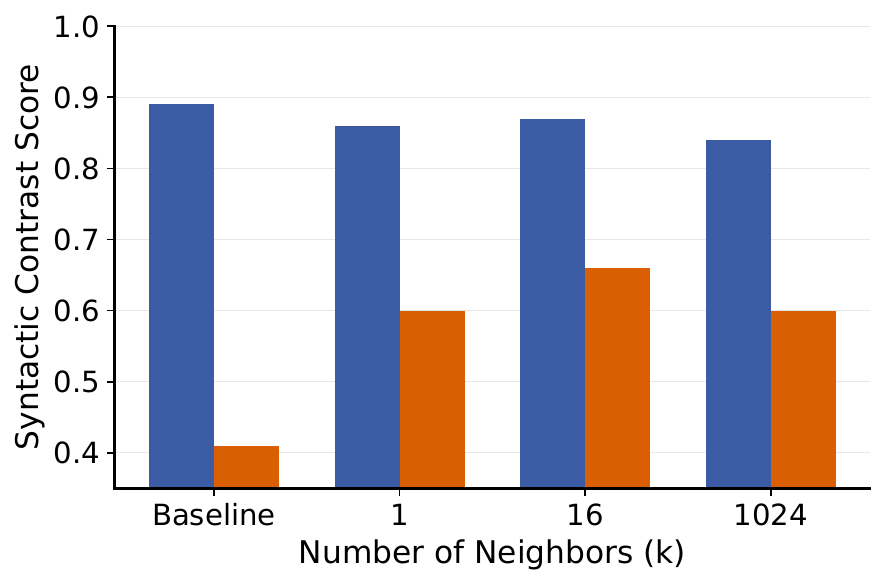}
    (g) RC -- token
\end{minipage}\hfill
\begin{minipage}[t]{0.32\textwidth}
    \centering
    \includegraphics[width=\linewidth]{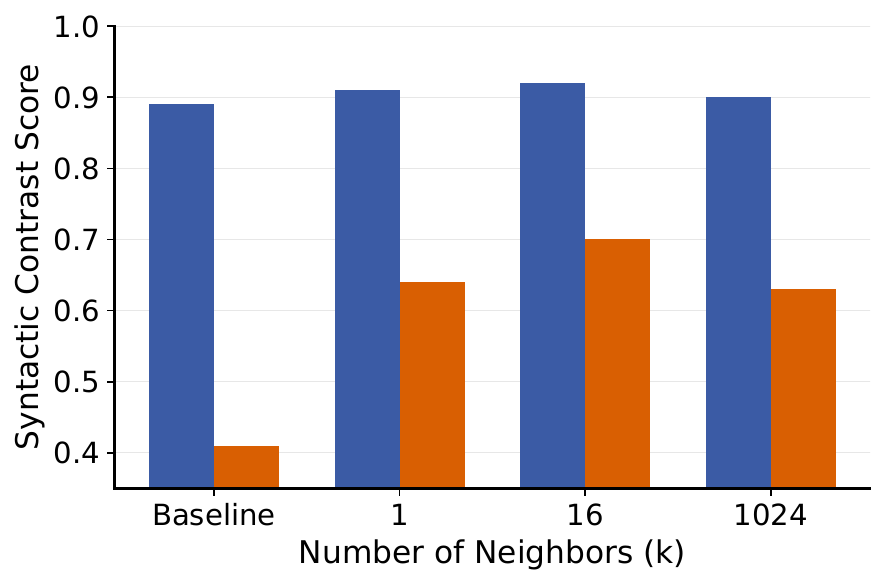}
    (h) RC -- phrase
\end{minipage}\hfill
\begin{minipage}[t]{0.32\textwidth}
    \centering
    \includegraphics[width=\linewidth]{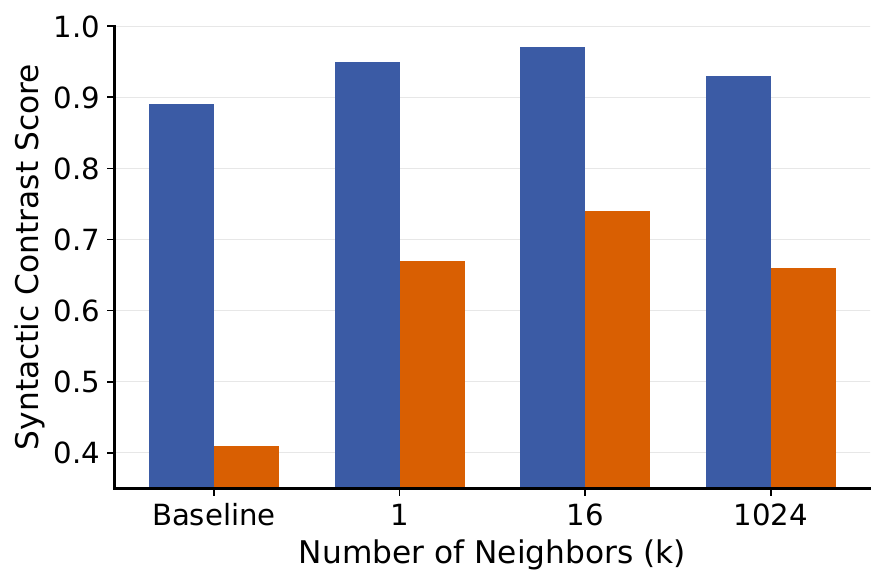}
    (i) RC -- sequence
\end{minipage}
\caption{Effect of number of retrieved neighbors ($k \in \{1,16,1024\}$) for GPT-2 XL at $\tau=1$. Blue bars indicate high-frequency items and orange bars indicate low-frequency items; baseline results are shown for reference.}
\label{fig:app_neighbor_gpt2xl_tau1}
\end{figure*}

\begin{figure*}[tp]
\centering
\begin{minipage}[t]{0.32\textwidth}
    \centering
    \includegraphics[width=\linewidth]{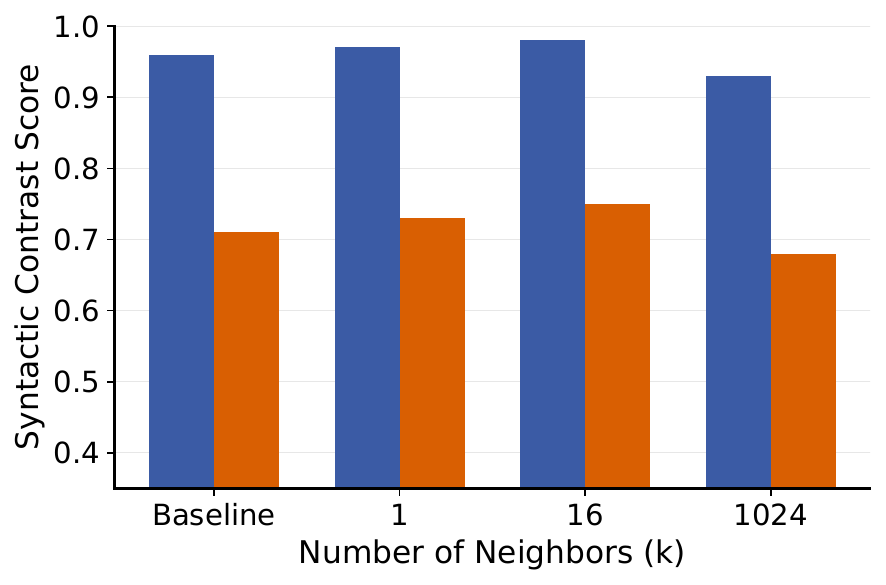}
    (a) SV -- token
\end{minipage}\hfill
\begin{minipage}[t]{0.32\textwidth}
    \centering
    \includegraphics[width=\linewidth]{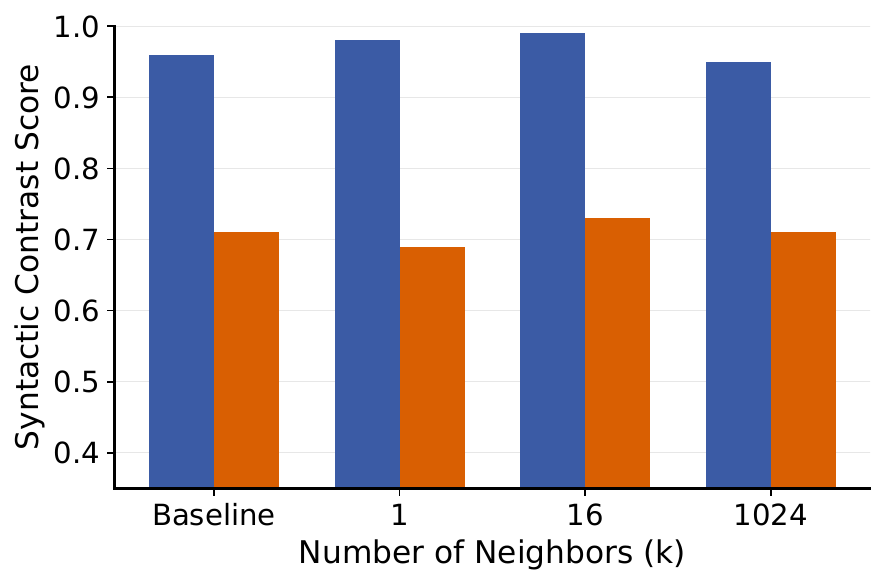}
    (b) SV -- phrase
\end{minipage}\hfill
\begin{minipage}[t]{0.32\textwidth}
    \centering
    \includegraphics[width=\linewidth]{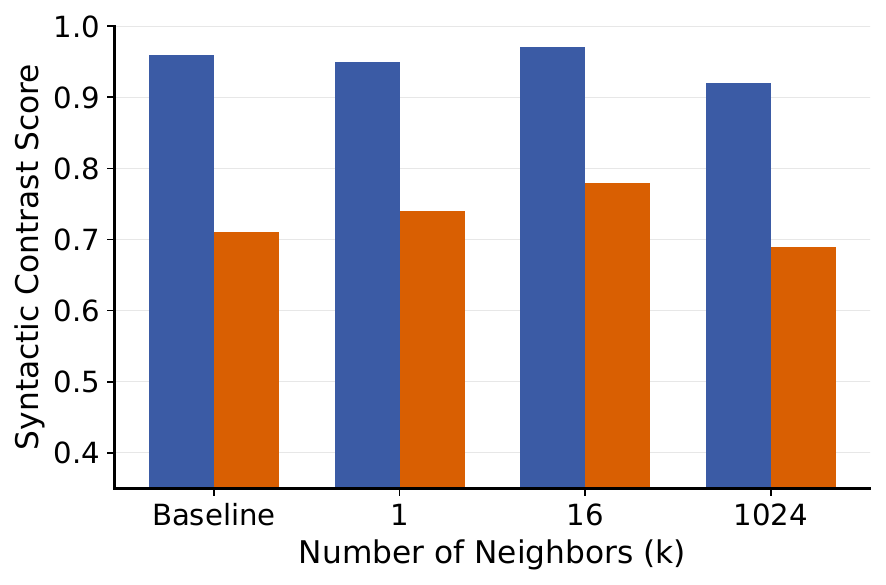}
    (c) SV -- sequence
\end{minipage}
\vspace{2mm}
\begin{minipage}[t]{0.32\textwidth}
    \centering
    \includegraphics[width=\linewidth]{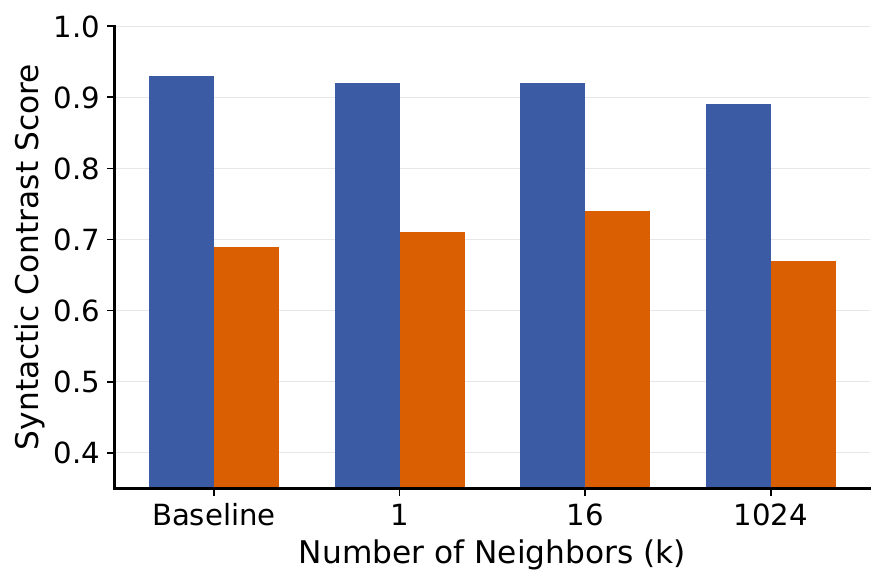}
    (d) Wh -- token
\end{minipage}\hfill
\begin{minipage}[t]{0.32\textwidth}
    \centering
    \includegraphics[width=\linewidth]{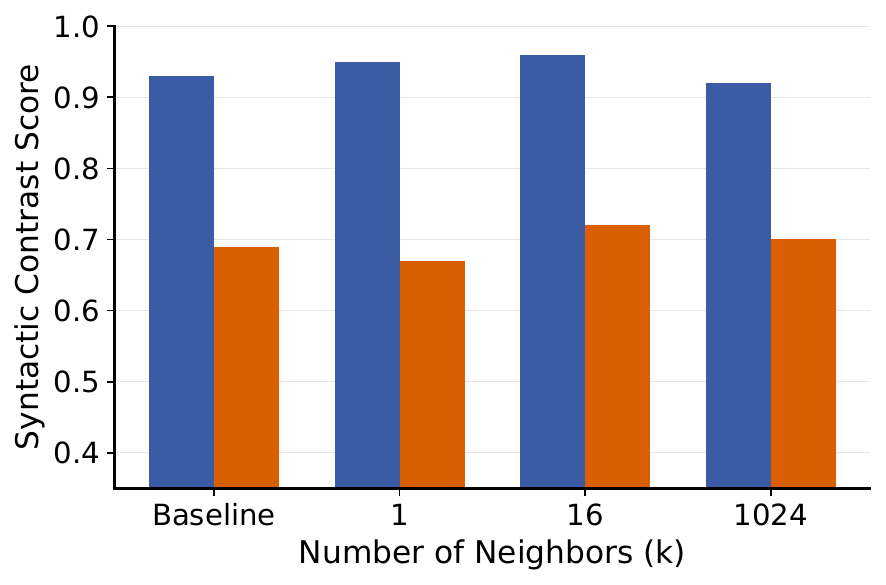}
    (e) Wh -- phrase
\end{minipage}\hfill
\begin{minipage}[t]{0.32\textwidth}
    \centering
    \includegraphics[width=\linewidth]{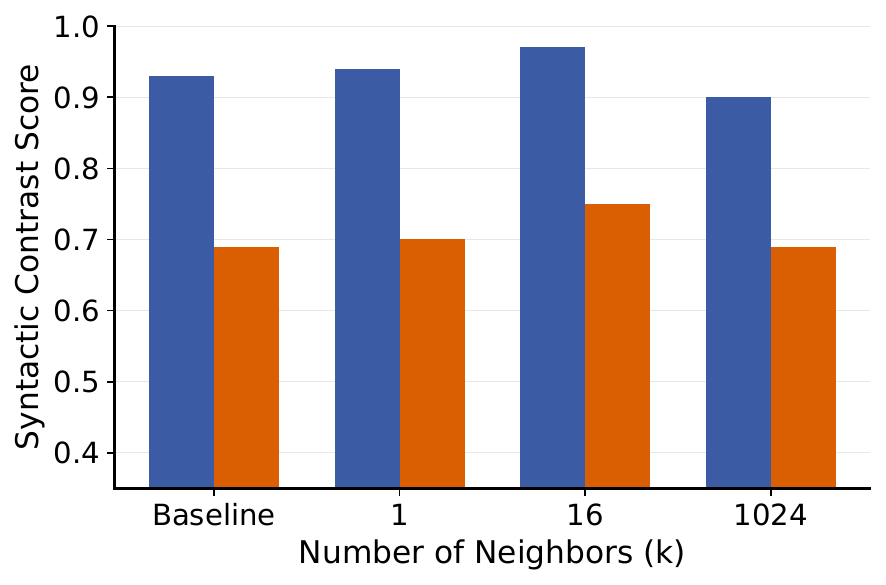}
    (f) Wh -- sequence
\end{minipage}
\vspace{2mm}
\begin{minipage}[t]{0.32\textwidth}
    \centering
    \includegraphics[width=\linewidth]{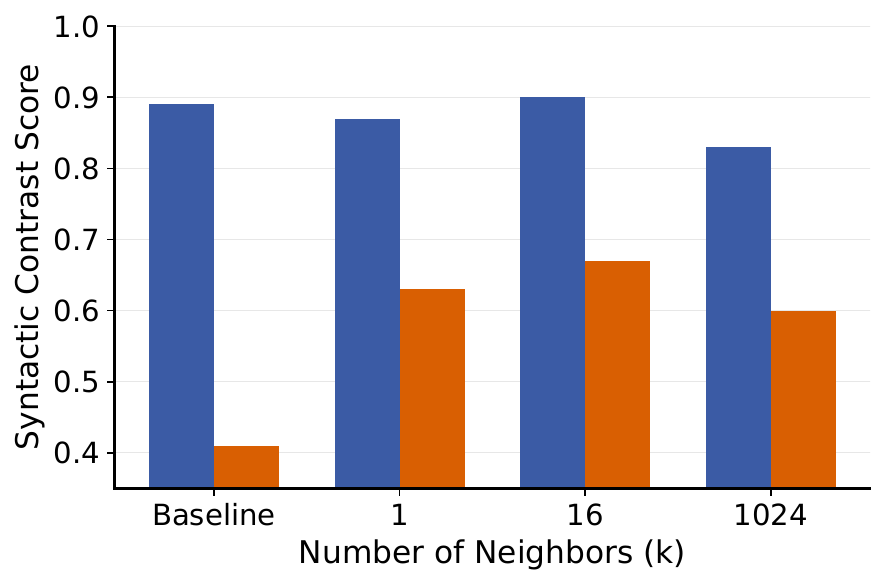}
    (g) RC -- token
\end{minipage}\hfill
\begin{minipage}[t]{0.32\textwidth}
    \centering
    \includegraphics[width=\linewidth]{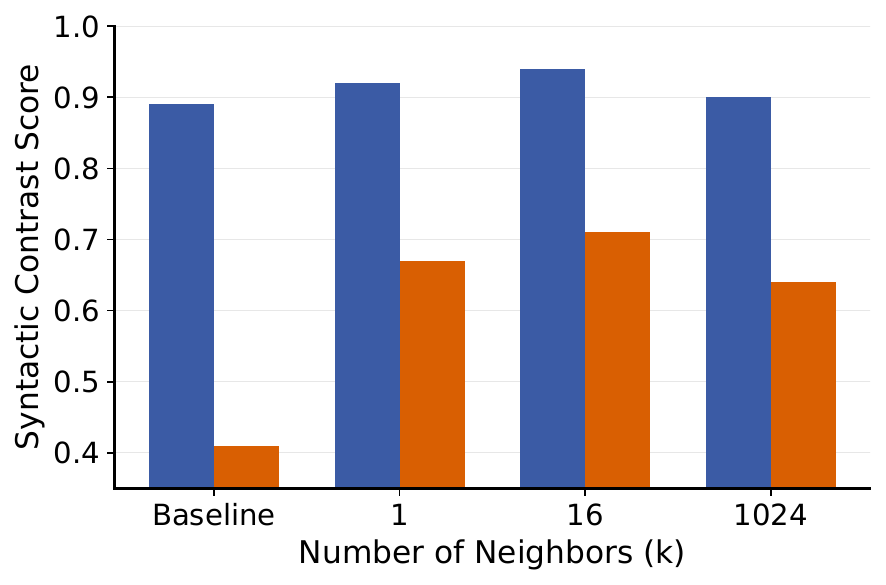}
    (h) RC -- phrase
\end{minipage}\hfill
\begin{minipage}[t]{0.32\textwidth}
    \centering
    \includegraphics[width=\linewidth]{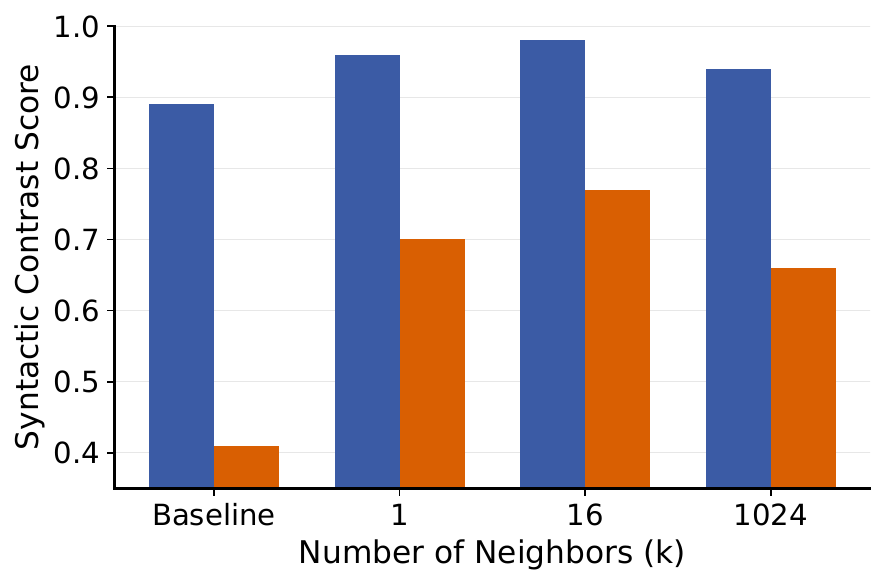}
    (i) RC -- sequence
\end{minipage}
\caption{Effect of number of retrieved neighbors ($k \in \{1,16,1024\}$) for GPT-2 XL at $\tau=10$. Blue bars indicate high-frequency items and orange bars indicate low-frequency items; baseline results are shown for reference.}
\label{fig:app_neighbor_gpt2xl_tau10}
\end{figure*}

\begin{figure*}[ht]
\centering
\begin{minipage}[t]{0.32\textwidth}
    \centering
    \includegraphics[width=\linewidth]{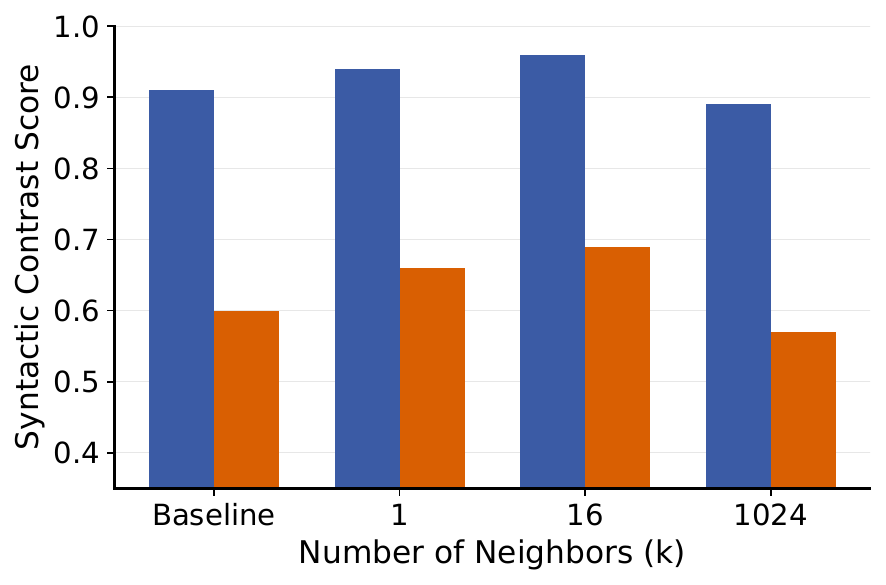}
    (a) SV -- token
\end{minipage}\hfill
\begin{minipage}[t]{0.32\textwidth}
    \centering
    \includegraphics[width=\linewidth]{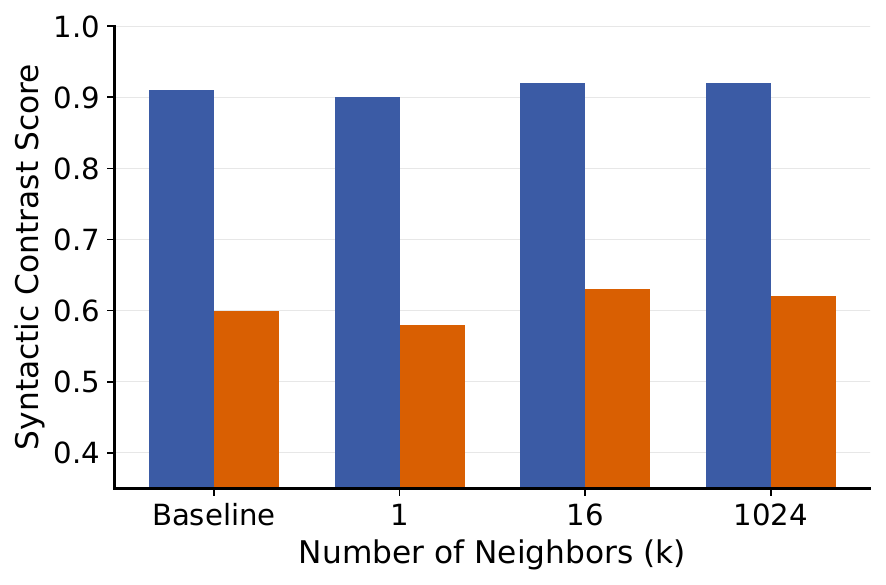}
    (b) SV -- phrase
\end{minipage}\hfill
\begin{minipage}[t]{0.32\textwidth}
    \centering
    \includegraphics[width=\linewidth]{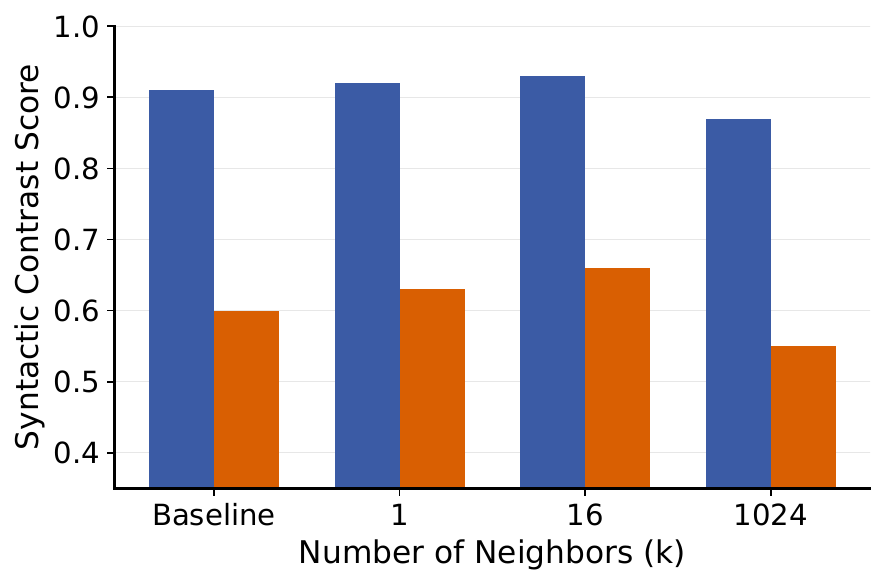}
    (c) SV -- sequence
\end{minipage}
\vspace{2mm}
\begin{minipage}[t]{0.32\textwidth}
    \centering
    \includegraphics[width=\linewidth]{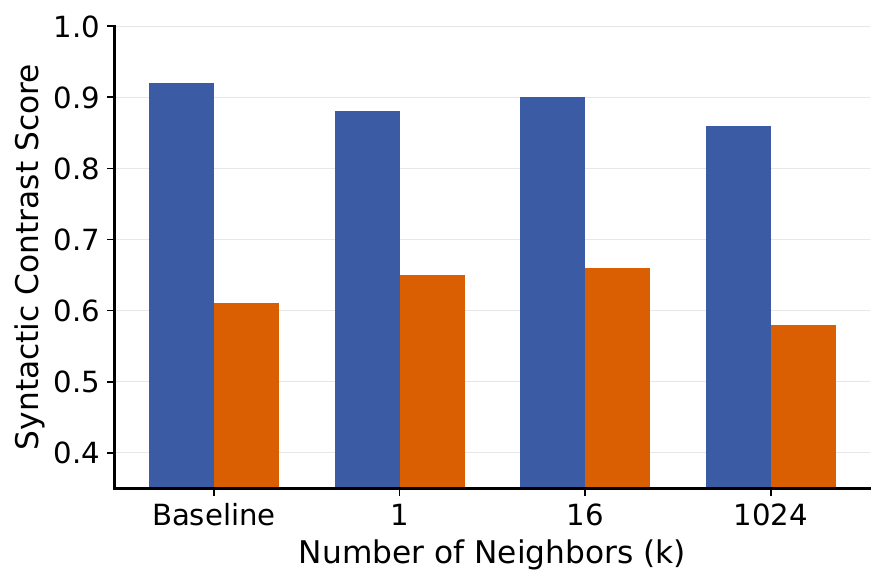}
    (d) Wh -- token
\end{minipage}\hfill
\begin{minipage}[t]{0.32\textwidth}
    \centering
    \includegraphics[width=\linewidth]{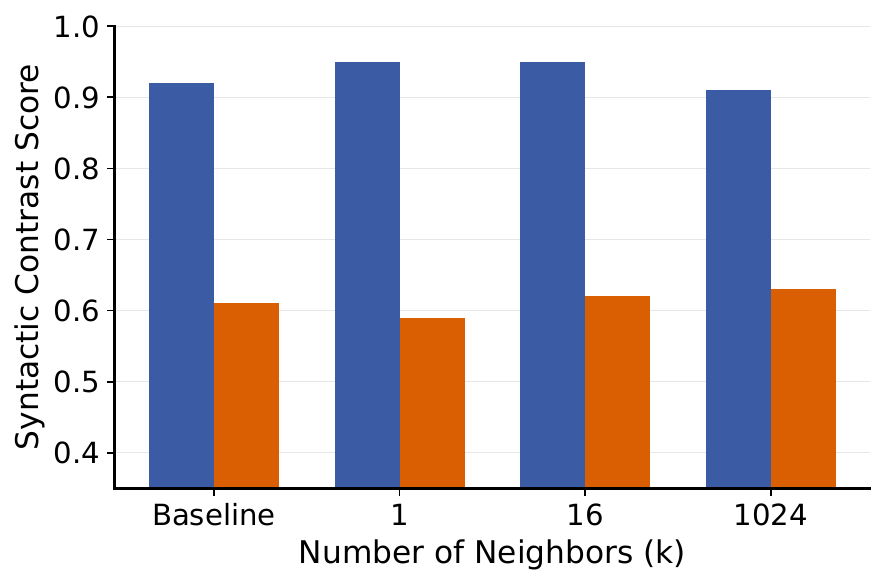}
    (e) Wh -- phrase
\end{minipage}\hfill
\begin{minipage}[t]{0.32\textwidth}
    \centering
    \includegraphics[width=\linewidth]{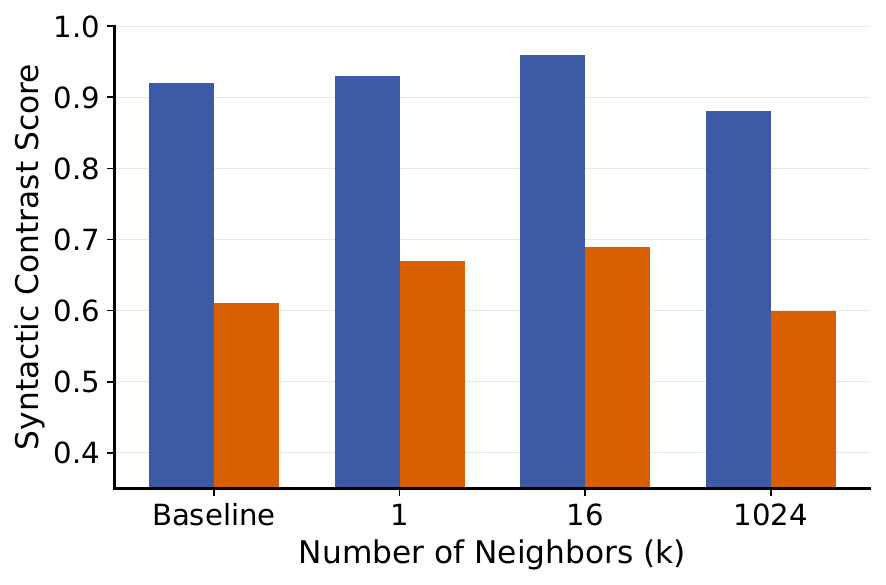}
    (f) Wh -- sequence
\end{minipage}
\vspace{2mm}
\begin{minipage}[t]{0.32\textwidth}
    \centering
    \includegraphics[width=\linewidth]{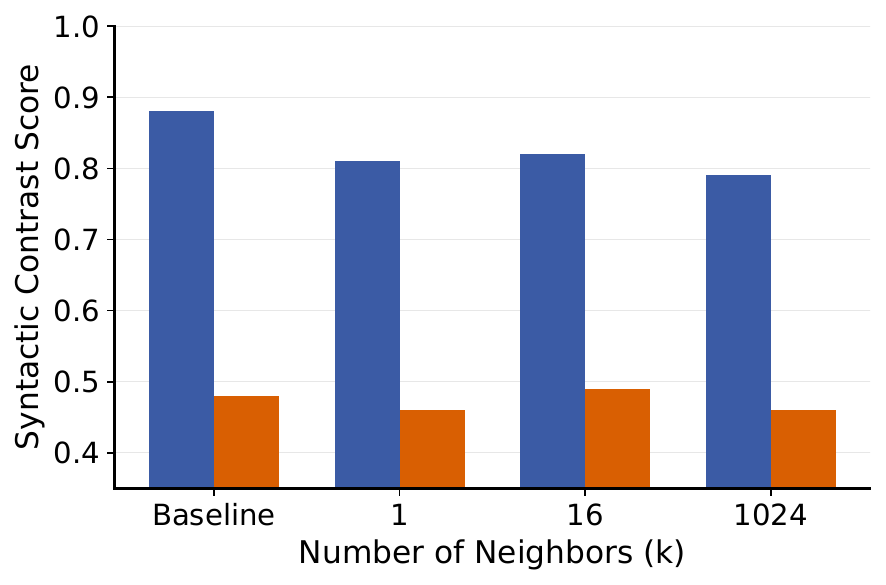}
    (g) RC -- token
\end{minipage}\hfill
\begin{minipage}[t]{0.32\textwidth}
    \centering
    \includegraphics[width=\linewidth]{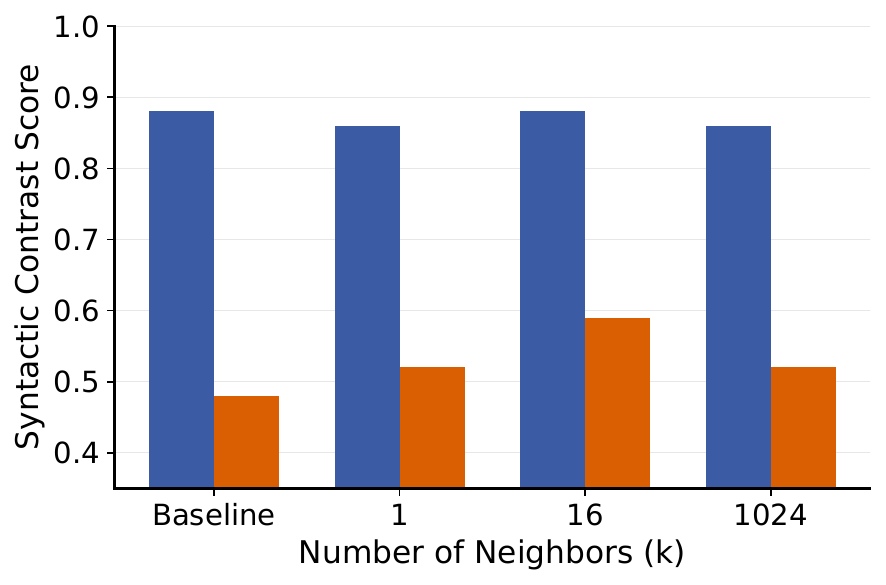}
    (h) RC -- phrase
\end{minipage}\hfill
\begin{minipage}[t]{0.32\textwidth}
    \centering
    \includegraphics[width=\linewidth]{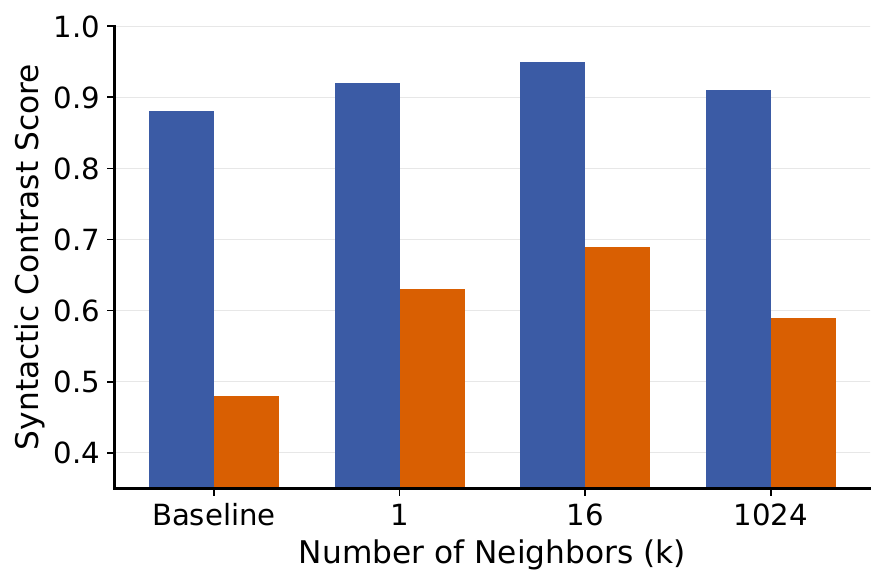}
    (i) RC -- sequence
\end{minipage}
\caption{Effect of number of retrieved neighbors ($k \in \{1,16,1024\}$) for the child-realistic model at $\tau=1$. Blue bars indicate high-frequency items and orange bars indicate low-frequency items; baseline results are shown for reference.}
\label{fig:app_neighbor_child_tau1}
\end{figure*}

\begin{figure*}[ht]
\centering
\begin{minipage}[t]{0.32\textwidth}
    \centering
    \includegraphics[width=\linewidth]{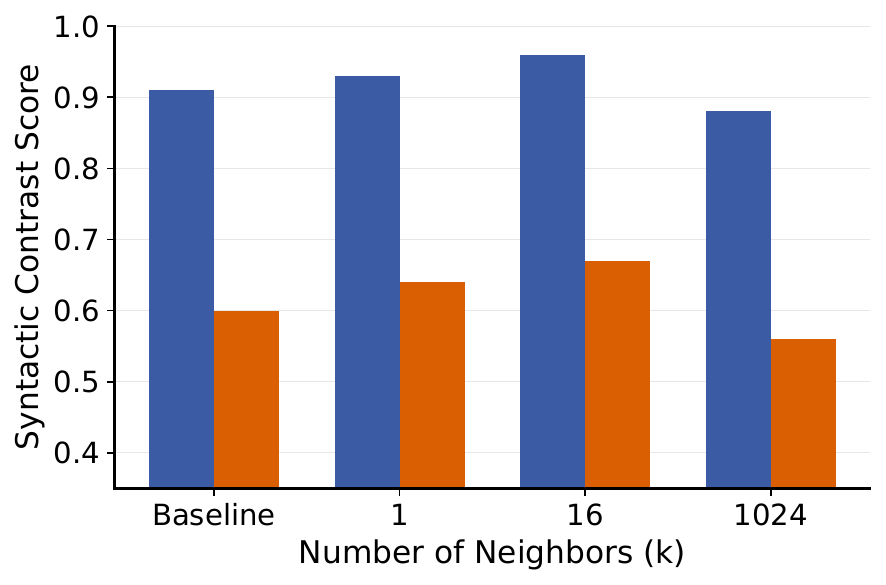}
    (a) SV -- token
\end{minipage}\hfill
\begin{minipage}[t]{0.32\textwidth}
    \centering
    \includegraphics[width=\linewidth]{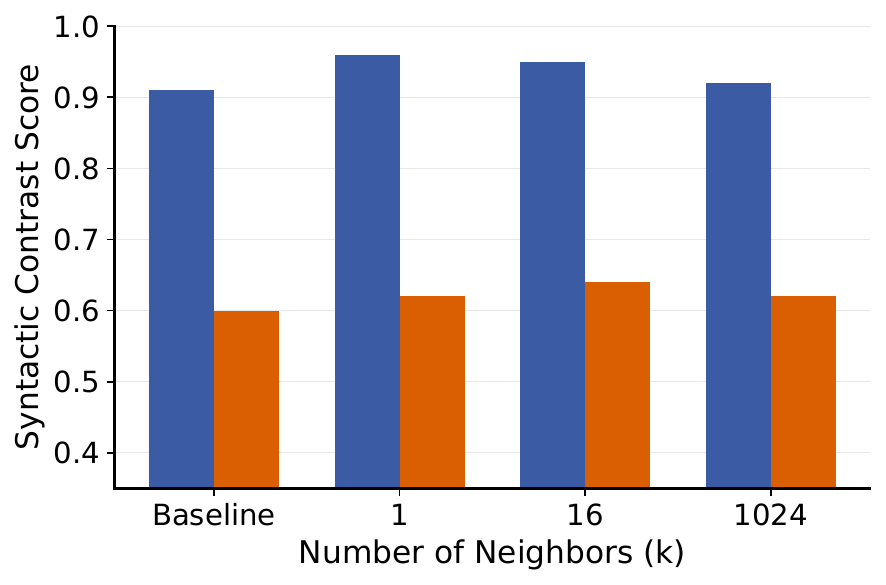}
    (b) SV -- phrase
\end{minipage}\hfill
\begin{minipage}[t]{0.32\textwidth}
    \centering
    \includegraphics[width=\linewidth]{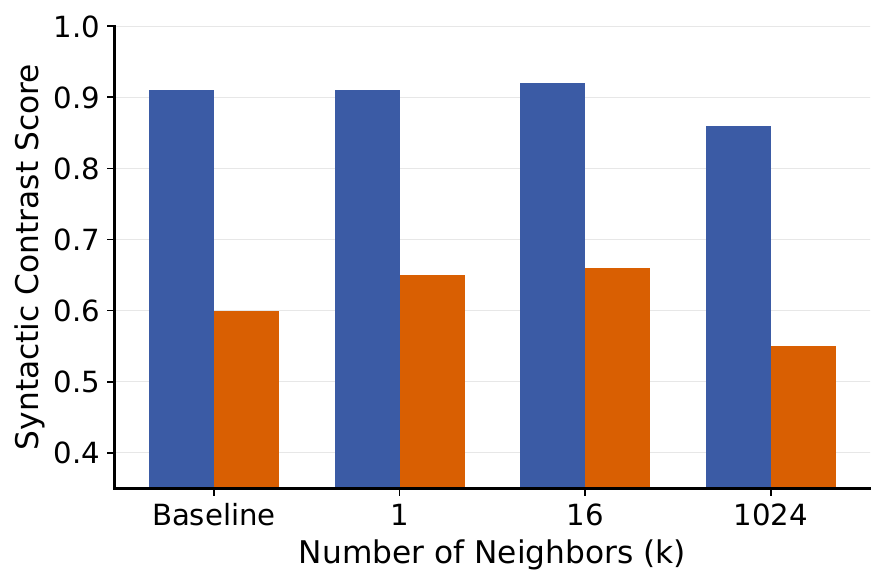}
    (c) SV -- sequence
\end{minipage}
\vspace{2mm}
\begin{minipage}[t]{0.32\textwidth}
    \centering
    \includegraphics[width=\linewidth]{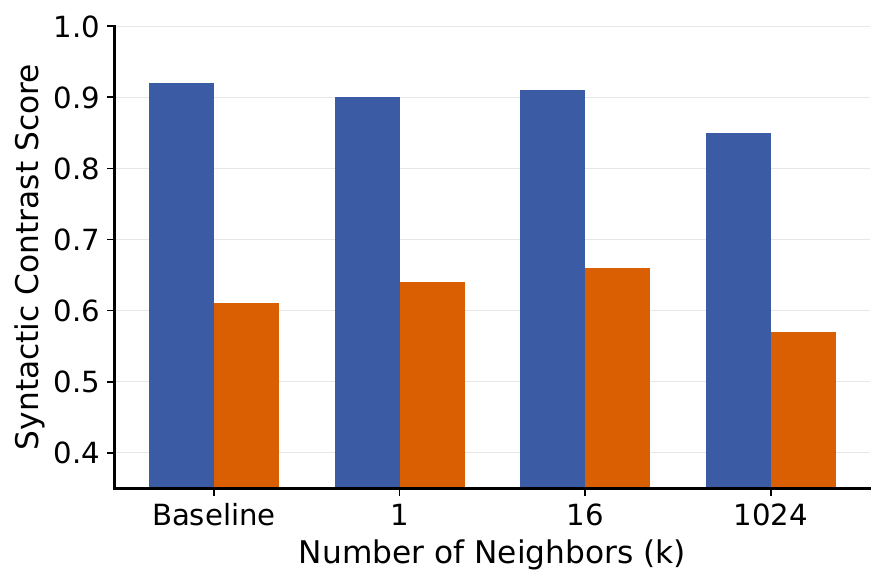}
    (d) Wh -- token
\end{minipage}\hfill
\begin{minipage}[t]{0.32\textwidth}
    \centering
    \includegraphics[width=\linewidth]{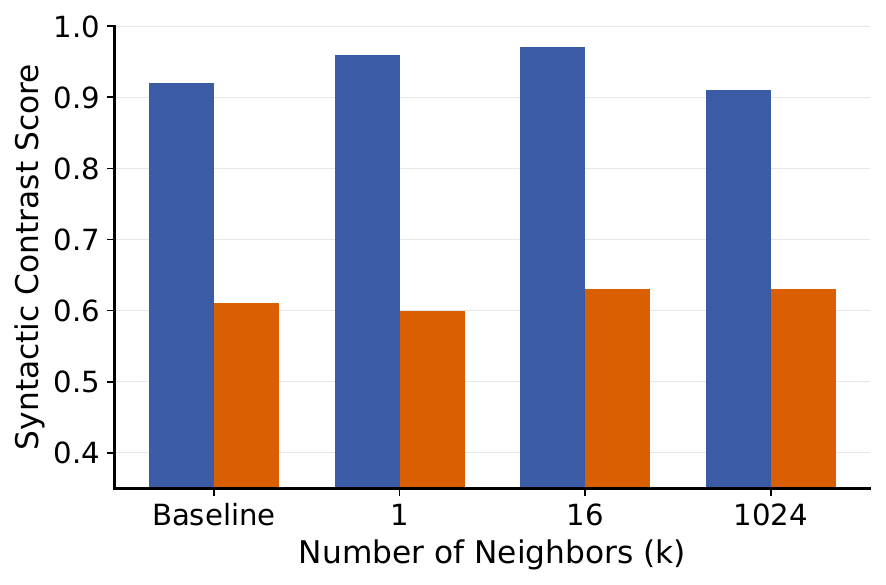}
    (e) Wh -- phrase
\end{minipage}\hfill
\begin{minipage}[t]{0.32\textwidth}
    \centering
    \includegraphics[width=\linewidth]{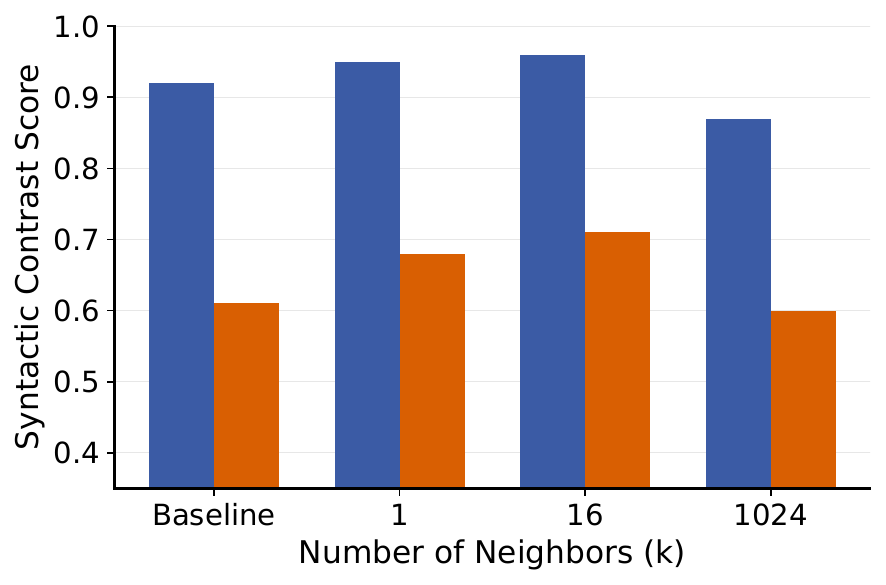}
    (f) Wh -- sequence
\end{minipage}
\vspace{2mm}
\begin{minipage}[t]{0.32\textwidth}
    \centering
    \includegraphics[width=\linewidth]{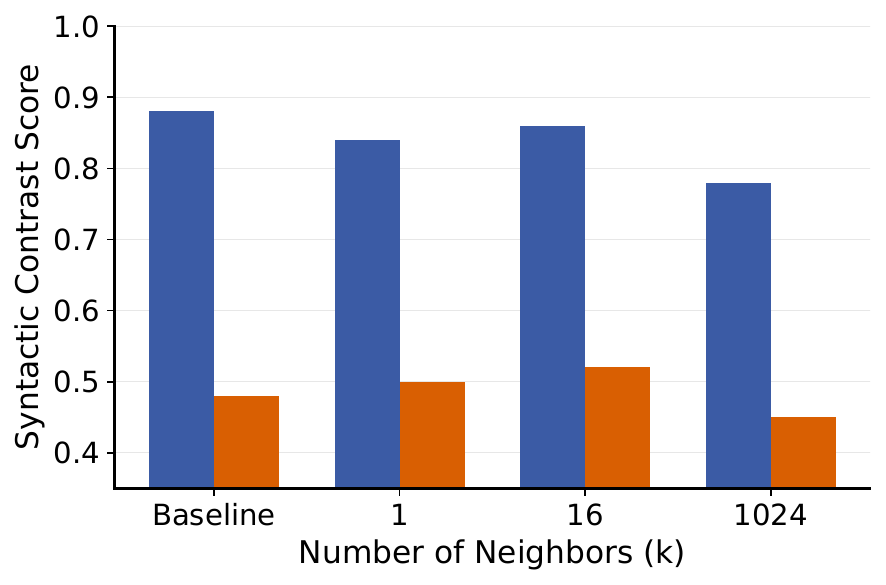}
    (g) RC -- token
\end{minipage}\hfill
\begin{minipage}[t]{0.32\textwidth}
    \centering
    \includegraphics[width=\linewidth]{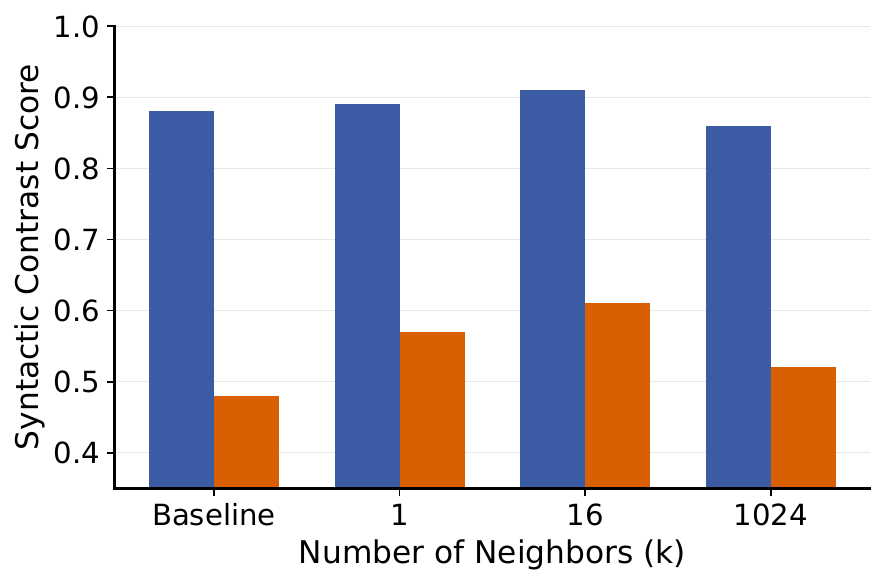}
    (h) RC -- phrase
\end{minipage}\hfill
\begin{minipage}[t]{0.32\textwidth}
    \centering
    \includegraphics[width=\linewidth]{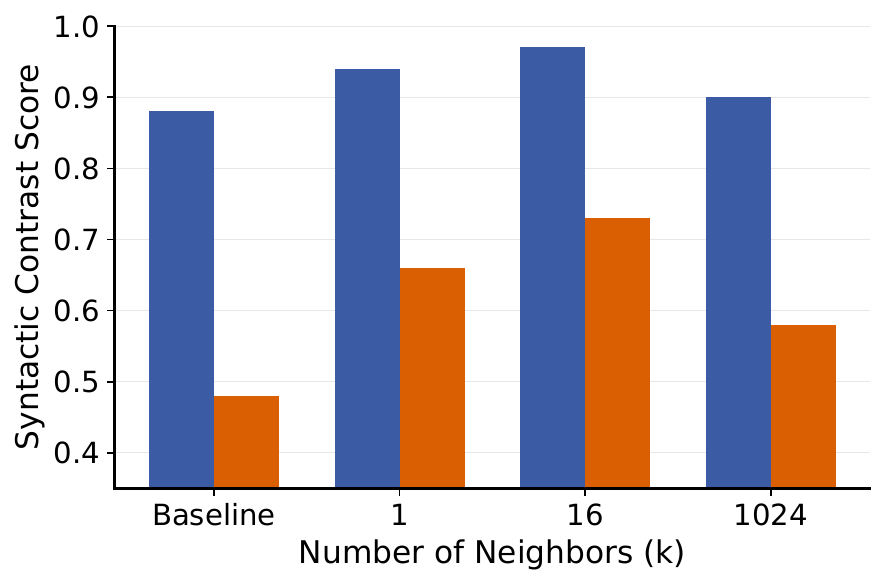}
    (i) RC -- sequence
\end{minipage}
\caption{Effect of number of retrieved neighbors ($k \in \{1,16,1024\}$) for the child-realistic model at $\tau=10$. Blue bars indicate high-frequency items and orange bars indicate low-frequency items; baseline results are shown for reference.}
\label{fig:app_neighbor_child_tau10}
\end{figure*}

This section provides additional results for the retrieval configuration analyses in the main text. All results use the \textit{Full Retrieval} datastore defined in Section~\ref{subsec:datastores}; baseline results are shown for reference. Figures~\ref{fig:app_context_gpt2xl_k1}--\ref{fig:app_context_child_k1024} vary context-window size ($\tau \in \{1, 3, 10\}$) at $k \in \{1, 1024\}$, whereas Figures~\ref{fig:app_neighbor_gpt2xl_tau1}--\ref{fig:app_neighbor_child_tau10} vary the number of retrieved neighbors ($k \in \{1, 16, 1024\}$) at $\tau \in \{1, 10\}$. Together with the main analyses, these figures cover the evaluated combinations of neighbor count, context-window size, and retrieval granularity for the displayed high- and low-frequency bands in both pretraining settings. Given many differences between individual configurations are small, we focus here on broad patterns rather than numerical rankings within individual panels.

\noindent \textbf{Effects of context window size.}
Figures~\ref{fig:app_context_gpt2xl_k1}--\ref{fig:app_context_child_k1024} extend Figure~\ref{fig:context_effects} to $k \in \{1, 1024\}$ and to all retrieval granularities. As in the main analysis, accuracy is non-monotonic with context window size. The direction and magnitude of the effects vary by number of neighbors, retrieval granularity, phenomenon, and frequency band; and are often small. These additional results therefore do not identify a context window size that is uniformly preferred across conditions.

\noindent \textbf{Effects of the number of retrieved neighbors.}
Figures~\ref{fig:app_neighbor_gpt2xl_tau1}--\ref{fig:app_neighbor_child_tau10} extend Figure~\ref{fig:neighbor_effects} to $\tau \in \{1, 10\}$ and to all retrieval granularities. The results are broadly consistent with the pattern reported in the main text: for low-frequency items, the intermediate setting $k=16$ generally yields the strongest performance among the tested values, whereas effects on high-frequency items are smaller and more mixed. The magnitude and direction of the effects vary across conditions, and the effects are often small. We therefore treat $k=16$ as a useful representative setting rather than as a universally optimal neighbor count. Comparisons between $\tau=1$ and $\tau=10$ similarly provide no evidence for an interesting interaction effect with context window size.

\end{document}